\documentclass[journal]{IEEEtran}
\usepackage{lineno,hyperref}
\usepackage{amsmath}
\usepackage{amssymb}
\usepackage{slashbox}
\usepackage{graphicx}

\usepackage{algorithm}
\usepackage{algorithmic}
\usepackage{amsfonts}
\usepackage{subfig}
\usepackage{helvet}  %
\usepackage{courier}  %
\usepackage{multirow}

\usepackage{cite}
\usepackage{color}

\usepackage{microtype}
\usepackage{booktabs} %
\usepackage{url}  %

\usepackage[utf8]{inputenc} %
\usepackage[T1]{fontenc}    %

\usepackage{enumitem}

\newcommand{\Bmat}[0]{\ensuremath{{\bf B}} }

\newcommand{\Dmat}[0]{\ensuremath{{\bf D}} }

\newcommand{\Gmat}[0]{\ensuremath{{\bf G}} }

\newcommand{\Imat}[0]{\ensuremath{{\bf I}} }

\newcommand{\Qmat}[0]{\ensuremath{{\bf Q}} }

\newcommand{\Tmat}[0]{\ensuremath{{\bf T}} }

\newcommand{\Wmat}[0]{\ensuremath{{\bf W}} }
\newcommand{\Xmat}[0]{\ensuremath{{\bf X}} }
\newcommand{\Ymat}[0]{\ensuremath{{\bf Y}} }
\newcommand{\Zmat}[0]{\ensuremath{{\bf Z}} }

\newcommand{\bv}[0]{\ensuremath{\boldsymbol{b}} }

\newcommand{\hv}[0]{\ensuremath{\boldsymbol{h}} }

\newcommand{\kv}[0]{\ensuremath{\boldsymbol{k}} }

\newcommand{\mv}[0]{\ensuremath{\boldsymbol{m}} }

\newcommand{\pv}[0]{\ensuremath{\boldsymbol{p}} }

\newcommand{\rv}[0]{\ensuremath{\boldsymbol{r}} }
\newcommand{\sv}[0]{\ensuremath{\boldsymbol{s}} }

\newcommand{\wv}[0]{\ensuremath{\boldsymbol{w}} }
\newcommand{\xv}[0]{\ensuremath{\boldsymbol{x}} }
\newcommand{\yv}[0]{\ensuremath{\boldsymbol{y}} }
\newcommand{\zv}[0]{\ensuremath{\boldsymbol{z}} }

\newcommand{\Pimat}[0]{\ensuremath{\boldsymbol{\Pi}} }

\newcommand{\Phimat}[0]{\ensuremath{\boldsymbol{\Phi}}}

\newcommand{\Omegamat}[0]{\ensuremath{\boldsymbol{\Omega}}}

\newcommand{\thetav}[0]{\ensuremath{\boldsymbol{\theta}} }

\newcommand{\lambdav}[0]{\ensuremath{\boldsymbol{\lambda}} }
\newcommand{\muv}[0]{\ensuremath{\boldsymbol{\mu}} }

\newcommand{\rhov}[0]{\ensuremath{\boldsymbol{\rho}} }
\newcommand{\sigmav}[0]{\ensuremath{\boldsymbol{\sigma}} }

\newcommand{\phiv}[0]{\ensuremath{\boldsymbol{\phi}} }

\newcommand{\varphiv}[0]{\ensuremath{\boldsymbol{\varphi}} }

\newcommand{\cdotv}[0]{\ensuremath{\boldsymbol{\cdot}}}
\newcommand{\zerov}[0]{\ensuremath{\boldsymbol{0}} }
\newcommand{\onev}[0]{\ensuremath{\boldsymbol{1}} }

\newcommand{\mc}{\multicolumn}
\newcommand{\mr}{\multirow}

\begin{document}
\title{Deep %
	Autoencoding Topic Model with \\ Scalable Hybrid Bayesian Inference}

\author{Hao~Zhang,
	Bo~Chen,~\IEEEmembership{Senior member,~IEEE,}
	Yulai~Cong,
	Dandan Guo,
	Hongwei~Liu,~\IEEEmembership{Member,~IEEE,}
	and Mingyuan~Zhou
	\thanks{B. Chen acknowledges the support of the Program for Young Thousand Talent by Chinese Central Government, the 111 Project (No. B18039), and NSFC (61771361)}
	\thanks{H. Liu acknowledges the support of NSFC for Distinguished Young Scholars (61525105) and Shaanxi Innovation Team Project.}
	
	\thanks{M. Zhou acknowledges the support of Award IIS-1812699 from the U.S. National Science Foundation.}
	
	\thanks{H. Zhang, B. Chen, Yu. Cong, D. Guo and H. Liu are with National Lab of Radar Signal Processing, Collaborative Innovation Center of Information Sensing and Understanding, Xidian University, Xi'an, Shaanxi 710071, China.}
	\thanks{M. Zhou is with McCombs School of Business, The University of Texas at Austin, Austin, TX 78712, USA.}
	
	\thanks{Corresponding author: Bo Chen, bchen@mail.xidian.edu.cn.}
	
	\thanks{E-mail: zhanghao\_xidian@163.com, bchen@mail.xidian.edu.cn, yulaicong@gmail.com, gdd\_xidian@126.com, hwliu@xidian.edu.cn, mingyuan.zhou@mccombs.utexas.edu.}%
}

\markboth{IEEE Transactions on Pattern Analysis and Machine Intelligence}%
{Shell \MakeLowercase{\textit{et al.}}: Deep Autoencoding Topic Model}
\maketitle

\begin{abstract}
	To build a flexible and interpretable model for document analysis, we develop deep autoencoding topic model (DATM) that uses a hierarchy of gamma distributions to construct its multi-stochastic-layer generative network. In order to provide scalable posterior inference for the parameters of the generative network, we develop topic-layer-adaptive stochastic gradient Riemannian MCMC that jointly learns simplex-constrained global parameters across all layers and topics, with topic and layer specific learning rates. Given a posterior sample of the global parameters, in order to efficiently infer the local latent representations of a document under DATM across all stochastic layers, we propose a Weibull upward-downward variational encoder that deterministically propagates information upward  via a deep neural network, followed by a Weibull distribution based stochastic downward generative model. To jointly model documents and their associated labels, we further propose supervised DATM that  enhances the discriminative power of its latent representations.  The efficacy and scalability of our models are demonstrated on both unsupervised and supervised learning tasks on big corpora.
\end{abstract}


\section{Introduction}

\IEEEPARstart{T}o analyze a collection of documents with high-dimensional, spare, and over-dispersed bag-of-words representation, a common task is to perform topic modeling that extracts topics and topic proportions in an unsupervised manner, and another common task is to perform supervised topic modeling that jointly models the documents and their labels. While  latent Dirichlet allocation (LDA) \cite{blei2003latent} and a variety of its extensions, such as these for capturing correlation structure \cite{lafferty2006correlated}, inferring the number of topics \cite{HDP,NBP2012}, document categorization \cite{zhu2012medlda}, multimodal learning \cite{srivastava2012multimodal,wang2018MPGBN}, collaborative filtering \cite{wang2011collaborative}, and scalable inference \cite{hoffman2013stochastic}, have been widely used for document analysis, the representation power of these shallow probabilistic generative models is constrained by having only a single stochastic hidden layer.

{\subsection{Related work}

\subsubsection{Deep generative models for documents}
 
 To address the constraint of a shallow generative model, there is a surge of research interest in multilayer representation learning for documents.
 To analyze the term-document count matrix of a text corpus, Srivastava et al. \cite{srivastava2013modeling} extend a deep Boltzmann machine (DBM) with the replicated softmax topic model \cite{hinton2009replicated} to infer a multilayer representation with  binary hidden units, but its inference network is not trained to match the true posterior \cite{mnih2014neural2} and the higher-layer neurons learned by DBM are difficult to visualize.
 Deep Poisson factor analysis (DPFA)  \cite{gan2015scalable} is introduced to generalize Poisson factor analysis \cite{zhou2012beta}, with a deep structure restricted to model binary topic usage patterns.
Extending LDA, a hierarchical LDA  (hLDA)  is developed based on the nested Chinese restaurant process
\cite{griffiths2004hierarchical}, in which the topics are arranged in an $L$-level tree and a document draws its words from a mixture of $L$ topics within a document-specific root-to-leaf path.
 Deep exponential families (DEF) \cite{def} construct more general probabilistic deep networks with non-binary hidden units, in which a count matrix can be factorized under the Poisson likelihood, with the gamma distributed hidden units of adjacent layers linked via the gamma scale parameters. 
 The Poisson gamma belief network (PGBN) \cite{zhou2015poisson,GBN} also factorizes a count matrix under the Poisson likelihood, but factorizes the shape parameters of the gamma distributed hidden units of each layer into the product of a connection weight matrix and the gamma hidden units of the next layer, resulting in strong nonlinearity and readily interpretable multilayer latent representations. %
However, the inference of PGBN in Zhou et al. \cite{zhou2015poisson} is based on Gibbs sampling,  making it hard to be applied to big corpora, slow in out-of-sample prediction, and difficult to be jointly trained with a downstream task.

\subsubsection{Scalable inference}

These multilayer probabilistic models  are often characterized by a top-down generative structure, with the distribution of a hidden layer typically acting as a prior for  the layer below. In order to perform scalable inference for big corpora, both stochastic gradient Markov chain Monte Carlo (SG-MCMC) \cite{ma2015a, patterson2013stochastic, welling2011bayesian, cong2017deep}  and stochastic variational inference (SVI) \cite{hoffman2013stochastic, foulds2013stochastic, ranganath2013black, mnih2014neural2} have been developed for topic models. Despite being able to infer a multilayer representation of a text corpus,
they usually rely on an iterative procedure to infer the latent representation of a new document at the testing stage, regardless of whether variational inference or MCMC is used. The potential need of a large number of iterations per testing document  makes them unattractive when real-time processing is desired.
For example, one may need to rapidly extract the topic-proportion vector of a document and use it for downstream analysis, such as identifying key topics and retrieving related documents. In addition, they often make the restrictive mean-field assumption \cite{ranganath2013black}, require sophisticated variance reduction techniques \cite{mnih2014neural2}, and use a single learning rate for different variables across all layers \cite{welling2011bayesian, Chen2014Stochastic, ding2014bayesian}, making it difficult to generalize them to deep probabilistic models.

A potential solution is to construct a variational autoencoder (VAE) that learns the parameters of an inference network (recognition model or encoder) jointly with those of the generative model (decoder) \cite{kingma2014stochastic,rezende2014stochastic}. However, most  existing VAEs rely on Gaussian latent variables, with the neural networks (NNs) acting as nonlinear connections between adjacent layers \cite{sonderby2016ladder,dai2016variational,pixelVAE}. A primary reason is that there is a simple reparameterization trick for Gaussian latent variables that allows efficiently computing the noisy gradients of  the evidence  lower bound (ELBO)  with respect to the NN parameters.  Unfortunately, Gaussian based distributions often fail to well approximate the posterior distributions of sparse, nonnegative, and skewed  document  latent representations. For example, Srivastava et al. \cite{srivastava2017autoencoding} propose autoencoding variational inference for topic models (AVITM),  as shown in Fig. \ref{fig:AVITM-strucure}, which utilizes the logistic-normal distribution to approximate the posterior distribution of the latent representation of a document; even though the generative model  is LDA  \cite{blei2003latent}, a basic single-hidden-layer topic model, due to the insufficient ability of the logistic-normal distribution to model sparsity, AVITM has to rely on some heuristics to force the latent representation of a document to be sparse.
To overcome this limitation, Knowles \cite{knowles2015stochastic} introduces a reparameterization method for the gamma distribution that relies on inefficient numerical approximation; 
Ruiz et al. \cite{ruiz2016generalized} develop generalized reparameterization (Grep)  to extend the reparameterization gradient to a wider class of variational distributions utilizing invertible transformations, leading to transformed distributions that only weakly depend on the variational parameters; and Naesseth et al. \cite{RSVI} further improve Grep with rejection sampling variational inference (RSVI) that achieves lower variance and faster speed via a rejection sampling algorithm at the cost of introducing more random noisy.
Another common shortcoming of existing VAEs is that they often only provide a point estimate  for the global parameters of the generative model, and hence their inference network is optimized to approximate the posterior of the local parameters conditioning on the data and that point estimate, rather than a full posterior, of the global parameters. In addition, from a probabilistic modeling point of view,  the VAE inference network  is often merely a  shallow probabilistic model, whose parameters are deterministically nonlinearly transformed from the observations via a non-probabilistic deep NN.

\subsection{Motivations and contributions}

To address the aforementioned constrains of existing %
topic models and move beyond Gaussian latent variable based deep generative models and inference, we develop deep autoencoding topic model (DATM).
DATM uses a deep topic model  as its decoder and a deterministic-upward--stochastic-downward network as its encoder, and jointly trains them with a hybrid Bayesian inference, integrating both SG-MCMC \cite{welling2011bayesian,ma2015complete,cong2017deep} and a multilayer Weibull distribution based inference network.
The distinctions of DATM are summarized as follows.

\begin{itemize}[itemindent=7mm, leftmargin=0mm]

			\item %
			 DATM is related to a usual VAE in having both a decoder and encoder, but differs from it in a number of ways: 1) Deep latent Dirichlet allocation (DLDA), a probabilistic deep topic model equipped with a gamma belief network, acts as the generative model; 2) Inspired by the upward-downward Gibbs sampler of DLDA, as sketched in Fig. \ref{fig:DLDA-strucure},  the inference network of DATM uses an upward-downward structure, as shown in Fig. \ref{fig:DWVAE-strucure}, to combine a non-probabilistic bottom-up deep NN and a probabilistic top-down deep generative model, with the $\ell$th hidden layer of the generative model linked to both the $(\ell+1)$th hidden layer of itself and the $\ell$th hidden layer of the deep NN; 3) A hybrid of SG-MCMC and autoencoding variational inference is employed to infer both the posterior distribution of the global parameters, represented as collected posterior MCMC samples, and a VAE that approximates the posterior distribution of the local parameters given the data and a posterior sample (rather than a point estimate) of the global parameters; 4) We use the Weibull distributions in the inference network to approximate gamma distributed conditional posteriors, exploiting the facts that the Weibull and gamma distributions have similar probability density functions (PDFs), the Kullback--Leibler (KL) divergence from the Weibull to gamma distributions is analytic,  and a Weibull random variable can be reparameterized with uniform random noise.
			
			\item In probabilistic topic models, the document-specific topic proportions are commonly used as features for downstream analysis such as document classification.
			Although  the unsupervisedly extracted features can be used to train a classifier  \cite{blei2003latent,zhou2015poisson,GBN}, they provide  relatively poor discrimination power \cite{lee2016deep}. Although some discriminative models \cite{zhu2012medlda,mcauliffe2008supervised,lacostejulien2009disclda} are integrated into LDA to improve its discrimination power, their single-hidden-layer structure clearly limits their ultimate potential for satisfactorily representing high-dimensional and sparse document data.
			Exploiting the multi-layer structure of DATM, we further propose a supervised deep topic model, referred to as supervised DATM (sDATM), that combines the flexibility of DATM in describing the documents and the discriminative power of deep NNs under a principled probabilistic framework. Distinct from supervised LDA and its extensions \cite{mcauliffe2008supervised,lacostejulien2009disclda,zhu2012medlda}, the features at different layers of sDATM exhibit different statistical properties, and hence are combined together to boost their discriminative power.

			\item DATM provides %
			interpretable hierarchical topics, which vary from very specific to increasingly more general when moving towards deeper layers. In DATM, 
			 the number of topics in a layer, i.e., the width of that layer, is automatically learned from the data given a fixed budget on the width of the first layer, with the help of the gamma-negative binomial process %
			 and a greedy layer-wise training strategy \cite{NBP2012}.
		\end{itemize}

The remainder of the paper is organized as follows. Section II introduces a deep probabilistic autoencoder for topic modeling and develops a hybrid Bayesian inference algorithm to perform efficient scalable inference.
With label information, a supervised deep topic model is introduced in Section III. Section IV reports a series of experiments on document representation and classification to evaluate the proposed models.
Section~V concludes the paper. We note that parts of the work presented here first appeared in Cong et al. \cite{cong2017deep} and Zhang et al. \cite{Zhang2018WHAI}.  In this paper, we unify related materials in both conference  publications and provide expansion to supervised deep topic modeling.
Furthermore, to infer from the data rather than predetermining  the network structure, given a fixed budget on the width of the first layer, we combine Bayesian nonparametrics and greedy layer-wise training for DATM to learn the width of each added hidden layer, with all the added hidden layers jointly trained, leading to further improved performance.

\section{Deep Autoencoding Topic Model}
In what follows, we propose DATM %
that uses a deep hierarchical Bayesian model as the generative model (decoder), and a deterministic-upward--stochastic-downward network as the recognition model (encoder, inference network).

\subsection{Document decoder: deep latent Dirichlet allocation}\label{sec2A}
To capture a hierarchical document latent representation, DATM uses PGBN \cite{zhou2015poisson}, a deep probabilistic topic model, as the decoder. %
Choosing a deep generative model as its decoder distinguishes DATM from AVITM \cite{srivastava2017autoencoding}, which uses a ``shallow'' LDA as its decoder, and from the conventional VAE, which often employs a ``shallow'' (transformed) Gaussian distribution as its decoder with parameters deterministically and
nonlinearly transformed from the observation via ``black-box'' deep NNs.

To model  high-dimensional multivariate sparse count vectors $\xv_n \in \mathbb{Z}^{K_0}$, where $\mathbb{Z}=\{0,1,\ldots\}$, under the Poisson likelihood, %
the PGBN generative model  with $L$ hidden layers, from top to bottom, can be expressed as
 \begin{align}\label{PGBN_generate}
  &\thetav_n^{(L)} \sim \mbox{Gam}\left(\rv,1/c_n^{(L+1)} \right), \rv \sim \mbox{Gam} (\gamma_0/K_L,1/c_0)%
  \nonumber \\
  &\thetav_n^{(l)} \sim \mbox{Gam}\left(\Phimat^{(l+1)} \thetav_n^{(l+1)} ,1/c_n^{(l+1)} \right), \quad l=1, \cdots, L-1 \nonumber\\
  & \xv_n \!\sim \!\mbox{Pois} \left(\Phimat^{(1)} \thetav_n^{(1)} \right),\!
\end{align}
where the hidden units (topic weights) $\thetav_n^{(l)} \in \mathbb{R}_+^{K_l}$  of layer $l$ are factorized into the product of the factor loading $\Phimat^{(l+1)} \in \mathbb{R}_+^{K_{l} \times K_{l+1}}$ and hidden units of layer $l+1$ under the gamma distribution.
For scale identifiability and ease of inference and interpretation, PGBN further places a simplex constraint on %
each column of $\{ \Phimat^{(l)} \}_{l=1}^L$ via %
a Dirichlet prior as
$\Phimat_k^{(l)} \sim \mbox{Dir} \left( \eta^{(l)} \Imat_{K_{l-1}}\right)$, where $ \Imat_{K_{l-1}}$ is a vector of $K_{l-1}$ ones.
The gamma shape parameters $\rv = (r_1,\cdots,r_{K_L})^T$ at the top layer are shared across all $\xv_n$ and $\{ 1/c_n^{(l)} \}_{l=2}^{L+1}$ are gamma scale parameters.

Using the law of total expectation, we have
\begin{equation}\label{Phi_notation}
\mathbb{E}\left[ \xv_n \,\Big |\, \thetav_n^{(l)}, \left\{\Phimat^{(t)} ,c_n^{(t)} \right\}_{t=1}^l \right] = \left[ \prod_{t=1}^{l} \Phimat^{(t)} \right] \frac{\thetav_n^{(l)}}{\prod_{t=2}^{l}c_n^{(t)}},
\end{equation}
which means the conditional expectation of $\xv_n$ on layer $l$ is a linear combination of the columns in $ \prod_{t=1}^{l} \Phimat^{(t)} $, with $\thetav_n^{(l)}$ viewed as a document-dependent topic-weight vector that can be used for downstream analysis, such as document classification and retrieval.
Furthermore, $\prod_{t=1}^{l-1} \Phimat^{(t)} \phi_k^{(l)}$ can be viewed as the projection of topic $\phi_k^{(l)}$ to the bottom data layer, which can be  used to visualize the topics at different layers.
An example of the hierarchical topic structure is illustrated in Fig.~\ref{fig:topic_wiki}.
The inferred topics of this model %
tend to be very specific at the bottom layer and become increasingly more general when moving upwards (deeper).

Denote $q_n^{(l+1)} = \log (1+q_n^{(l)}/c_n^{(l+1)})$ for $l=1,\cdots,L$, where $q_n^{(1)}=1$, $p_j^{(l)}=1-e^{-q_j^{(l)}}$, and $m\sim \mbox{SumLog}(x,p)$ as the sum-logarithmic distribution \cite{NBP_CountMatrix}.
With all the gamma distributed hidden units marginalized out, %
PGBN can also  be
represented as deep LDA (DLDA) \cite{cong2017deep}, expressed as
\begin{align}\label{DLDA-another expression}
& x_{kn}^{(L+1)}\sim \mbox{Pois}(r_kq_n^{(L+1)}),~k=1,\ldots,K_L,\notag\\
&m_{kn}^{(L)(L+1)} \sim \mbox{SumLog} (x_{kn}^{(L+1)},p_n^{(L+1)}),~k=1,\ldots,K_L, \nonumber \\
& \quad \quad \quad \quad \quad \quad \quad \quad \quad \cdots \nonumber \\
& \left( x_{vkn}^{(l)} \right)_{v=1,K_{l-1}}\! \sim \mbox{Mult}\left( m_{kn}^{(l)(l+1)}, \phiv_k^{(l)} \right),~k=1,\ldots,K_l, \nonumber \\
& x_{kn}^{(l)}=\sum_{k'=1}^{K_l} x_{kk'n}^{(l)}, ~~k=1,\ldots,K_{l-1}, \nonumber \\
& m_{kn}^{(l-1)(l)} \sim \mbox{SumLog} \left( x_{kn}^{(l)}, p_j^{(l)} \right), ~~k=1,\ldots,K_{l-1},%
\notag\\
& \quad \quad \quad \quad \quad \quad \quad \quad \quad \cdots \nonumber \\
&  \left( x_{vkn}^{(1)} \right)_{v=1,K_0}\!\! \sim \mbox{Mult}\left( m_{kn}^{(1)(2)}, \phiv_k^{(1)} \right),~k=1,\ldots,K_1\notag\\
&x_{kn}^{(1)}=\sum_{k'=1}^{K_1} x_{kk'n}^{(1)},~k=1,\ldots,K_0.
\end{align}
For simplicity, below we use DLDA to refer to both the PGBN and DLDA representations of the same underlying deep generative model.
Note the single-hidden-layer version of DLDA reduces to Poisson factor analysis \cite{zhou2012beta}, which is closely related to LDA.
To make DLDA be scalable to big corpora, in the following, we develop a SG-MCMC based algorithm.

{\bf{SG-MCMC.}}
For a statistical model with likelihood $p(\xv\,|\, \zv)$ and prior $p(\zv)$, where $\zv$ denotes the set of all global variables, one may follow the general framework for
SG-MCMC \cite{ma2015a}
to express the sampling equation as
\begin{align}\label{SGMCMC}
  \zv_{t+1} =\zv_t \,+ & \,\epsilon_t \left\{ -\left[ \Dmat (\zv_t) + \Qmat (\zv_t) \right]\nabla H(\zv_t)+\Gamma(\zv_t) \right\} \nonumber \\
  & + \mathcal{N} \left( \zerov, \epsilon_t \left[ 2 \Dmat (\zv_t) - \epsilon_t \Bmat_t \right]  \right),
\end{align}
where %
$\epsilon_t$ denotes the step size at step $t$,
$H(\zv)=-\ln p(\zv) - \rho \Sigma _ {\xv \in \hat{\Xmat}} \ln p(\xv | \zv)$,
$\Gamma_i(\zv_t) = \sum_j \frac{\partial}{\partial z_{jt}} [\Dmat_{ij}(\zv_t) + \Qmat_{ij}(\zv_t)] $,
$\hat{\Xmat}$ the mini-batch, $\rho $ the ratio of the dataset size $|\Xmat|$ to the mini-batch size $|\hat{\Xmat}|$, and $\Bmat_t$ an estimate of the stochastic gradient noise variance satisfying a positive definite constraint as $2\Dmat(\zv_t)-\epsilon_t \Bmat_t \succ \zerov$.
Under this framework, stochastic gradient Riemannian Langevin dynamics (SGRLD) \cite{patterson2013stochastic} is proposed for LDA, with $\Dmat(\zv) = \Gmat(\zv)^{-1}$, $\Qmat(\zv)=\zerov$, and $\Bmat_t=\zerov$, where
\begin{equation}\label{FIM}
  \Gmat(\zv) = \mathbb{E}_{{\Pimat}|\zv} \left[ -\frac{\partial^{2}}{\partial \zv^2} \ln p({\Pimat}|\zv) \right]
\end{equation}
 is the Fisher information matrix (FIM) that is widely used to precondition the gradients to adjust the learning rates,
where $\Pimat$ denotes the set of all observed and local variables.
In general, it is difficult to calculate the FIM. %
Fortunately,
DLDA admits a block-diagonal FIM that is easy to work with.

{\bf{Fisher information matrix.}} %
Since the topic $\phiv_k^{(l)} \in \mathbb{R}_{+}^{K_{l-1}}$ is a vector that lies on the probabilistic simplex, we use $\phiv_k^{(l)}=\left( \varphi^{(l)}_{1k}, \cdots, \varphi^{(l)}_{(K_{l-1}-1)k},1-\Sigma_{v<K_{l-1}}\varphi^{(l)}_{vk} \right)^T$ as the reduced-mean parameterization of $\phiv_k^{(l)}$, which is different from the expanded-mean parameterization used by  SGRLD  \cite{patterson2013stochastic}.
Under the DLDA representation shown in \eqref{DLDA-another expression}, %
the likelihood is fully factorized with respect to the global parameters $\zv=\{\varphiv_1^{(1)},\ldots,\varphiv_{K_L}^{(L)},\rv\}$, leading to a FIM that admits  a block diagonal form as
\begin{align}\label{blockFIM}
&  ~~~~~ \Gmat(\zv)=\mbox{diag} \left[ \Imat \left( \varphiv_1^{(1)} \right), \cdot \cdot \cdot, \Imat \left( \varphiv_{K_L}^{(L)} \right),
\Imat \left( \rv \right)  \right],\\
\label{FIM}
&  \Imat\left( \varphiv_k^{(l)} \right)=%
  M_k^{(l)}\left[ \mbox{diag}\left( 1/\varphiv_k^{(l)} \right) +\onev \onev^T / (1-\varphiv_{\cdotv k}^{(l)}) \right],\\
& ~~~~~~~~~~\Imat \left( \rv \right)= M^{(L+1)} \mbox{diag} \left( r_1^{-1},\ldots,r_{K_L}^{-1} \right),
\end{align}
where the symbol ``$\cdotv$'' denotes summing over the corresponding index, $M_l^{(l)} = \mathbb{E}\left[ m_{k \cdotv}^{(l)(l+1)} \right] = \mathbb{E}\left[ x_{\cdotv k \cdotv }^{(l)} \right]$, and  $M^{(L+1)} = \mathbb{E}\left[ q_{\cdotv}^{(L+1)} \right]$.
Note that the block diagonal structure of the FIM for DLDA makes it computationally appealing to apply its inverse for preconditioning.
To utilize \eqref{SGMCMC} to develop SG-MCMC for DLDA under the reduced-mean parameterization, similar to SGRLD, we let
$\Dmat(\zv) = \Gmat(\zv)^{-1}$, $\Qmat(\zv)=\zerov$, and $\Bmat_t=\zerov$.
Relying on the FIM to automatically adjust relative learning rates for different parameters across all layers and topics, we only need to choose a single step size $\varepsilon _t$ for all.
Moreover, the block-diagnoal structure of FIM will be carried over to its inverse $\Dmat(\zv)$, leading to a computationally efficient way to perform updating by \eqref{SGMCMC}, as described below.

\begin{figure*}
  \centering
  \subfloat[DATM]{\includegraphics[scale=0.65]{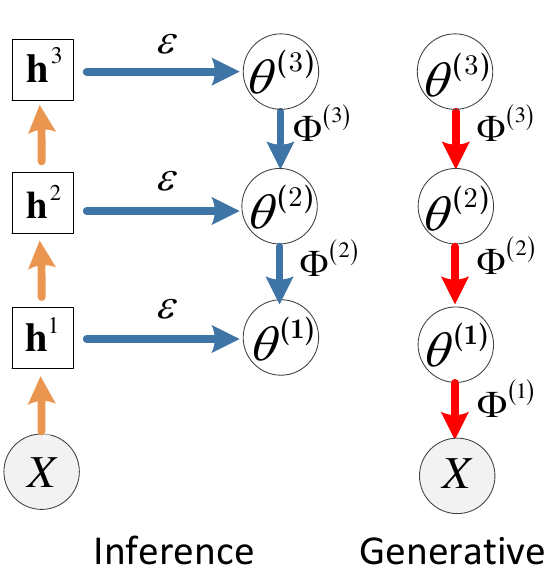}\label{fig:DWVAE-strucure} } $\quad$ $\quad$ $\quad$
  \subfloat[AVITM]{\includegraphics[scale=0.65]{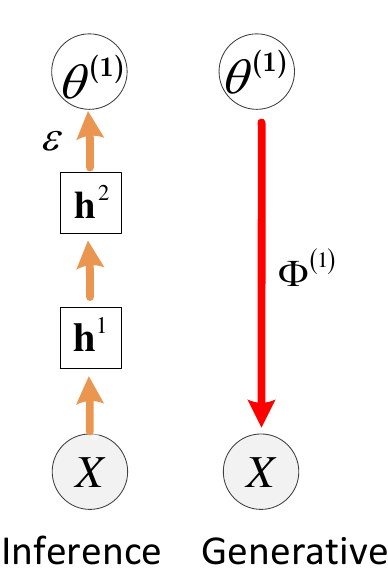}\label{fig:AVITM-strucure}} $\quad$ $\quad$ $\quad$
  \subfloat[DLDA]{\includegraphics[scale=0.65]{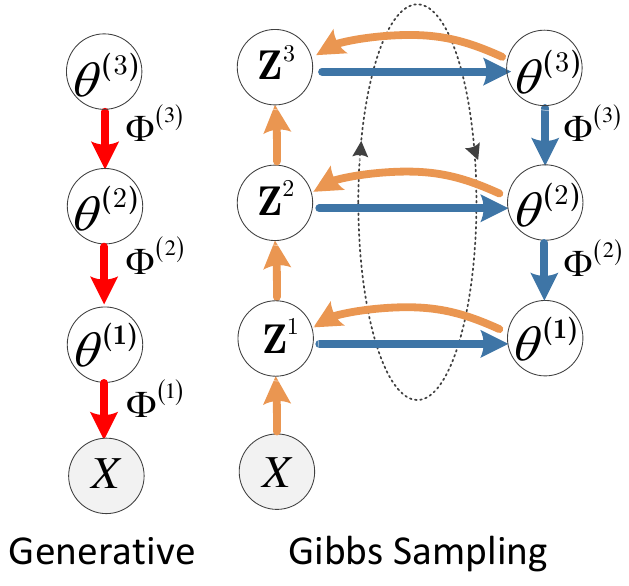}\label{fig:DLDA-strucure}} $\quad$
  $\quad$ $\quad$
  \subfloat[sDPATM]{\includegraphics[scale=0.65]{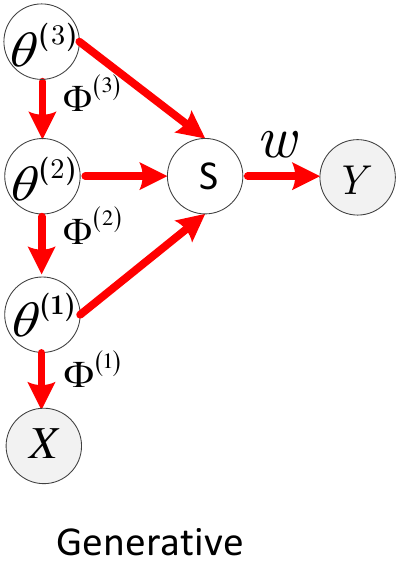}\label{fig:SDLDA-strucure}}
  \caption{(a-b): Inference (or encoder/recognition) and generative (or decoder) models for (a) DATM and (b) AVITM; (c) the generative model and a sketch of the upward-downward Gibbs sampler of DLDA, where $\Zmat^l$ are augmented latent counts that are upward sampled in each Gibbs sampling iteration. Circles are stochastic variables and squares are deterministic variables. The orange and blue arrows denote the upward and downward information propagation respectively, and the red ones denote the data generation; (d) the generative model of sDPATM.}
  \label{fig:network-structure}
\end{figure*}

{\bf{Inference on the probability simplex.}}
Using the DLDA representation in \eqref{DLDA-another expression} and reduced-mean parameterization of simplex-constrained vectors,
we derive a block-diagonal FIM as in \eqref{FIM}.
Besides this advantage, we describe another reason for our choice in the following discussion,
where we ignore the layer-index superscript $^{(l)}$ for brevity and assume $\phiv_k=(\varphiv_k^T, 1-\varphi_{\cdotv k})^T\in\mathbb{R}_+^{V}$ lies on a $V$-dimensional simplex. %

With \eqref{DLDA-another expression} and the Dirichlet-multinomial conjugacy, taking the gradient with respect to $\varphiv_k$ of the summation of the  log-likelihood of a mini-batch $\tilde{X}$ scaled by $\rho = |X|/|\tilde{X}|$ and the logarithm of the Dirichlet prior, we have
\begin{equation}\label{gradient_H}
  \nabla_{\varphiv_k} \left[ -H(\varphiv_k) \right] = \frac{\rho \bar{\xv}_{:k\cdotv}+\eta-1}{\varphiv_k} - \frac{\rho \tilde{x}_{vk\cdotv}+\eta-1}{1-\varphi_{\cdotv k}},
\end{equation}
where $\tilde{x}_{vk\cdotv}=\sum_{n:x_n \in \tilde{X} }x_{vkn}$ , $\bar{\xv}_{:k\cdotv} = \left( \tilde{x}_{1k\cdotv}, \cdots,  \tilde{x}_{(V-1)k\cdotv} \right)^T$.
Note the gradient in \eqref{gradient_H} becomes unstable when some components of $\varphiv_k$ approach zeros, a key reason that this approach is mentioned but considered as an unsound choice in Patterson \& Teh \cite{patterson2013stochastic}.
However, after preconditioning the noisy gradient in \eqref{gradient_H} with the inverse of the FIM,
it is intriguing to find out that the stability issue completely disappears.
More specifically, by substituting \eqref{FIM},  \eqref{gradient_H}, and $\Gamma (\varphiv_k)$, whose derivation is given in the Supplement, into the SG-MCMC update equation \eqref{SGMCMC}, the sampling of $\varphiv_k$ becomes
\begin{align}\label{Phi-updata-tmp}
&\left( {\varphiv_k} \right)_{t + 1} \! =  \! \bigg[ \! \left( {\varphiv_k} \right)_t \! + \! \frac{\varepsilon _t}{M_k} \! \left[ \left(\rho \tilde \xv_{:k\cdotv} \! + \! \eta\right) \! - \! \left(\rho \tilde x_{\cdotv k\cdotv} \! + \! \eta V\right) \! \left( {\varphiv_k} \right)_t \right] \nonumber \\
&~~~~~ +  \mathcal{N} \left( 0, \frac{2 \varepsilon _t}{M_k}\left[ \mbox{diag}(\varphiv_k)_t - (\varphiv_k)_t (\varphiv_k)_t^T  \right] \right) \bigg]_\triangle,
\end{align}
where $[\cdot]_{\triangle}$ represents the constraint that $\varphi_{vk}\geq 0 $ and $ \varphi_{\cdotv k}=\sum_{v=1}^{V-1} \varphi_{vk} \leq 1$.

Note while the stability issue has now been solved, naively sampling from the  multivariate normal (MVN) distribution in \eqref{Phi-updata-tmp}, even without the $[\cdot]_{\triangle}$ constraint, is computationally expensive as a Cholesky decomposition of the non-diagonal covariance matrix has $O((V-1)^3)$ complexity \cite{Golub1996Matrix}. %
Fortunately, following Theorem 2 in Cong et al.  \cite{cong2017fast},  %
 we may equivalently draw
 $\phiv_k$  from a $V$-dimensional MVN that has a diagonal covariance matrix and is subject to the simplex constraint,
 with $O(V)$ complexity, as
\begin{align}\label{Phi-updata}
\left( {\phiv_k} \right)_{t + 1} \! = & \! \bigg[ \! \left( {\phiv_k} \right)_t \! + \! \frac{\varepsilon _t}{M_k} \! \left[ \left(\rho \tilde \xv_{:k\cdotv} \! + \! \eta\right) \! - \! \left(\rho \tilde x_{\cdotv k\cdotv} \! + \! \eta V\right) \! \left( {\phiv_k} \right)_t \right] \nonumber \\
& +  \mathcal{N} \left( 0, \frac{2 \varepsilon _t}{M_k} \mbox{diag} \left( \phiv_k \right)_t \right)\bigg]_{\angle},
\end{align}
where $[\cdot]_\angle$ denotes a simplex constraint that $\phi_{vk} \geq 0$ and $\Sigma_{v=1}^{V} \phi_{vk} = 1$. More details about \eqref{Phi-updata-tmp} and \eqref{Phi-updata} can be found in Examples 1-3 of Cong et al.\cite{cong2017fast}.

Similarly, with the gamma-Poisson conjugacy for $\rv$, we have $\Gamma_k (\rv) = 1/M^{(L+1)}$, whose detailed derivation is deferred to the Supplement, and hence the update of $\rv$ as
\begin{align}\label{r_update}
&  \rv_{t+1}= \bigg| \rv_t+\frac{\varepsilon _t}{M^{(L+1)}} \bigg[ \left( \rho \tilde{\xv}^{(L+1)}_{:\cdotv} + \frac{\gamma_0}{K_L} \right) \nonumber \\
  &~-\rv_t \left( c_0 + \rho \tilde{q}^{(L+1)}_{\cdotv} \right) \bigg]
  + \mathcal{N}\left( \zerov, \frac{2 \varepsilon _t}{M^{(L+1)}} \mbox{diag} (\rv_t) \right) \bigg|.
\end{align}

{\bf{Topic-layer-adaptive step-size.}}
Note that $\left\{ M_k^{(l)} \right\}_{l=1}^{L}$ and $M^{(L+1)}$ %
appearing in \eqref{Phi-updata} and \eqref{r_update} are expectations over all local variables that need to be approximately calculated. We update them using annealed weighting \cite{polatkan2015a} as
\begin{align}\label{update_M}
  M_k^{(l)} &= \left( 1-\varepsilon_t^{'} \right) M_k^{(l)} + \varepsilon_t^{'} \rho \mathbb{E}\left[ x_{\cdotv k \cdotv}^{(l)} \right],\\
  M_k^{(L+1)} &= \left( 1-\varepsilon_t^{'} \right) M_k^{(L+1)} + \varepsilon_t^{'} \rho \mathbb{E}\left[ \tilde{q}_{\cdotv}^{(L+1)} \right],
\end{align}
where the expectation $\mathbb{E}[\cdot]$ denotes averaging over the collected MCMC samples.
In DATM to be discussed below, this can be approximated by one sample from the introduced document encoder \eqref{posterior-completely}, instead of collected MCMC samples. %
For simplicity, we set $\varepsilon_t^{'}=\varepsilon_t$, which is found to work well in practice.
Note although it appears that a common step-size $\varepsilon_t$ is set for all layers, the effective step-sizes are $\varepsilon_t / M_k^{(l)}$, which differ for all layers ($l \in \{1,\cdot,\cdot,\cdot L\}$) and topics ($k \in \{1,\cdot,\cdot,\cdot K_l\}$).
For this reason,
We refer to the proposed SG-MCMC as topic-layer-adaptive stochastic gradient Riemannian (TLASGR) MCMC.

Despite having attractive properties and scalable inference via TLASGR-MCMC, the power of DLDA is limited in that %
it has to take a potentially large number of MCMC iterations to infer the latent representation of a test observation, and it is difficult to utilize  available side information such as class labels. %
To address these issues, in the following we first develop a deep document encoder network.

\begin{figure*}
  \centering
  \subfloat[$\mbox{KL}=0.96$]{\includegraphics[scale=0.205]{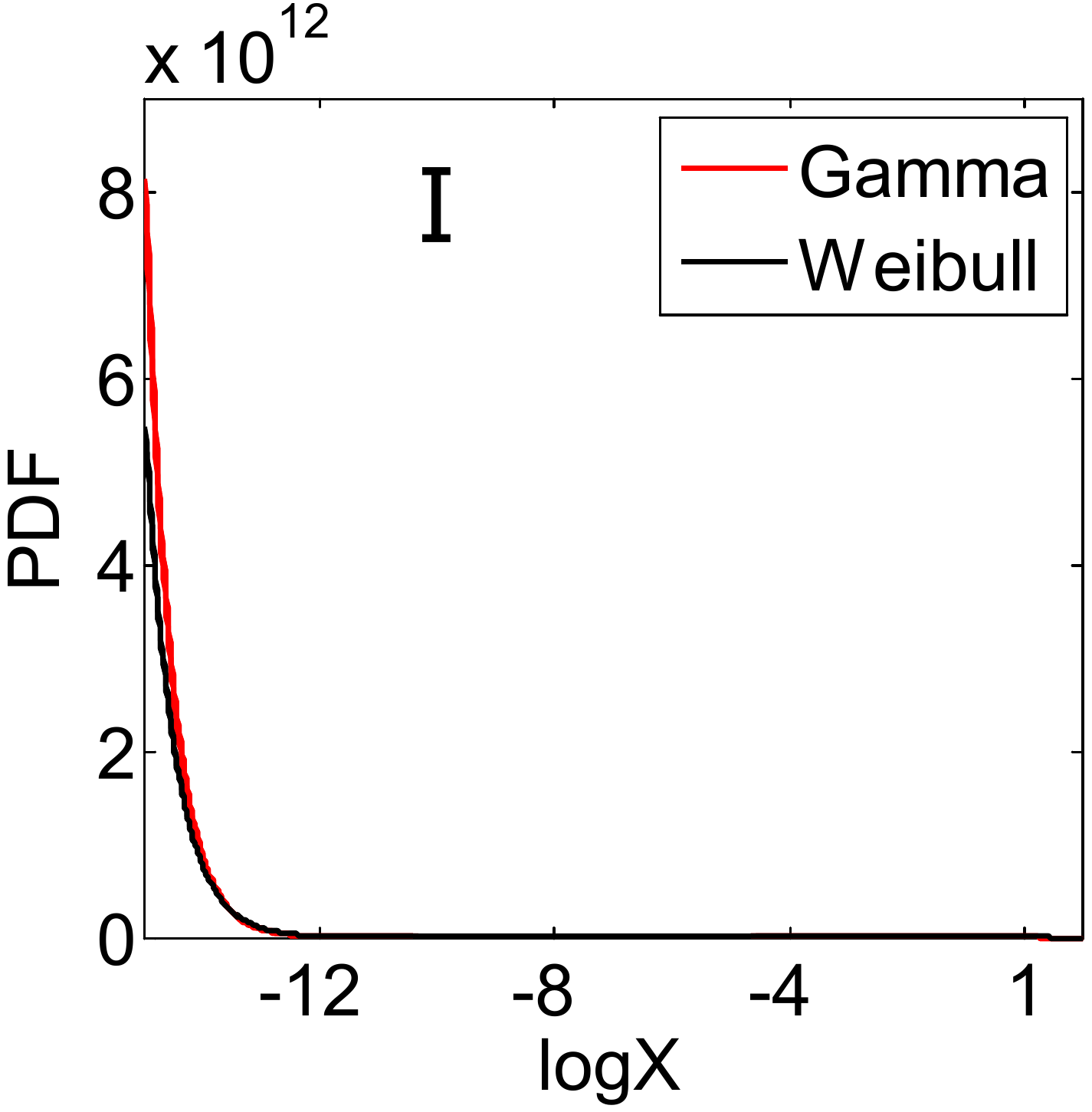}}
  \subfloat[$\mbox{KL}=0.01$]{\includegraphics[scale=0.2]{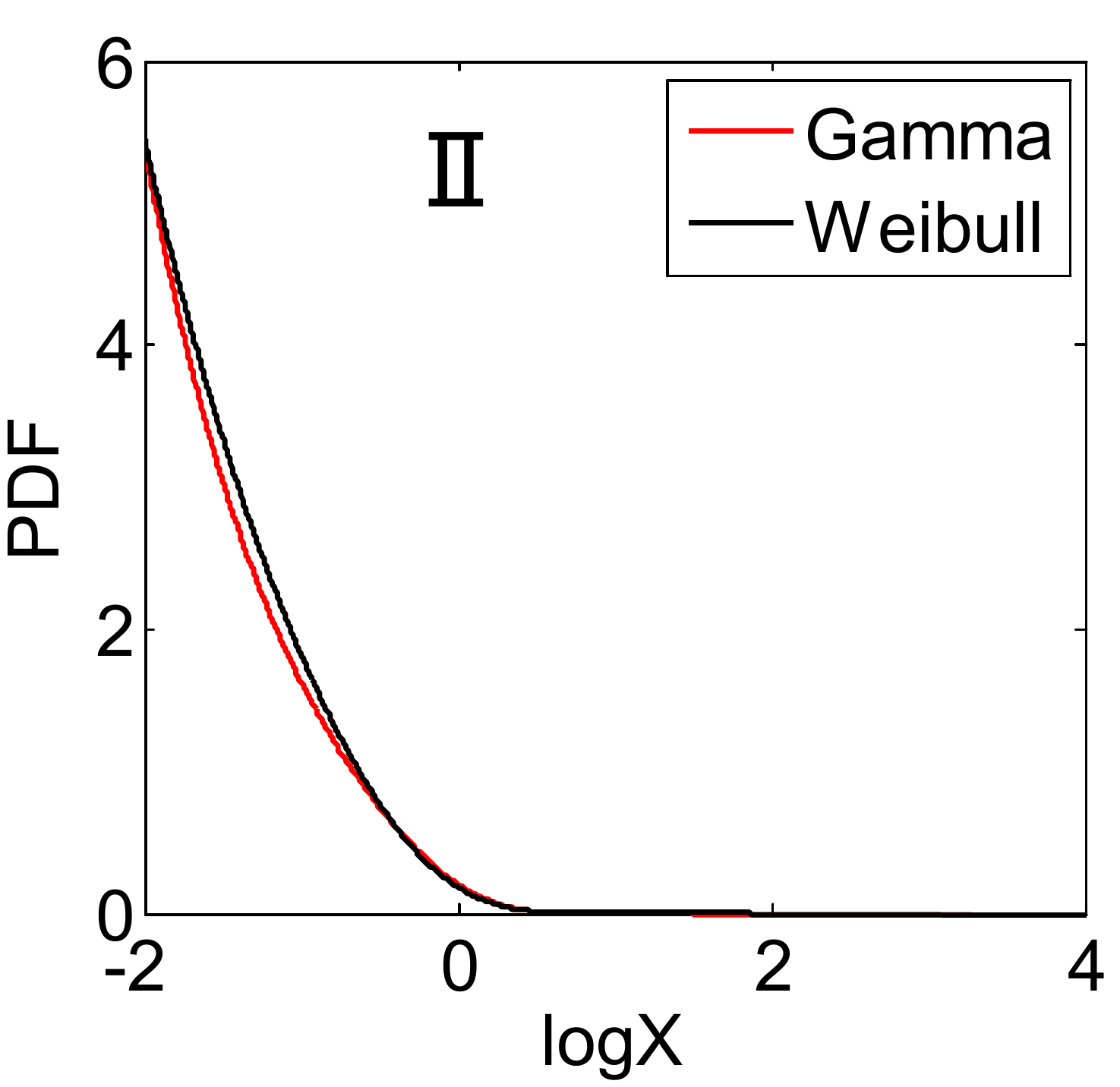}}
  \subfloat[$\mbox{KL}=0.06$]{\includegraphics[scale=0.2]{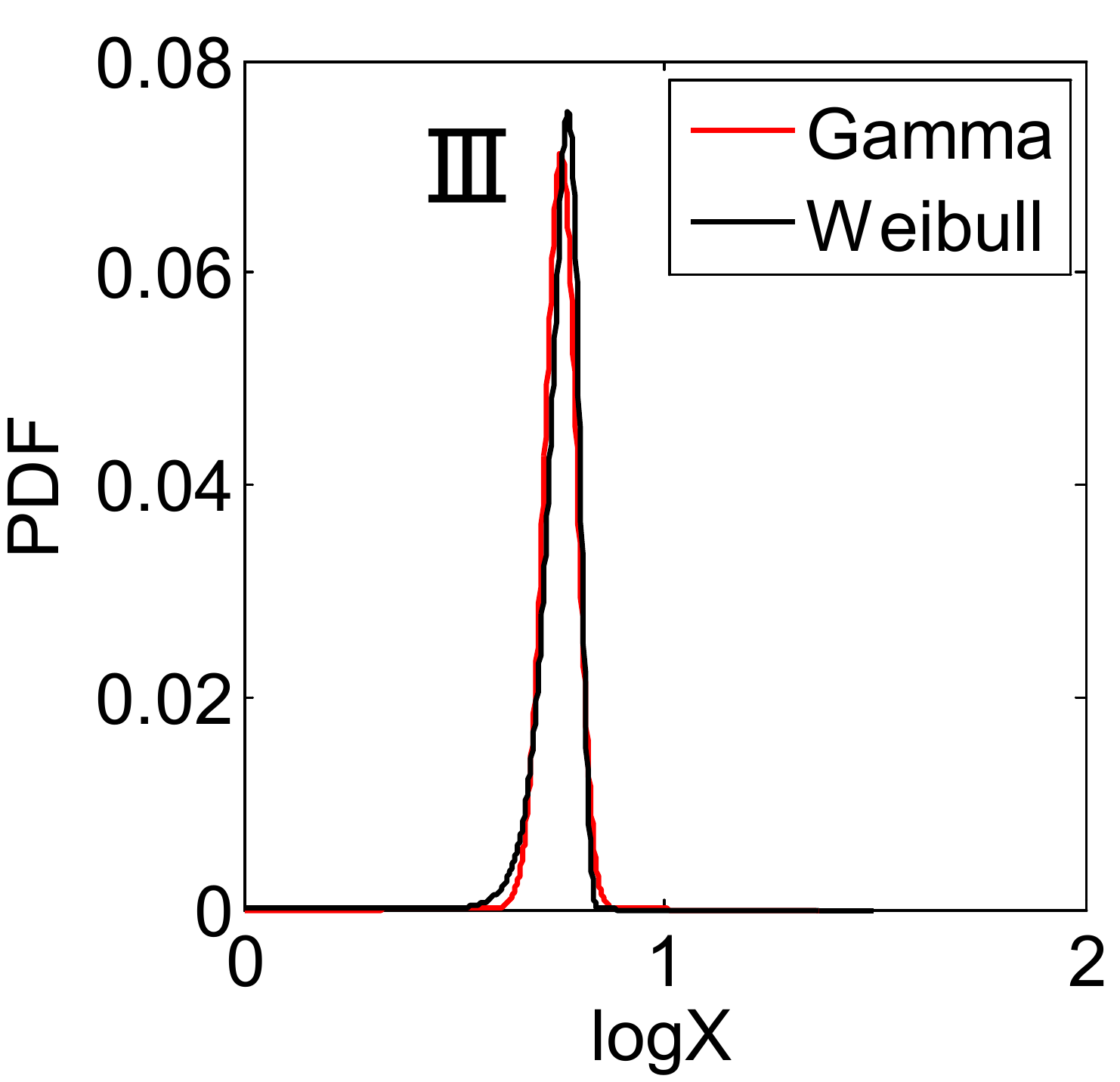}}
  \subfloat[$\mbox{KL}$ change]{\includegraphics[scale=0.2]{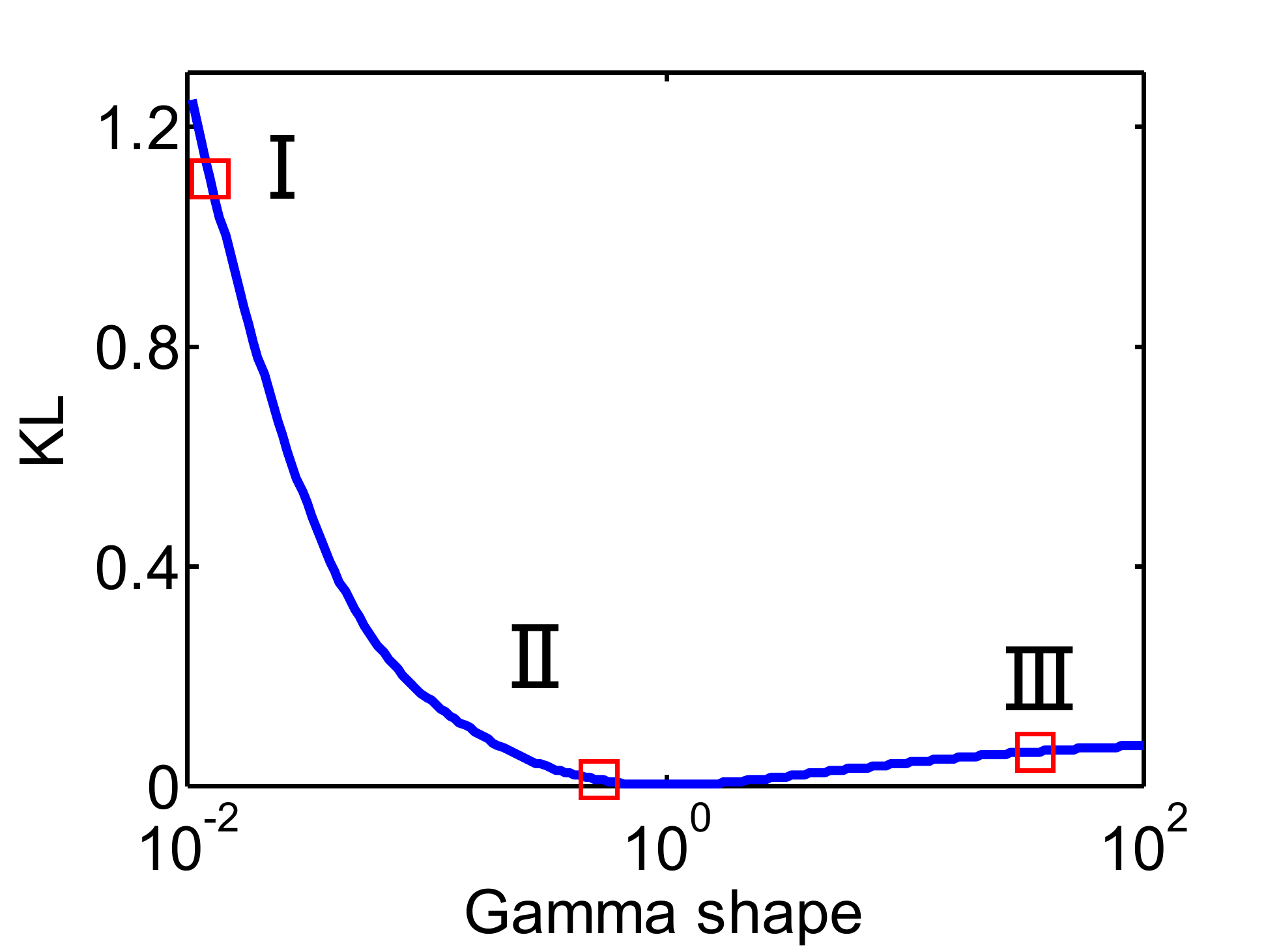}}
  \vspace{-2mm}
  \caption{The KL divergence from the inferred Weibull distribution to the target gamma one as %
  (a) $\mbox{Gamma}(0.05,1)$, (b) $\mbox{Gamma}(0.5,1)$, and (c) $\mbox{Gamma}(5,1)$. Subplot (d) shows the KL divergence as a function of the gamma shape parameter, where  the gamma scale parameter is fixed at 1. %
  }\label{fig:gamma-weibull-kl}
\end{figure*}

\subsection{Document encoder: Weibull upward-downward variational encoder} \label{sec:dist}

To perform fast inference for out-of-sample predictions, we are motivated to construct an inference network that maps the observations directly to the posterior distributions of their latent representations and hence avoid performing any iterative updates at the test time. Variational auto-encoder  (VAE)
\cite{kingma2014stochastic,rezende2014stochastic} becomes an idea candidate for this purpose. %
However, its success so far is mostly restricted to Gaussian distributed latent variables, and does
not generalize well to model sparse, nonnegative, and skewed latent document representations.
To this end, below we propose Weibull upward-downward variational encoder (WUDVE) to efficiently produce a document's multi-layer latent representation under DLDA.

To maximize the marginal likelihood $p(\xv)$ under DLDA, one may choose a usual strategy of variational Bayes \cite{jordan1999introduction} to maximize the ELBO of $p(\xv)$ that can be expressed as
\begin{align}\label{ELBO-of-VADTM}
L = &\sum \limits_{n=1}^N \mathbb{E} \left[ \ln p\left( \xv_n \,|\, \Phimat^{(1)}, \thetav_n^{(1)} \right) \right] \nonumber \\
&- \sum \limits_{n=1}^N \sum \limits_{l=1}^L \mathbb{E} \left[ \ln \frac{q\left( \thetav_n^{(l)}\,|\,  \Phimat^{(l+1)}, \thetav_n^{(l+1)}\right)}{p \left( \thetav_n^{(l)} \,|\, \Phimat^{(l+1)}, \thetav_n^{(l+1)} \right)} \right],
\end{align}
where $\Phimat^{(L+1)}:=\rv$, $\thetav_n^{(L+1)}:=\emptyset$, and the expectations are taken with respect to the variational distribution as
\begin{equation}\label{approximation-pfa}
  q \left(  \{
   \thetav_n^{(l)} \}_{n=1,l=1}^{N,L} \right) =
    \prod \limits_{n=1}^N
   \prod \limits_{l=1}^L q \left(  \thetav_{n}^{(l)} \,|\,  \Phimat^{(l+1)}, \thetav_n^{(l+1)} \right).
\end{equation}
To simplify the optimization, one often resorts to the mean-field assumption that factorizes the variational distribution as
\begin{equation}\label{meanfield-approximation-pfa}
  q \left(  \{
   \thetav_n^{(l)} \}_{n=1,l=1}^{N,L} \right) =
    \prod \limits_{n=1}^N
   \prod \limits_{l=1}^L q \left(  \thetav_{n}^{(l)} \right).
\end{equation}
Furthermore, to achieve fast out-of-sample prediction with autoencoding variational inference, one may consider a gamma distribution based inference network %
 as $  q(\thetav_n^{(l)} \,|\, \xv_n) = \mbox{Gamma}(f_{\Wmat}^{(l)}(\xv_n),g_{\Wmat}^{(l)}(\xv_n)) $ to  model sparse and nonnegative latent document representation, where $f^{(l)}$ and $g^{(l)}$ are related DNNs parameterized by $\Wmat$. However, it is hard to efficiently compute the gradient of the ELBO with respect to $\Wmat$, especially if $L \geq 2$,  due to the difficulty to reparameterize  a gamma random variable \cite{kingma2014stochastic,ruiz2016generalized,knowles2015stochastic}, motivating us to identify a surrogate distribution that can not only  well approximate the gamma distribution, but also be easily reparameterized.  Below we show the Weibull distribution is an ideal choice.

{\bf{Weibull variational posterior.}}
A main reason that we choose the Weibull distribution to construct the inference network is that the Weibull and gamma distributions have similar PDFs, which makes it possible to model sparse and nonnegative latent representation:
\begin{align} %
\notag
 &\mbox{Weibull PDF: } P(x\,|\, k,\lambda)= \frac{k}{{\lambda}^k} x^{k-1} e^{(-x/\lambda)^k},~~~ \\
  &\mbox{Gamma PDF: } P(x\,|\, \alpha, 1/\beta) =\frac{{\beta}^{\alpha}}{\Gamma(\alpha)} x^{\alpha-1} e^{-\beta x},
\end{align}
where $x\in\mathbb{R}_+$.
Another reason is due to a simple reparameterization for $x\sim\mbox{Weibull}(k,\lambda)$ as
\begin{equation} %
  x = \lambda (-\ln(1-\epsilon)) ^ {1/k},~ \epsilon \sim \mbox{Uniform}(0,1),
\end{equation}
leading an easy-to-compute gradient when maximizing the ELBO.
Moreover, denoting $\gamma$ as the Euler--Mascheroni constant, the KL-divergence from the gamma to Weibull distribution has an analytic expression as
\begin{align} \label{Gamma-Weibull-KL}
 &\mbox{KL}(\mbox{Weibull}(k,\lambda)||\mbox{Gamma}(\alpha, 1/\beta))
   = -\alpha \ln \lambda + \frac{\gamma \alpha}{k}+ \ln k \notag \\
    &+ \beta \lambda \Gamma\Big(1+\frac{1}{k}\Big) - \gamma - 1 - \alpha \ln \beta + \ln \Gamma (\alpha),
\end{align}
which helps reduce the variance 
when evaluating the gradient of the ELBO \cite{kingma2014stochastic}.
Minimizing this KL divergence, one can identify the two parameters of a Weibull distribution to approximate a given gamma one.
As shown in Fig. \ref{fig:gamma-weibull-kl}, the inferred Weibull distribution in general  accurately approximates the target gamma one, as long as the gamma shape parameter is neither too close to zero nor  too large.

{\bf{Upward-downward information propagation.}}
With the DLDA upward-downward Gibbs sampler  sketched in Fig. \ref{fig:DLDA-strucure} and the corresponding sampling equation
\begin{equation}\label{theta-gibbs-equation}
 \small ( \thetav_n^{(l)} \,|\, -) \sim \mbox{Gamma}\left(\mv_n^{(l)(l+1)} + \Phimat^{(l+1)}\thetav_n^{(l+1)}, f(p_n^{(l)},c_n^{(l+1)})\right),
\end{equation}
where $\mv_n^{(l)(l+1)}$ and $p_n^{(l)}$ are latent random variables constituted by information upward propagated to layer $l$,
it is clear that the conditional posterior of $\thetav_n^{(l)}$ is related to both the information at the higher (prior) layer, and that  upward propagated to the current layer via a series of data augmentation and marginalization steps; see Zhou et al.\cite{GBN} for more details. %
Considering that VAE-like models usually  build  the upward propagation but ignore the impact of the prior,
inspired by the instructive upward-downward information propagation in Gibbs sampling, as shown in Fig. \ref{fig:DWVAE-strucure}, we construct WUDVE, the inference network of our model, as $q(\thetav_n^{(L)} \,|\,
\hv_n^{(L)}) \prod_{l=1}^{L-1} q(\thetav_n^{(l)} \,|\, \Phimat^{(l+1)} ,
\hv_n^{(l)},\thetav_n^{(l+1)})$,
where
\begin{align}\label{posterior-completely}
&q(\thetav_n^{(l)} \,|\, \Phimat^{(l+1)} ,
\hv_n^{(l)},\thetav_n^{(l+1)}) \nonumber \\
&= \mbox{Weibull}(\kv_n^{(l)}+\Phimat^{(l+1)} \thetav_n^{(l+1)},\lambdav_n^{(l)}).
\end{align}
The Weibull distribution is used to approximate the gamma distributed conditional posterior, and its parameters $\kv_n^{(l)}\in \mathbb{R}^{K_l}$ and $\lambdav_n^{(l)}\in \mathbb{R}^{K_l}$ are both deterministically transformed from the observation~$\xv_n$ using the NNs, as illustrated in Fig. \ref{fig:DWVAE-strucure} and specified  as
\begin{align}
&\kv_n^{(l)} = \ln[1+\exp(\Wmat_1^{(l)}\hv_n^{(l)}+ \bv_1^{(l)})], \\
&\lambdav_n^{(l)} =  \ln[1+\exp(\Wmat_2^{(l)}\hv_n^{(l)}+ \bv_2^{(l)})], \label{MLP2}\\
&\hv_n^{(l)} = \ln[1+\exp(\Wmat_3^{(l)}\hv_n^{(l-1)}+\bv_3^{(l)})],%
\end{align}
where $\hv_n^{(0)}=\log (1+\xv_n)$, $\Wmat_1^{(l)}\in\mathbb{R}^{K_l\times K_{l}}$, $\Wmat_2^{(l)}\in\mathbb{R}^{K_l\times K_{l}}$, $\Wmat_3^{(l)}\in\mathbb{R}^{K_l\times K_{l-1}}$, $\bv_1^{(l)}\in\mathbb{R}^{K_l}$, $\bv_2^{(l)}\in\mathbb{R}^{K_l}$, and $\bv_3^{(l)}\in\mathbb{R}^{K_l}$.
This upward-downward inference network is distinct from that of a usual VAE,  where it is common that the inference network has a pure bottom-up structure and only interacts with the generative model via the ELBO \cite{kingma2014stochastic,pixelVAE}.
Note that it does not follow mean-field variational Bayes to make a fully factorized assumption as in \eqref{meanfield-approximation-pfa}.

Comparing Figs. \ref{fig:DLDA-strucure} and \ref{fig:DWVAE-strucure} shows that in each iteration, both Gibbs sampling in DLDA and the hybrid Bayesian inference in DATM have not only upward information propagations (orange arrows), but also downward ones (blue arrows), but there are distinctions between their underlying implementations.
Gibbs sampling in  Fig.  \ref{fig:DLDA-strucure} does not have an inference network and needs the local variables $\thetav_n^{(l)}$ to help perform stochastic upward information propagation, whereas DATM in Fig. \ref{fig:DWVAE-strucure} uses a ladder network to combine a deterministic upward and stochastic downward information propagation, without relying on the local variables $\thetav_n^{(l)}$.
It is also interesting to notice that the upward-downward  structure, motivated by the upward-downward Gibbs sampler of DLDA,  is closely related to the ladder structure used in the ladder VAE \cite{sonderby2016ladder}.
However, to combine the bottom-up and top-down information, ladder VAE relies on some heuristics  restricted to Gaussian latent variables.

\subsection{Weibull hybrid autoencoding inference (WHAI)}
Based on the above discussion, in DATM, we need to infer the topic parameters $\{\Phimat ^{(l)}\} _{l=1}^{L}$ of the decoder network and the NN parameters  $\Omegamat=\{\Wmat_{1}^{(l)}, \bv_1^{(l)},\Wmat_{2}^{(l)}, \bv_2^{(l)},\Wmat_{3}^{(l)}, \bv_3^{(l)}\}_{1,L}$ of the encoder network.

Rather than merely finding point estimates, we describe in Algorithm \ref{Algorithm} how to combine TLASGR-MCMC  and  WUDVE into a hybrid SG-MCMC/VAE inference algorithm,  which infers  posterior samples for $\{\Phimat ^{(l)}\} _{l=1}^{L}$ and $\Omegamat$.
An important step of Algorithm \ref{Algorithm} is calculating the gradient of the ELBO in \eqref{ELBO-of-VADTM} with respect to the NN parameters $\Omegamat$, which is important to the success of a variational inference algorithm
\cite{hoffman2013stochastic, blei2012variational, kingma2014stochastic, mnih2014neural2, ruiz2016generalized, rezende2014stochastic}.
Thanks to the choice of the Weibull distribution, the second term of the ELBO in \eqref{ELBO-of-VADTM} is analytic, and due to simple reparameterization of the Weibull distribution, the gradient of the first term of the ELBO with respect to $\Omegamat$ can be accurately evaluated, achieving satisfactory performance using as few as a single Monte Carlo sample, as shown in our experimental results.
Thanks to the architecture of DATM using the inference network, for a new mini-batch, different from Cong et al.  \cite{cong2017deep} that run hundreds of MCMC iterations to collect posterior samples for local variables, we can directly find the conditional posterior of $\{ \thetav_n^{(l)} \}_{l=1}^{L}$ given $\{ \Phimat^{(l)} \}_{l=1}^{L}$ and the stochastically updated $\Omegamat$, with which we can sample the local parameters and then use TLASGR-MCMC to stochastically update the global parameters  $\{\Phimat^{(l)}\}_{1,L}$.

\begin{algorithm}[!t]
\caption{Hybrid stochastic-gradient MCMC and autoencoding variational inference for DATM}
\begin{algorithmic}
    \STATE {\bf{Input:}} Observed data $\{ \xv_n \}_n$, the structure of DATM, and hyper-parameters.

    \STATE {\bf{Output:}} Global parameters of DATM $\{\Phimat^{(l)}\}_{1,L}$ and $\Omegamat$.

    \STATE Set mini-batch size $m$;
    \STATE Initialize encoder parameters $\Omegamat$ and decoder parameters $ \{ \Phimat^{(l)} \}_{1,L}$.

    \FOR{$iter = 1,2, \cdots$ }
    \STATE %
    Randomly select a mini-batch of $m$ documents to form
    a subset $\Xmat = \{ \xv_i \}_{1,m}$;

    Draw random noise $\left\{ {{\varepsilon _i^l}} \right\}_{i = 1,l=1}^{m,L}$ from uniform distribution;

    Calculate  $ {\nabla _{\Omegamat }} L\left( \Omegamat,\Phimat ^{ \{ l \}} ;{\Xmat},{\varepsilon _i^l} \right)$ according to \eqref{ELBO-of-VADTM}, and update $\Omegamat$;

    Sample $\left\{\thetav_i^{(l)}\right\}_{i=1,l=1}^{m,L}$ from \eqref{posterior-completely} via $\Omegamat$ ;

    \FOR{$l=1, \cdots, L+1$ and $k=1, \cdots, K_l$}

    \STATE Update $M_k^{(l)}$ with \eqref{update_M}; then topics $ \{ \Phimat^{(l)} \}_{l=1}^{L}$ with \eqref{Phi-updata} and $\rv$ with \eqref{r_update}.

    \ENDFOR

    \ENDFOR

\end{algorithmic}\label{Algorithm}
\end{algorithm}

\subsection{Learning the network structure with layer-wise training}\label{sec:layer-wise}
{{Distinct from some existing unsupervised  learning algorithms that train deep networks in a greedy layer-wise manner, such as the one proposed in Hinton et al. \cite{Hinton06} for training the deep belief networks,
 DATM is equipped with a SG-MCMC/VAE hybrid Bayesian inference algorithm that can jointly train all its hidden layers, as described in Algorithm \ref{Algorithm}.
However, the same as most existing algorithms in deep learning, it still needs to specify the %
width of each layer,

In this paper, motivated by related work in Zhou et al. \cite{GBN,PBDN2018}, we adopt the idea of layer-wise training for DATM for the purpose of learning the width of each hidden layer in a greedy layer-wise manner, given a fixed budget on the width of the first layer.
The proposed layer-wise training strategy is summarized in Algorithm \ref{Algorithm2}.
With a DATM of $L-1$ layers that has already been trained, the key idea is to use a truncated gamma-negative binomial process \cite{NBP2012} to model the latent count matrix for the newly added top layer as $m_{kn}^{(L)(L+1)} \sim \mbox{NB} (r_k,p_n^{(L+1)})$, $r_k \sim \mbox{Gam} (\gamma_0/K_{L max},1/c_0)$, and rely on that stochastic process's shrinkage mechanism to prune inactive factors of layer $L$ according to the values of $\{r_k\}_k$.
Generally speaking, the inferred $K_L$ would be clearly smaller than $K_{L max}$ if $K_{L max}$ is sufficiently large.
As in Algorithm \ref{Algorithm2},  $K_{1 max}$ is a parameter to set, whereas the inferred width of layer $l-1$, $K_{l-1}$, is set as the maximum number of factors of a newly added layer $K_{l max}$.
More details on this greedy layer-wise learning strategy can be found in Zhou et al. \cite{GBN}. }}

\subsection{Variations of WHAI}\label{sec:Variations}
To clearly understand how each component contributes to the overall performance of WHAI, below we consider some different variations.

{\bf{Gamma hybrid autoencoding inference (GHAI): }}In the inference network of DATM, the reparameterizable Weilbull distribution is chosen to be the variational posterior and used to connect adjacent stochastic layers for the reasons specified in Section \ref{sec:dist}.
One may also choose some other distributions to construct the variational posterior. For example, one may %
replace \eqref{posterior-completely} with \begin{align}\label{RSVI-posterior}
&q(\thetav_n^{(l)} \,|\, \Phimat^{(l+1)} ,
\hv_n^{(l)},\thetav_n^{(l+1)}) \nonumber \\
&=
\mbox{Gamma}(\kv_n^{(l)}+\Phimat^{(l+1)} \thetav_n^{(l+1)},\lambdav_n^{(l)}).
\end{align}
While the gamma distribution does not have a simple reparameteriation, one may use RSVI %
\cite{RSVI}  to define an approximate reparameterization procedure via rejection sampling.
More specifically, following  Naesseth et al. \cite{RSVI}, to generate a gamma random variable $z \sim \mbox{Gamma}(\alpha, \beta)$,  one may first use  the rejection sampler \cite{marsaglia2000simple} to generate $\tilde{z} \sim \mbox{Gamma}(\alpha+B,1)$, for which the proposal distribution is expressed as
$$
\tilde{z}=\left(\alpha+B-\frac{1}{3}\right)\left(1+\frac{\varepsilon}{\sqrt{9(\alpha+B)-3}}\right)^3,~~\varepsilon\sim\mathcal{N}(0,1),
$$
where $B$ is a pre-set integer to make the acceptance probability be close to 1; one
then lets
$z=1/\beta*\tilde{z}\prod_{i=1}^B {u_i}^{1/(\alpha+i-1)}$, where
$u_i \sim \mbox{Uniform}(0,1)$.
The gradients
with respect to the ELBO, however, could still suffer from relatively high variance, as how likely a proposed $\varepsilon$ will be accepted depends on the gamma distribution parameters, and $B$ extra uniform random numbers $\{u_i\}_{1,B}$ need to be introduced.

{\bf{Weibull autoencoding inference (WAI): }}
To illustrate the effectiveness of the proposed hybrid Bayesian inference, we also consider WAI that has the same inference network as WHAI but infers
$\{ \Phimat ^{(l)}\} _{1,L}$ and $\Omegamat$ only using SGD.
Although as argued in Mandt et al. \cite{Mandt2017Stochastic}, SGD can also be used for approximate Bayesian inference, and it performs well in AVITM \cite{srivastava2017autoencoding},
we will show in  experiments that sampling the global parameters via TLASGR-MCMC
provides improved performance in comparison to updating them via SGD.

{\bf{Independent WHAI (IWHAI): }}To understand the importance of the stochastic-downward  structure used in the inference network,
we also consider IWHAI
that remove the stochastic-downward connections of DATM-WHAI.
More specifically, IWHAI redefines
$q(\thetav_n^{(l)} \,|\, \Phimat^{(l+1)} ,
\hv_n^{(l)},\thetav_n^{(l+1)})$ in \eqref{posterior-completely} as $\mbox{Weilbull}(\kv_n^{(l)},\lambdav_{n}^{(l)})$, and uses the same hybrid Bayesian inference  to infer $\{ \Phimat ^{(l)}\} _{1,L}$ and $\Omegamat$.

\begin{algorithm}[!t]
\caption{Hybrid stochastic-gradient MCMC and autoencoding variational inference for DATM, which uses a layer-wise training strategy to train a set
of networks, each of which adds an additional hidden layer on top of the previously inferred network, retrains all its layers jointly, and prunes inactive factors from the last layer.}
\begin{algorithmic}
    \STATE {\bf{Input:}} Observed data $\{ \xv_n \}_n$, upper bound of the width of the first layer $K_{1 max}$, the number of layers $L$, the pruned threshold $u$, and hyper-parameters.

    \STATE {\bf{Output:}} Global parameters of DATM $\{\Phimat^{(l)}\}_{1,L}$ and $\Omegamat$.

    \STATE Set mini-batch size $m$;

    \FOR{$l = 1,2, \cdots L$}

     \STATE Set $K_{l-1}$, the inferred width of layer $l-1$, as $K_{l max}$, the upper bound of layer $L$'s width.

    \STATE Initialize encoder parameters $\Omegamat^{(l)}$ and decoder parameters $ \Phimat^{(l)} $, combined with $\{\Omegamat^{(t)}\}_{t=1}^{l-1}$ and $ \{\Phimat^{(t)}\}_{t=1}^{l-1} $.

    \FOR{$iter = 1,2, \cdots$ }
    \STATE %
    Randomly select a mini-batch of $m$ documents to form
    a subset $\Xmat = \{ \xv_i \}_{1,m}$;

    Draw random noise $\left\{ {{\varepsilon _i^t}} \right\}_{i = 1,t=1}^{m,l}$ from uniform distribution;

    Calculate  $ {\nabla _{\Omegamat }} L\left( \Omegamat,\Phimat ^{ \{ t \}} ;{\Xmat},{\varepsilon _i^t} \right)$ according to \eqref{ELBO-of-VADTM}, and update $\Omegamat$;

    Sample $\left\{\thetav_i^{(t)}\right\}_{i=1,t=1}^{m,l}$ from \eqref{posterior-completely} via $\Omegamat$ ;

    \FOR{$l=1, \cdots, L+1$ and $k=1, \cdots, K_l$}

    \STATE Update $M_k^{(l)}$ with \eqref{update_M}, topics $ \{ \Phimat^{(l)} \}_{l=1}^{L}$ with \eqref{Phi-updata}, and $\rv$ with \eqref{r_update}.

    \ENDFOR
    \ENDFOR

    \STATE Delete the inactive topics of $\Phimat^{(l)}_k$ if $r_k<u$, and delete the corresponding paramters in $\Omegamat^{(l)}$. Output the inferred width $K_l$.

    \ENDFOR

\end{algorithmic}\label{Algorithm2}
\end{algorithm}

\section{Supervised DATM for classification}

With DATM, we are able to efficiently  infer the topics of DLDA \cite{zhou2015poisson, cong2017deep}  and directly project a document  into its latent representation at multiple stochastic hidden layers, providing a new opportunity to learn  interpretable  latent representation that can well generate not only the observed bag of words of documents, but also the class labels that are often associated with documents. Thus, rather than following a two-step procedure to first apply DATM and then build a classifier on its unsupervisedly extracted latent features, we generalize DATM to a generative model for both the observed bags of words and labels, referred to as supervised DATM (sDATM), exploiting the synergy between document generation and classification to achieve enhanced  performance.

\subsection{Label generation}
We consider a labeled document corpus %
$\{ \xv_n,y_n \}_{n=1}^N$, where  $y_n\in \{1,2,\cdots,C\}$ and $C$ is the total number of classes.
We assume the label is generated from a categorical distribution $y_n\sim\mbox{Categorical}(p_{n1},\ldots, p_{nC})$, where $p_{nc}$ is the probability that $\xv_n$ belongs to class $c$, which means
\begin{equation}\label{Catogorical distribution}
  p(y_n) = \prod_{c=1}^C p_{nc}^{\delta(y_n=c)},
\end{equation}
where
$\delta(\cdotv)$ is an indicator function that is equal to one if the argument is true and zero otherwise.

In a usual supervised-learning setting that maps an observation to its label via a deterministic deep NN, it is often only the features at the top hidden layer (furthest from the data) that are transformed to define the label probabilities $p_{nc}$.  For DATM, as the latent representation $\thetav_n^{(l)}$ at different hidden layers are stochastically connected,  the topics at different stochastic layers reveal different levels of abstraction, and it is the features at the bottom hidden layer (closest to the data) that are directly responsible for data generation, we are motivated to concatenate $\thetav_n^{(l)}$ across all hidden layers to construct a latent feature vector~as
\begin{equation}\label{fearure_fusion}
  \sv_n = \left[ \thetav_n^{(1)}, \cdots, \thetav_n^{(L)} \right].
\end{equation}
With this concatenation, %
the label information is directly used to influence the features across all layers, which helps improve
 the discrimination power and robustness of the learned features%
~\cite{lee2014deeply-supervised}. %

To map from $\sv_n$ to its label probability vector $\pv_n=(p_{n1},\ldots,p_{nC})$, we first consider a linear setting that lets
\begin{equation}\label{Linear-classification}
  \pv_n = \left[ \frac{e^{\wv_1^T \sv_n}}{\sum_{c=1}^C e^{\wv_c^T \sv_n}}, \cdots, \frac{e^{\wv_C^T \sv_n}}{\sum_{c=1}^C e^{\wv_c^T \sv_n}} \right],
\end{equation}
where $\Wmat_c = [\wv_1, \cdots, \wv_C]$ can be considered as the coefficients of a linear classifier, whose features are the concatenation of the latent features projected from $\xv_n$ using \eqref{posterior-completely}.

Note although the features $\sv_n$ in \eqref{Linear-classification} is nonlinear transformed from $\xv_n$, those nonlinear mappings are primarily used to approximate the posterior of $\{\thetav_n^{(l)}\}$. %
To further boost the performance of classification, $L$ layer-specific multi-layer perceptrons (MLPs) $\{g_1^{(l)}\}_{l=1}^L$ are used to map layer-specific feature spaces to a concatenated feature space as
\begin{equation}\label{fearure_fusion_nonlinear}
  \sv_n = \left[ g_1^{(l)}(\thetav_n^{(1)}), \cdots, g_1^{(L)}(\thetav_n^{(L)}) \right].
\end{equation}
Then, another MLP $g_2$ is used to transform the concatenated features $\sv_n$ to the probabilistic space as
\begin{equation}\label{nonlinear-classification}
  \pv_n = \left[ \frac{e^{\wv_1^T g_2(\sv_n)}}{\sum_{c=1}^C e^{\wv_c^T g_2(\sv_n)}}, \cdots, \frac{e^{\wv_C^T g_2(\sv_n)}}{\sum_{c=1}^C e^{\wv_c^T g_2(\sv_n)}} \right].
\end{equation}
We use $\Wmat_m$ to denote all parameters in MLPs.
No matter for the linear model or the nonlinear one, the label likelihood \eqref{Catogorical distribution} can be rewritten as $p(y_n\,|\,\{\thetav_n^{(l)}\}_{l=1}^L)$, resulting in a fully-generative model for $\{ \xv_n, \yv_n\}_{n=1}^N$ as shown in Fig. \ref{fig:SDLDA-strucure}.
The linear model and the nonlinear one are represented as $\text{sDATM-L}$ and sDATM-N, respectively, whose inference models are the same with DATM shown in Fig. \ref{fig:DWVAE-strucure}.

\subsection{Model learning and prediction}\label{sDPATM_learning}
With the generative process of sDATM, we can write the ELBO of $p(\xv,\yv)$ as
\begin{align}\label{ELBO-of-classification}
L = &\sum \limits_{n=1}^N \mathbb{E} \left[ \ln p\left( \xv_n \,|\, \Phimat^{(1)}, \thetav_n^{(1)} \right) + %
 \ln p\left( \yv_n \,|\,  \{\thetav_n^{(l)}\}_{l=1}^L \right) \right] \nonumber \\
&- \sum \limits_{n=1}^N \sum \limits_{l=1}^L \mathbb{E} \left[ \ln \frac{q\left( \thetav_n^{(l)} \right)}{p \left( \thetav_n^{(l)} \,|\, \Phimat^{(l+1)}, \thetav_n^{(l+1)} \right)} \right] \nonumber \\
& - \sum \limits_{c=1}^C KL\left[ q(\wv_c)||p(\wv_c) \right],
\end{align}
where the expectations are taken with respect to $q \left(\{
\thetav_n^{(l)} \}_{n=1,l=1}^{N,L} \right)$, modeled by \eqref{posterior-completely}, and $q(\{\wv_c\}_{c=1}^C)$.
 The prior and the variational posterior of $\{\wv_c\}_{c=1}^{C}$ are set as diagonal Gaussian distributions \cite{graves2011practical} \cite{Blundell2015Weight} as
\begin{align}\label{w-distribution}
  p(\wv_c) &= \mathcal{N}(\zerov,\Imat),\nonumber \\
  q(\wv_c) &= \mathcal{N}(\muv_c,\mbox{diag}(\sigmav_c)),
\end{align}
resulting in an analytic KL divergence as
\begin{equation}\label{KL-gaussian}
  \mbox{KL}\left[ q(\wv_c)||p(\wv_c) \right] = \frac{1}{2}\left( ||\muv_c||_2^2 + ||\sigmav_c||_2^2 \right) - \log ||\sigmav_c||,
\end{equation}
which can be viewed as a prior regularization on $\wv_c$.
To ensure $\sigmav_c$ to be nonnegative, we parameterize it pointwise as $\sigmav_c = \log(1+\exp(\rhov_c))$ and update  the variational parameters $\left\{ \muv_c,\rhov_c \right\}_{c=1}^C$ with a usual backpropagation algorithm.
{ For nonlinear sDATM, the network structure of $\Wmat_m$ in \eqref{fearure_fusion_nonlinear} and \eqref{nonlinear-classification} is set to:
\begin{align}\label{eq:Wm}
  g_1^{(l)}(\thetav_n^{(l)}) &= \ln [1+\exp(\Wmat_{m1}^{(l)} \thetav_n^{(l)} + \bv_{m1}^{(l)}) ], \nonumber \\
  \hv_n &= \ln [1+\exp(\Wmat_{m2} \sv_n + \bv_{m2}) ], \nonumber \\
  g_2(\sv_n) &= \ln [1+\exp(\Wmat_{m3} \hv_n + \bv_{m3}) ],
\end{align}
where $\{ \Wmat_{m1}^{(l)}\}_{l=1}^L \in \mathbb{R}^{K_l \times K_l} $, $\{ \Wmat_{m2}\} \in \mathbb{R}^{a_1 \times \sum_l K_l} $, $\{ \Wmat_{m3}\} \in \mathbb{R}^{a_2 \times a_1} $, $\{ \bv_{m1}^{(l)}\}_{l=1}^L \in \mathbb{R}^{K_l} $, $\bv_{m2} \in \mathbb{R}^{a_1} $,  and $\bv_{m3} \in \mathbb{R}^{a_2} $, with $a_1=400$, $a_2=200$ in the experiments.

With the inferred variational parameters of the inference network,
at the test stage,
approximating the intractable expectation $\mathbb{E}_{q(\sv,\wv_{1:C} \,|\, \xv)}[p(y\,|\,\sv,\wv_{1:C})]$ with Monte Carlo estimation, we can predict the label of a testing document as %
\begin{equation}\label{prediction}
\textstyle
  y=\mbox{argmax}_c \left( %
  \sum_{j=1}^{N_{collect}}
  p(y=c\,|\, \wv_1^{(j)},\sv^{(j)})%
  \right)_{c=1,C},
\end{equation}
where $\wv_c^{(j)} \sim q(\wv_c)$ and $\thetav^{(j)} \sim q(\thetav)$ accord to \eqref{w-distribution} and \eqref{posterior-completely}, respectively, $\sv^{(j)}$ is deterministically transformed from $\thetav^{(j)}$, and $N_{collect} = 50$ is used.

\begin{table*}[ht]
\centering
\caption{Comparisons of per-heldout-word perplexity and testing time (average seconds per document, with 3000 random samples) on three different datasets.}
\begin{tabular}{cc|ccc|ccc}
\hline
\mr{2}*{Model} & \mr{2}*{Size} & \mc{3}{c|}{Perplexity} & \mc{3}{c}{Test Time } \\ \cline{3-8}
      &           & 20News & RCV1 & Wiki & 20News & RCV1 & Wiki \\ \hline  \hline
LDA & 128 &  593 &  1039 &  1059 &  4.14 &  11.35 & 12.16 \\
OR-softmax & 128-64-32 &  592 &  1013 &  1024 &  3.20 &  8.64 & 9.77 \\
DocNADE & 128 &  591 &  969 &  999 &  0.42 &  0.90 & 1.04 \\
DPFA  & 128-64-32 &  637 & 1041 & 1056 &  20.12 &  34.21 & 35.41\\
AVITM & 128       &  654 & 1062 & 1088 &  0.23 &  0.68 & 0.80\\ \hline
DLDA-Gibbs & 128-64-32 &  571 &  938 &  966 &  10.46
 &  23.38 & 23.69 \\
DLDA-Gibbs & 128-64    &  573 &  942 &  968 &  8.73 &  18.50 & 19.79 \\
DLDA-Gibbs & 128       &  584 &  951 &  981 &  4.69 &  12.57 & 13.31 \\
DLDA-TLASGR & 128-64-32 &  579 &  950 &  978 &  10.46 & 23.38  & 23.69 \\
DLDA-TLASGR & 128-64    &  581 &  955 &  979 & 8.73  & 18.50  & 19.79 \\
DLDA-TLASGR & 128       &  590 &  963 &  993 & 4.69  & 12.57  & 13.31 \\  \hline
DATM-GHAI & 128-64-32 &  604 & 963 &  994 & 0.66  & 1.25  & 1.49 \\
DATM-GHAI & 128-64      &  608 & 965 & 997 & 0.44  & 0.96  & 1.05 \\
DATM-GHAI & 128 &  615 & 972 &  1003 & 0.22  & 0.69  & 0.80 \\ \hline
DATM-IWHAI & 128-64-32       & 588 & 964 & 990 & 0.58  & 1.15  & 1.38 \\
DATM-IWHAI & 128-64      &  589 & 965 &  992 & 0.38  & 0.87  & 0.97 \\
DATM-IWHAI & 128  &  592 & 966 &  996 & 0.20  & 0.66  & 0.78 \\ \hline
DATM-WAI & 128-64-32 &  581 &  954 &  984 & 0.63  & 1.20  & 1.43 \\
DATM-WAI  & 128-64    &  583 &  958 &  986 & 0.42  & 0.91 & 1.02 \\
DATM-WAI  & 128       &  593 &  967 &  999 & 0.20  & 0.66  & 0.78\\ \hline
DATM-WHAI  & 128-64-32 &  581 &  953 &  980 &  0.63 &  1.20 & 1.43\\
DATM-WHAI  & 128-64    &  582 &  957 &  982 &  0.42 &  0.91 & 1.02 \\
DATM-WHAI  & 128       &  591 &  965 &  996 &  0.20 &  0.66 & 0.78\\ \hline
\end{tabular}\label{Tab:perplexity and time}
\end{table*}

\section{Experimental results}
In this paper, 
DATM is proposed for extracting deep latent features and analyzing documents unsupervisedly, and sDATM is proposed for joint deep topic modeling and document classification.
In this section, the performance of the proposed models are demonstrated through both unsupervised and supervised learning tasks on big corpora.
Our code is written based on Theano \cite{Theano}.

\subsection{Unsupervised learning for document representation}
\subsubsection{Per-heldout-word perplexity}
We first compare the per-heldout-word perplexity \cite{zhou2015poisson,gan2015scalable,henao2015deep}, a widely-used performance measure, of different models on 20Newsgroups (20News), Reuters Corpus Volume I (RCV1), and Wikipedia (Wiki).
20News consists of 18,845 documents with a vocabulary size of 2,000.
RCV1 consists of 804,414 documents with a vocabulary size of 10,000.
Wiki, with a vocabulary size of 7,702, consists of 10 million documents randomly downloaded from Wikipedia using the script provided by Hoffman et al. \cite{OnlineLDA}.
For Wiki,
we randomly select 100,000 documents for testing, and
to be consistent with previous settings \cite{gan2015scalable,henao2015deep,cong2017deep}, no precautions are taken in the  Wikipedia downloading script to prevent a testing document from being downloaded into a mini-batch for training.

For comparison, the models included in our comparison are listed as follows, using the code provided by the authors:
\begin{itemize}
  \item {\bf{LDA}}: Latent Dirichlet allocation \cite{blei2003latent} is a basic probability topic model, which is closely related to a single-hidden-layer version of DLDA. We run it by onlineVB.
  \item {\bf{OR-softmax}}: Over-replicated softmax \cite{srivastava2013modeling} is a type of deep Boltzmann machine that is suitable for extracting distributed semantic representation from documents.
  \item {\bf{DocNADE}}: Document neural autoregressive distribution estimation  \cite{larochelle2012a} is an autoregressive distribution estimator based on feed-forward NNs for text analysis. %
  \item {\bf{DPFA}}: Deep Poisson factor analysis \cite{gan2015scalable} is a hierarchical model for text analysis based on Poisson factor analysis and sigmoid belief network.
  \item {\bf{AVITM}}: Autoencoding variational inference for topic modeling \cite{srivastava2017autoencoding} is an autoencoding topic model based on a single-hidden-layer  LDA.
  \item {{\bf{DLDA-Gibbs}} and {\bf{DLDA-TLASGR}}}: Deep latent Dirichlet allocation inferred by Gibbs sampling \cite{GBN} and by TLASGR-MCMC \cite{cong2017deep}, respectively.
  \end{itemize}
  In order to further demonstrate the advantages of the stochastic upward-downward structure and hybrid inference algorithm, some variants including {\bf{DATM-GHAI}}, {\bf{DATM-WAI}} and {\bf{DATM-IWHAI}} discussed in Section \ref{sec:Variations} are also included for  comparison.
Note that as shown in Cong et al. \cite{cong2017deep}, DLDA-Gibbs  and DLDA-TLASGR  are state-of-the-art topic modeling algorithms that outperform a large number of previously proposed ones, such as deep Poisson factor modeling \cite{henao2015deep} and  the nested hierarchical Dirichlet process \cite{nHDP}.

\begin{table*}[!t]
	\centering
	\caption{Comparisons of per-heldout-word perplexity by layer-wise training strategy to infer the network structure (the same settings with Table \ref{Tab:perplexity and time}) on three different datasets.}
	\begin{tabular}{c|ccc|ccc}
		\hline
		\mr{2}*{$K_{1 max}$}  & \mc{3}{c|}{Inferred structure} & \mc{3}{c}{Perplexity} \\ \cline{2-7}
		& 20News & RCV1 & Wiki & 20News & RCV1 & Wiki \\ \hline \hline
		64  &  64-62-55 &  64-64-61 &  64-64-59 &  584 & 959 & 987  \\ \hline
		128  &  121-110-84 &  126-118-102 &  123-114-96 &  578 &  949 & 978 \\ \hline
		256  &  248-211-183 &  253-220-196 &  250-217-188 &  574 &  943 & 972 \\ \hline
		512  &  470-197-155 &  482-201-167 &  471-193-170 &  574 &  941 & 971 \\ \hline
		1024  &  478-199-160 &  484-201-163 &  472-190-174 &  573 &  940 & 971 \\ \hline
	\end{tabular}\label{Tab:inferred structure}
\end{table*}

\begin{figure*}[!t]
	\centering
	\subfloat[]{\includegraphics[scale=0.17]{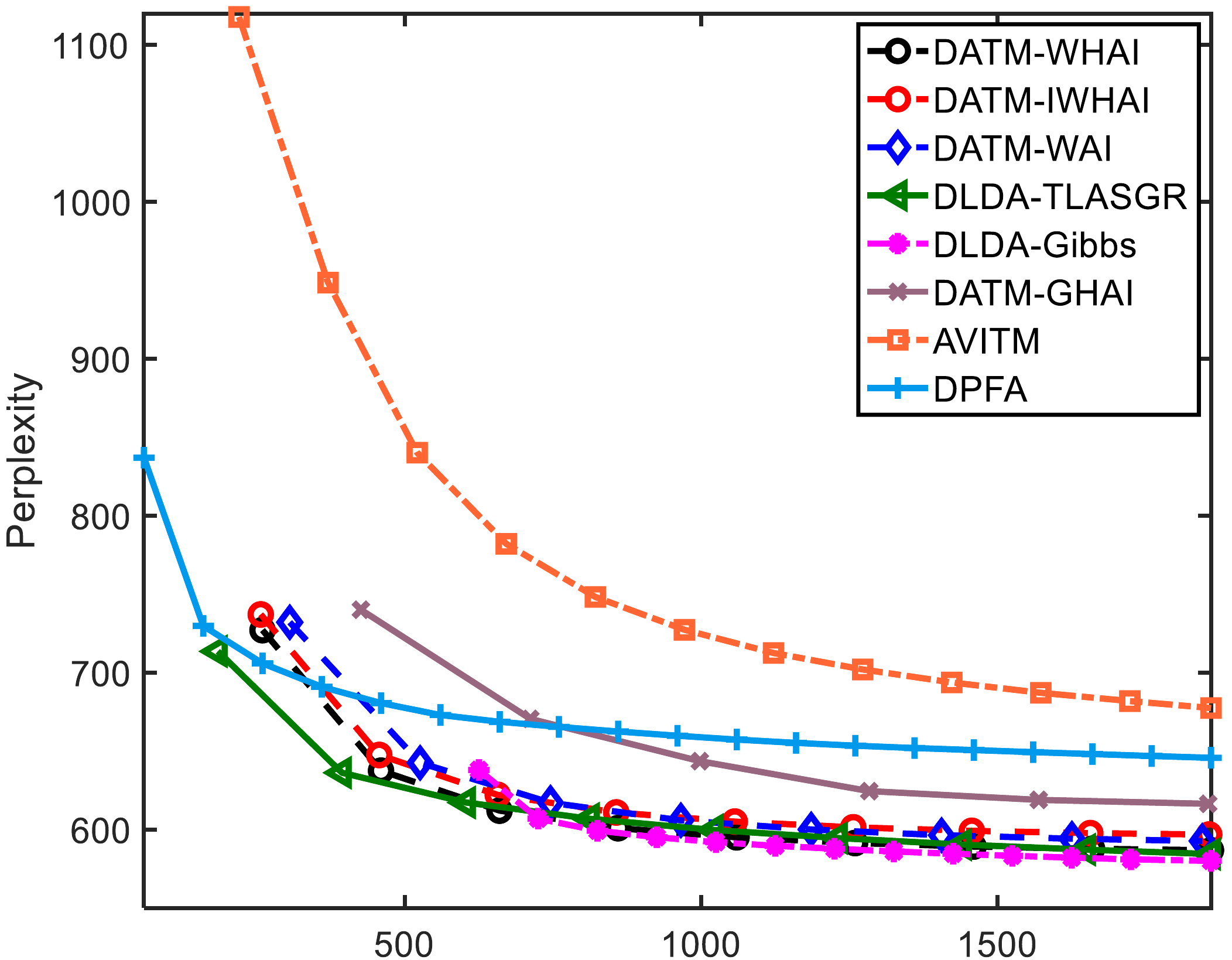}\label{20News_journal}} \quad
	\subfloat[]{\includegraphics[scale=0.17]{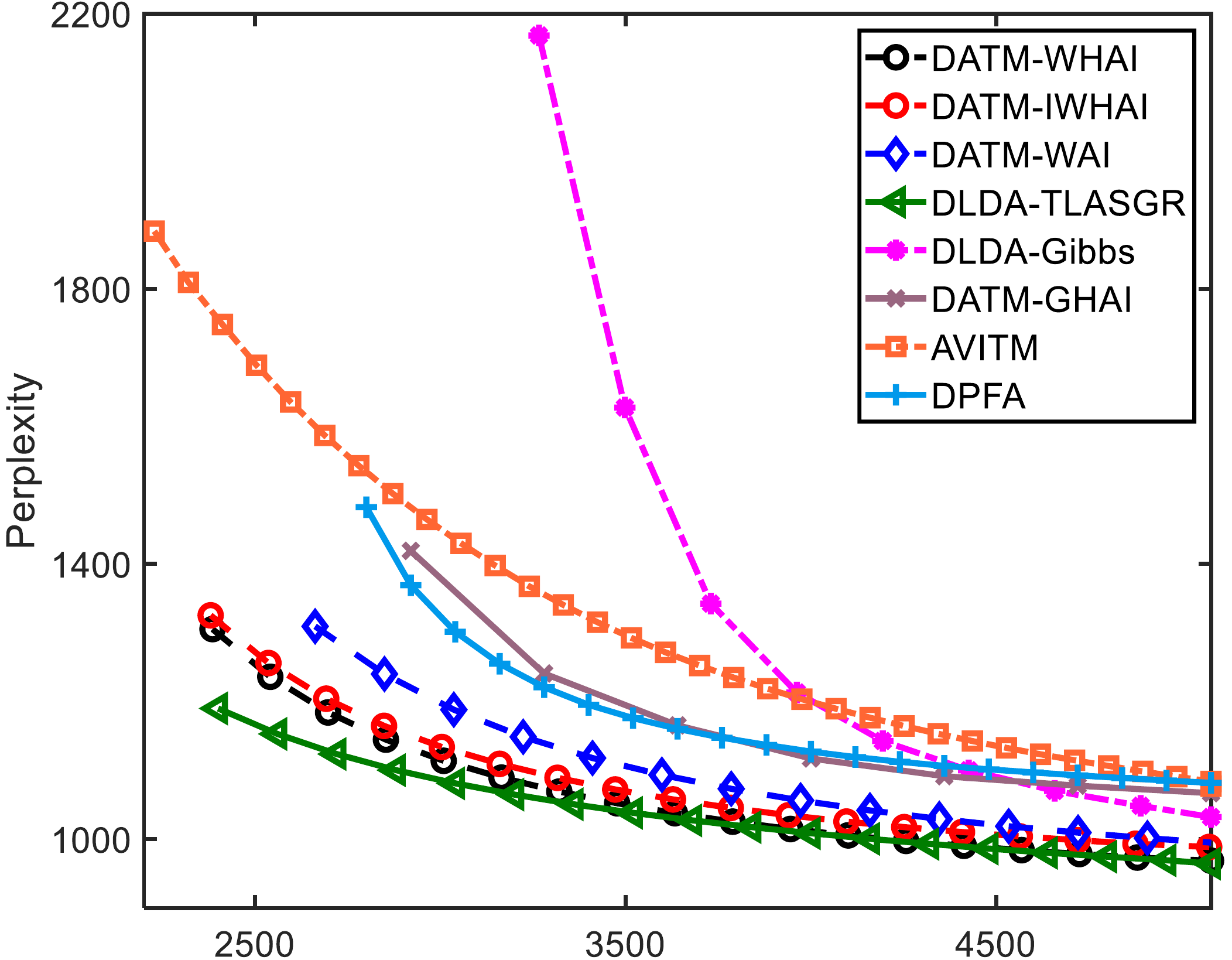}\label{RCV1_journal}}
	\quad
	\subfloat[]{\includegraphics[scale=0.17]{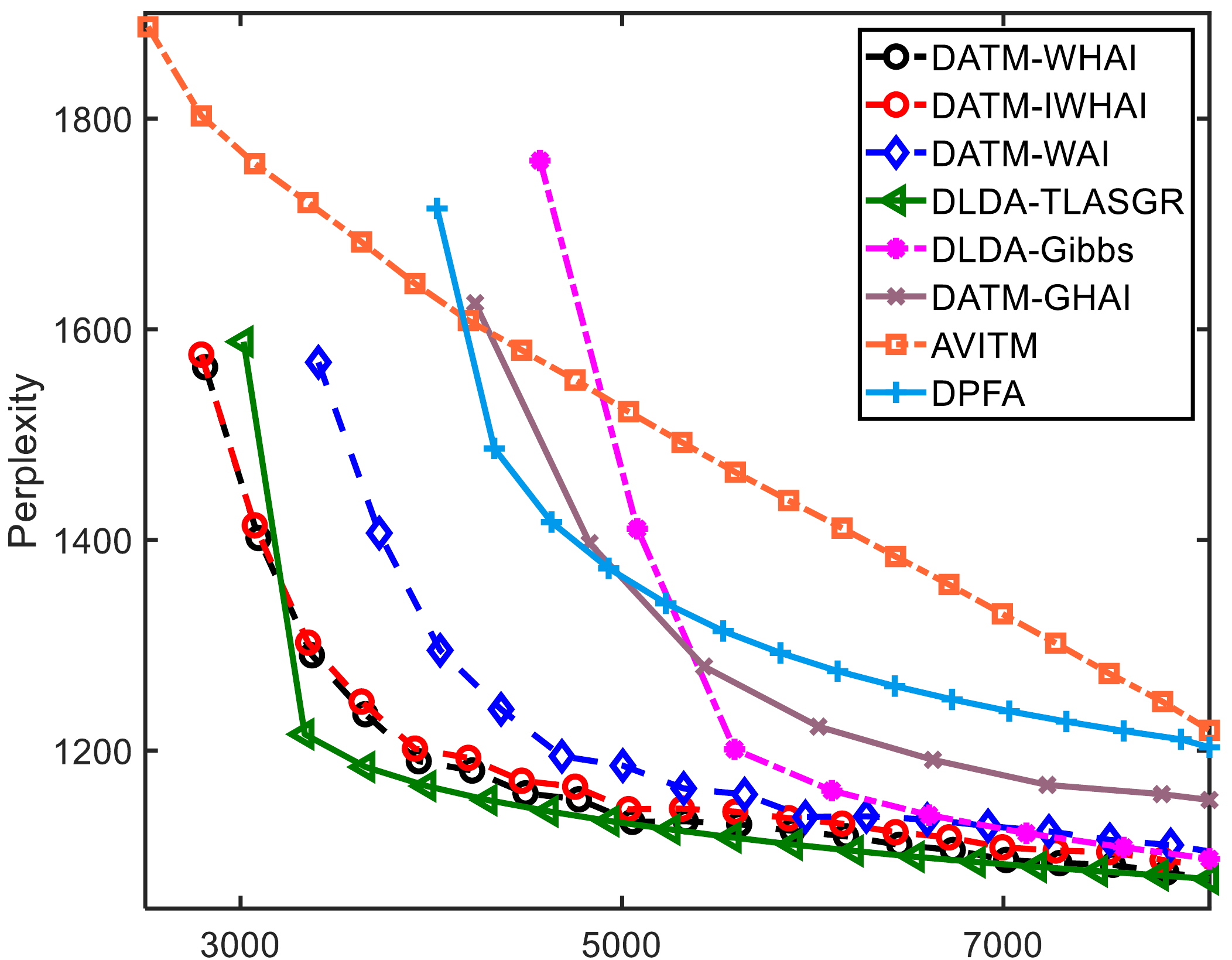}\label{WIKI_journal}}
	\caption{Plot of per-heldout-word perplexity as a function of time for (a) 20News, (b) RCV1, and (c) Wiki.  Except for AVITM that has a single hidden layer with 128 topics, all the other algorithms have the same network size of 128-64-32 for their deep generative models.}
	\label{fig:Train-time}
\end{figure*}

Similar to previous work \cite{wallach09,DILN,zhou2012beta}, for each corpus, we randomly select 70$\%$ of the word tokens from each document to form a training matrix $\Tmat$, holding out the remaining 30$\%$ to form a testing matrix $\Ymat$.
We use $\Tmat$  to train the model and calculate the per-heldout-word perplexity as
\begin{equation}\label{perplexity}
  \exp\left\{ -\frac{1}{y_{\cdotv \cdotv}} \sum \limits_{v=1}^V \sum \limits_{n=1}^{N} y_{vn} \ln \frac{\sum\nolimits_{s=1}^S \sum\nolimits_{k=1}^{K^{1}} \phi_{vk}^{(1)s} \theta_{kn}^{(1)s} }{\sum\nolimits_{s=1}^S \sum\nolimits_{v=1}^{V} \sum\nolimits_{k=1}^{K^{1}} \phi_{vk}^{(1)s} \theta_{kn}^{(1)s} }  \right\},
\end{equation}
where $S$ is the total number of collected samples and $y_{\cdotv \cdotv}=\sum\nolimits_{v=1}^V \sum\nolimits_{n=1}^{N} y_{vn}$.
For the proposed models, we set the mini-batch size as 200, and use 2000 mini-batches for burn-in on both 20News and RCV1 and 3500 on Wiki.
We collect 3000 samples after burn-in to calculate perplexity. The hyperparameters of WHAI are set as: $\eta^{(l)} = 1/K_l$, $\rv = \onev$, and $c_n^{(l)} = 1$.

Table \ref{Tab:perplexity and time} lists for various algorithms both the perplexity and the average run time per testing document given 3000 random samples of the global parameters.
For fair comparison, all the models are evaluated on the same 3.0 GHz CPU. We first compare different DLDA based models and then compare 
 the proposed DATM-WHAI with other non-DLDA based models.

Given the same generative network structure, DLDA-Gibbs performs the best in terms of predicting heldout word tokens, which is not surprising as this batch algorithm can sample from the true posteriors given a sufficiently large number of Gibbs sampling iterations.
DLDA-TLASGR is a mini-batch algorithm that is much more scalable in training than DLDA-Gibbs, at the expense of slightly  degraded performance in out-of-sample prediction.
Both DATM-WAI, using SGD to infer the global parameters, and DATM-WHAI, using a stochastic-gradient MCMC to infer the global parameters, slightly underperform DLDA-TLASGR.
Compared with DATM-GHAI approximately reparameterizing the gamma distributions, DATM-WHAI that has simple reparameterizations for its
Weibull distributions outperforms DATM-GHAI.
Besides, thanks to the use %
 TLASGR-MCMC rather than a simple SGD procedure, DATM-WHAI consistently outperforms  DATM-WAI.
It is also clear that except for DATM-IWHAI that has no stochastic-downward components in its inference, all the other variations of DATM %
have a clear trend of improvement as the generative network becomes deeper, indicating the importance of having stochastic downward information propagation during posterior inference.
Compared with DLDA-Gibbs and DLDA-TLASGR that need to perform Gibbs sampling at the testing stage, DATM-WHAI and its variations are considerably faster in processing a testing document, due to the use of an inference network.

Further, comparing DATM-WHAI with the methods in the first group in Table \ref{Tab:perplexity and time} shows that all algorithms with an inference network, including AVITM, DocNADE, and DATM-WHAI, clearly outperform those relying on an iterative procedure for out-of-sample prediction, including OR-softmax, LDA, and DPFA.
In terms of perplexity, it can be seen that DATM-WHAI with a single hidden layer already clearly outperforms AVITM, indicating that using the Weibull distribution is more appropriate than using the logistic normal distribution to model the document latent representation.
Compared with DocNADE, an outstanding autoregressive and shallow model, the single-layer DATM-WHAI with the same number of topics marginally improves the perplexity and test speed. 
Distinct from DocNADE, DATM-WHAI is able to add more stochastic hidden layers to extract hierarchical topic representations and further improve its perplexity, and its non auto-regressive structure makes it easier to be %
 accelerated with GPUs.

{As discussed in Section \ref{sec:layer-wise}, DATM is able to infer the network structure via a greedy layer-wise training strategy given a fixed budget on the width of the first layer.
We perform experiments with $L=3$, $K_{1 max} \in \{ 64,128,256,512,1024 \}$, and the pruning threshold as $u=0.01$. %
Shown in Table \ref{Tab:inferred structure} are the inferred network structure and perplexities over three different corpora.
We observe a clear trend of improvement by increasing $K_{1 max}$ until saturation (when $K_{1 max}$ becomes sufficiently large).
Moreover, when $K_{1 max}=128$, in comparison to the results of a fixed 128-64-32 network structure shown in Table \ref{Tab:perplexity and time}, we find that a better network structure with lower perplexity is inferred, illustrating the effectiveness of our proposed method.

}

Below we examine how various inference algorithms progress over time  during training, evaluated with per-holdout-word perplexity.
As clearly shown in~Fig. \ref{fig:Train-time},  DATM-WHAI outperforms DPFA and AVITM in providing lower perplexity as time progresses, which is not surprising as the DLDA multi-layer generative model is good at document representation, while AVITM is only ``deep'' in the deterministic part of its inference network
and DPFA is restricted to model binary topic usage patterns via its deep network.
 When DLDA is used as the generative model, in comparison to Gibbs sampling and TLASGR-MCMC on two large corpora, RCV1 and Wiki, the mini-batch based WHAI converges slightly slower than TLASGR-MCMC but much faster than Gibbs sampling;
WHAI consistently  outperforms WAI, which demonstrates the advantage of our proposed hybrid Bayesian inference algorithm;
in addition, the RSVI based DATM-GHAI clearly converges more slowly in time than DATM-WHAI does.
Note that for all three datasets, the perplexity of TLASGR decreases at a fast rate, followed closely by that of WHAI,
while that of Gibbs sampling decreases slowly, especially for RCV1 and Wiki, as shown in Figs. \ref{RCV1_journal} and \ref{WIKI_journal}.
This is expected as both RCV1 and Wiki are much larger corpora, for which a mini-batch based inference algorithm can already make significant progress in inferring the global model parameters, before a batch-learning Gibbs sampler finishes a single iteration that needs to go through all documents.
We also notice that although AVITM is fast for testing via the use of a VAE, its representation power is limited due to not only  the use of a shallow topic model, but also the use of a latent Gaussian based inference network that is not naturally suited  to model document latent representation.

\subsubsection{Topic hierarchy}
In addition to quantitative evaluations, we have also visually inspected the inferred topics at different layers and the inferred connection weights between the topics of adjacent layers.
Distinct from many existing deep topic models that build nonlinearity via ``black-box'' NNs,
we can easily visualize the whole  stochastic network, whose hidden units of layer $l-1$ and those of  layer $l$  are connected by $\phi_{k'k}^{(l)}$ that are sparse.
In particular, we can understand the meaning of each hidden unit by projecting it back to the original data space via $\left[ \prod_{t=1}^{l-1} \Phimat^{(t)} \right] \phiv_k^{(l)}$, as described in Section \ref{sec2A}.
 We show in Fig. \ref{fig:topic_wiki} a subnetwork, originating from Topics (units) 16, 19, and 24 of the top hidden layer, taken from the generative network of size 128-64-32 inferred on Wiki. 
Note plotting  the whole network at once is often unrealistic and hence we resort to extracting a subnetwork for visualization. The reason that these three topics are combined as the roots to form the subnetwork shown in Fig. \ref{fig:topic_wiki} is because they share similar key words and appear somewhat related to each other. Both the semantic meanings of the inferred topics and the connection weights  between them are highly interpretable.  These topics tend to be very specific at the bottom layer, and become increasingly more general at higher layers.
Note that the higher-layer topics gather the general semantics from their leaf nodes, leading to the fact that some words may appear with large weights in several  different higher-layer topics.
For example, in Fig. \ref{fig:topic_wiki},  both Topics 16 and 19 at layer 3 talk about ``international/group/company,''  %
but Topic 16 pays more attention to ``business'' while Topic 19 focuses more on ``organization.''
Several additional example topic subnetworks rooted at different top-layer nodes are shown in Figs. 1, 2, 3, and 6 in the Supplement. %
Moreover, comparisons of the hierarchical structures learned by hLDA \cite{griffiths2004hierarchical}, DEF \cite{def}, and DATM on 20News and NIPS12\footnote{http://www.cs.nyu.edu/\~{}roweis/data.html} are provided in the Supplement, which clearly demonstrate the unique hierarchical topic structure and its interpretability under the proposed model. %
More discussions can be found in the Supplement.

\begin{figure}
  \centering
  \subfloat[]{\includegraphics[scale=0.21]{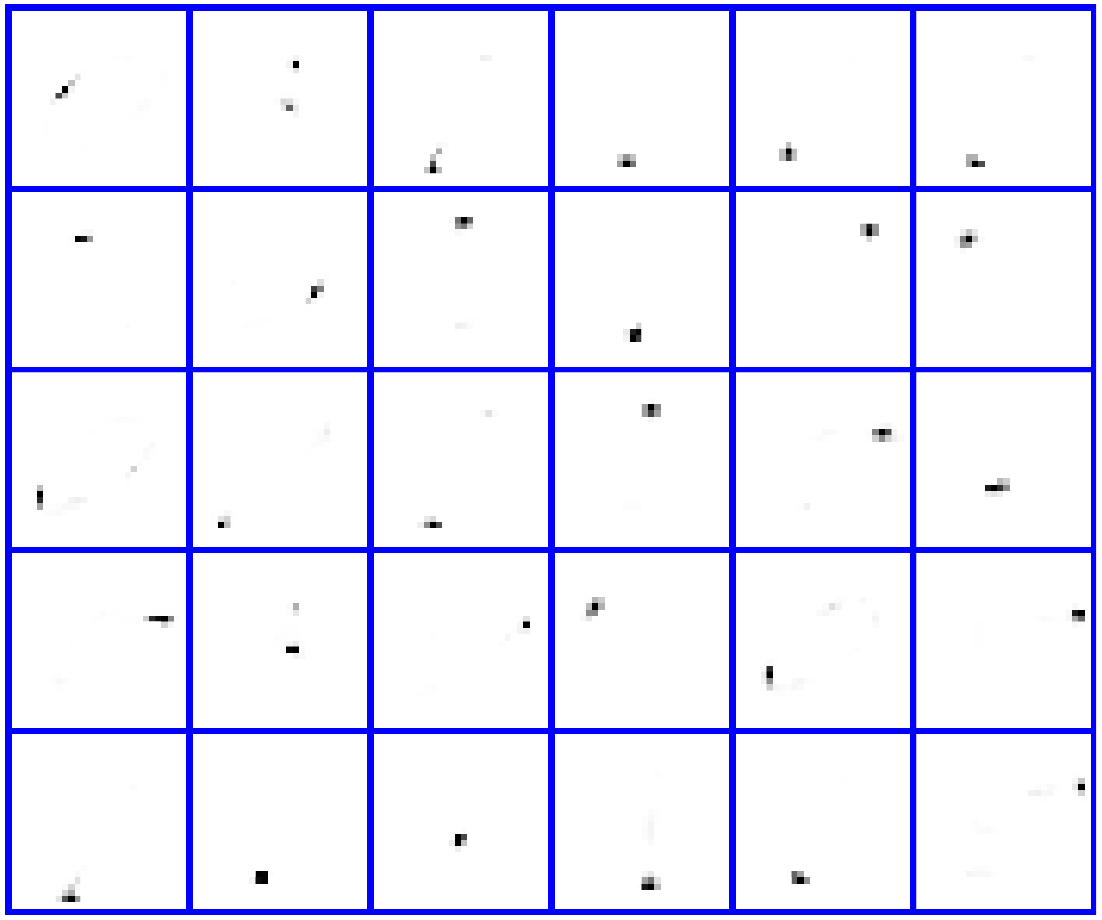}\label{fig:topic_MNISTa}} $\quad$
  \subfloat[]{\includegraphics[scale=0.21]{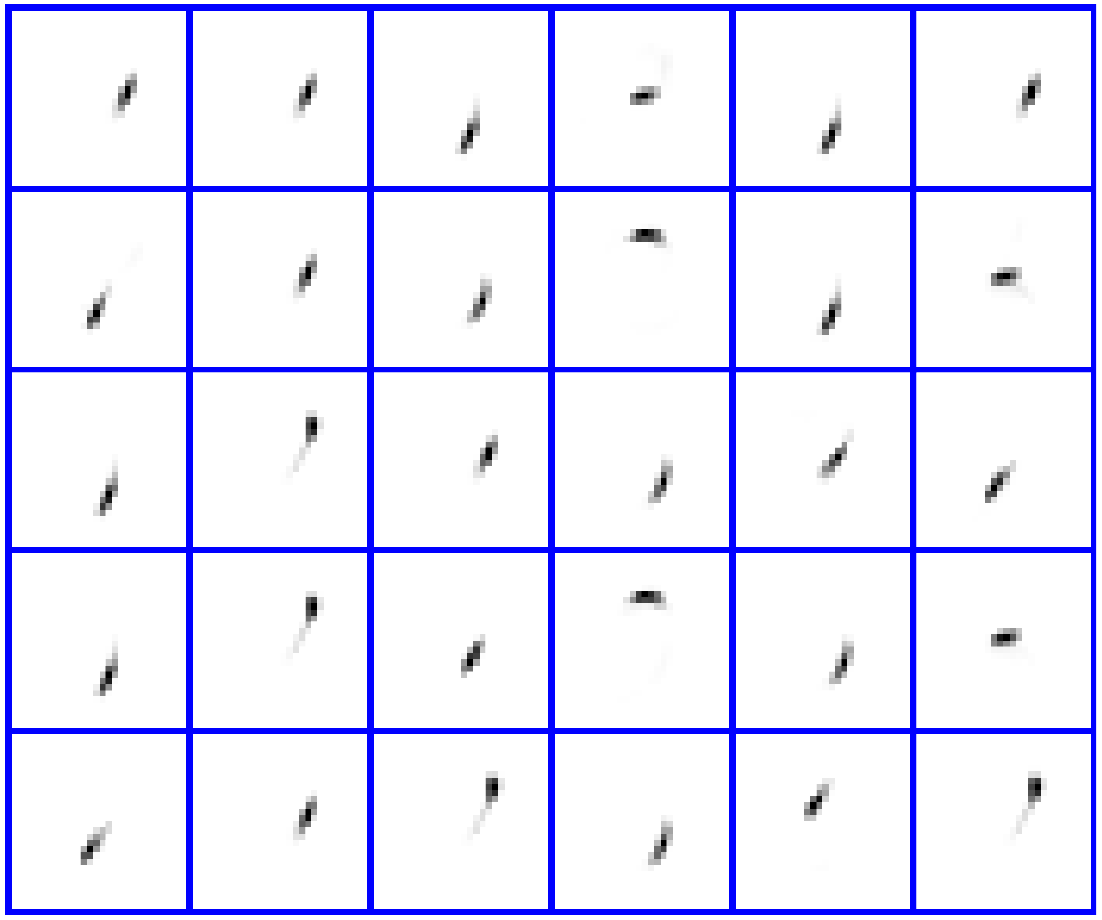}\label{fig:topic_MNISTb}} $\quad$
  \subfloat[]{\includegraphics[scale=0.21]{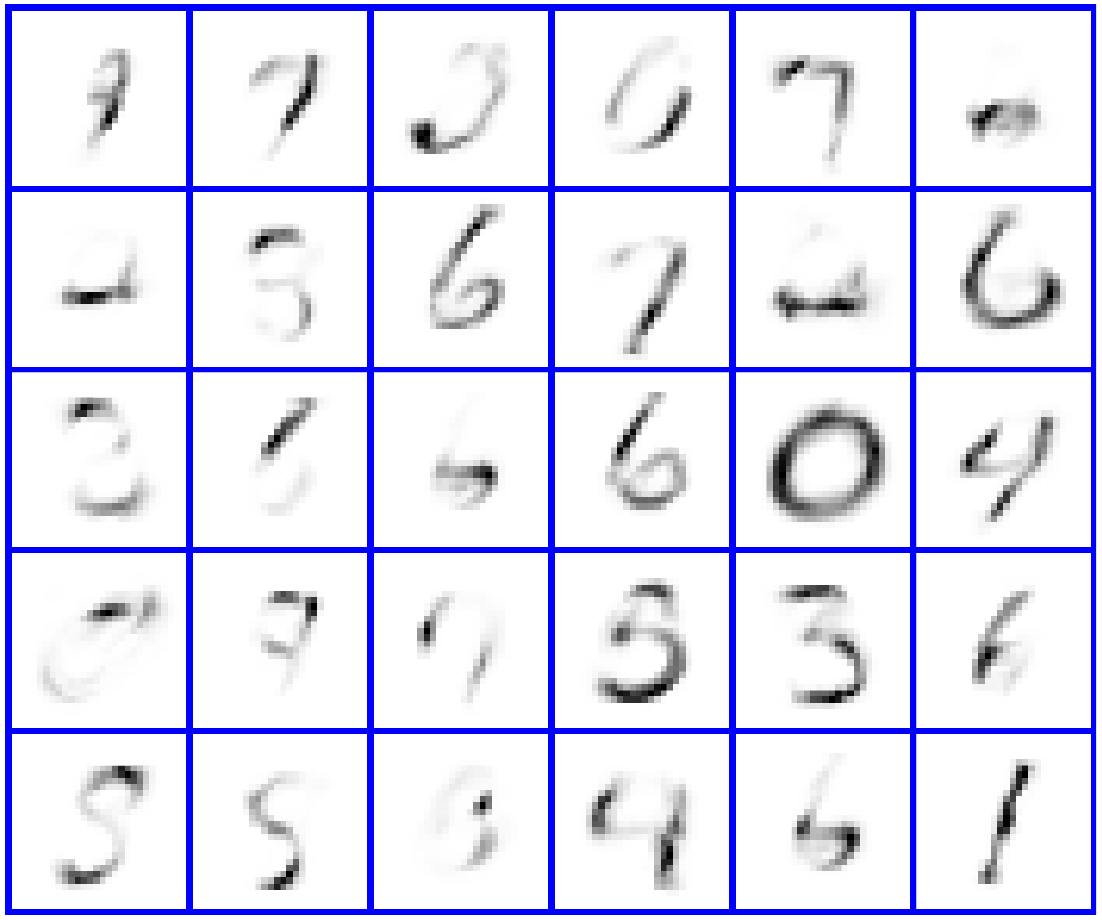}\label{fig:topic_MNISTc}} \\
  \subfloat[]{\includegraphics[scale=0.21]{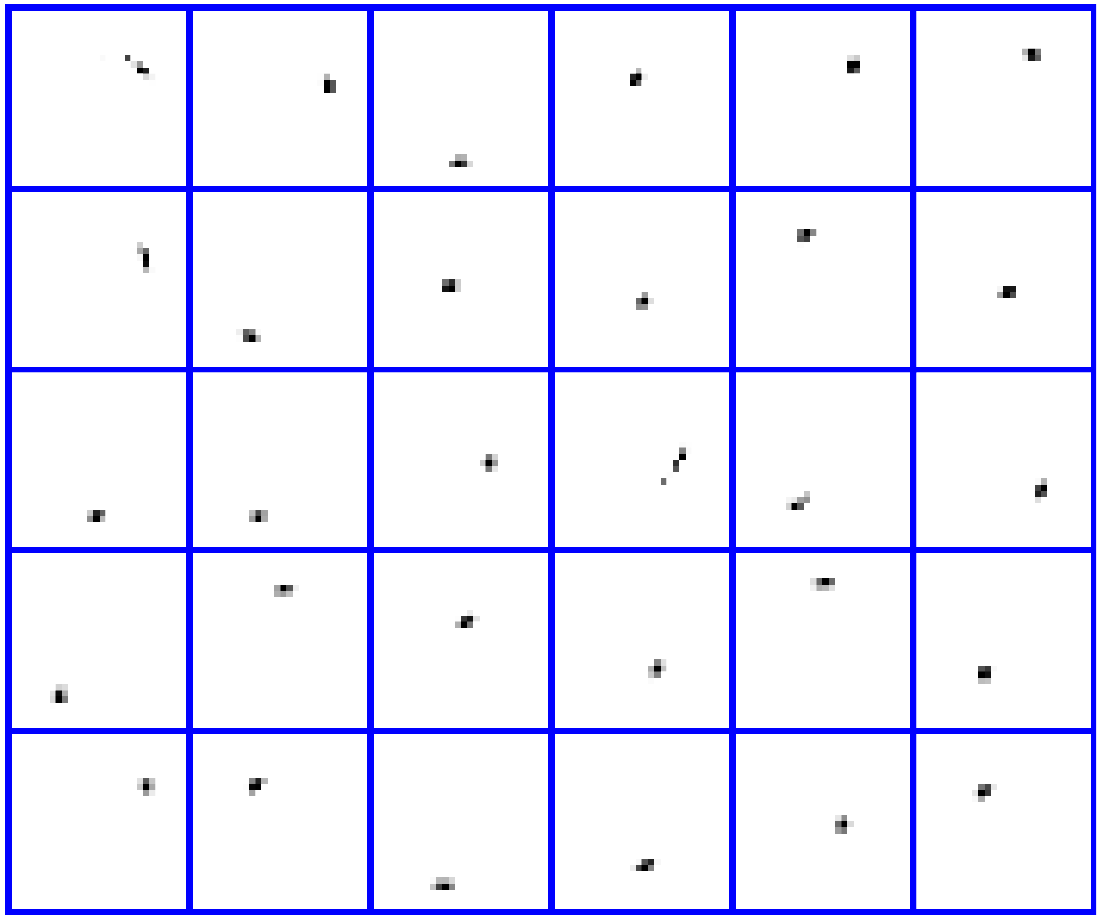}\label{fig:topic_MNISTd}} $\quad$
  \subfloat[]{\includegraphics[scale=0.21]{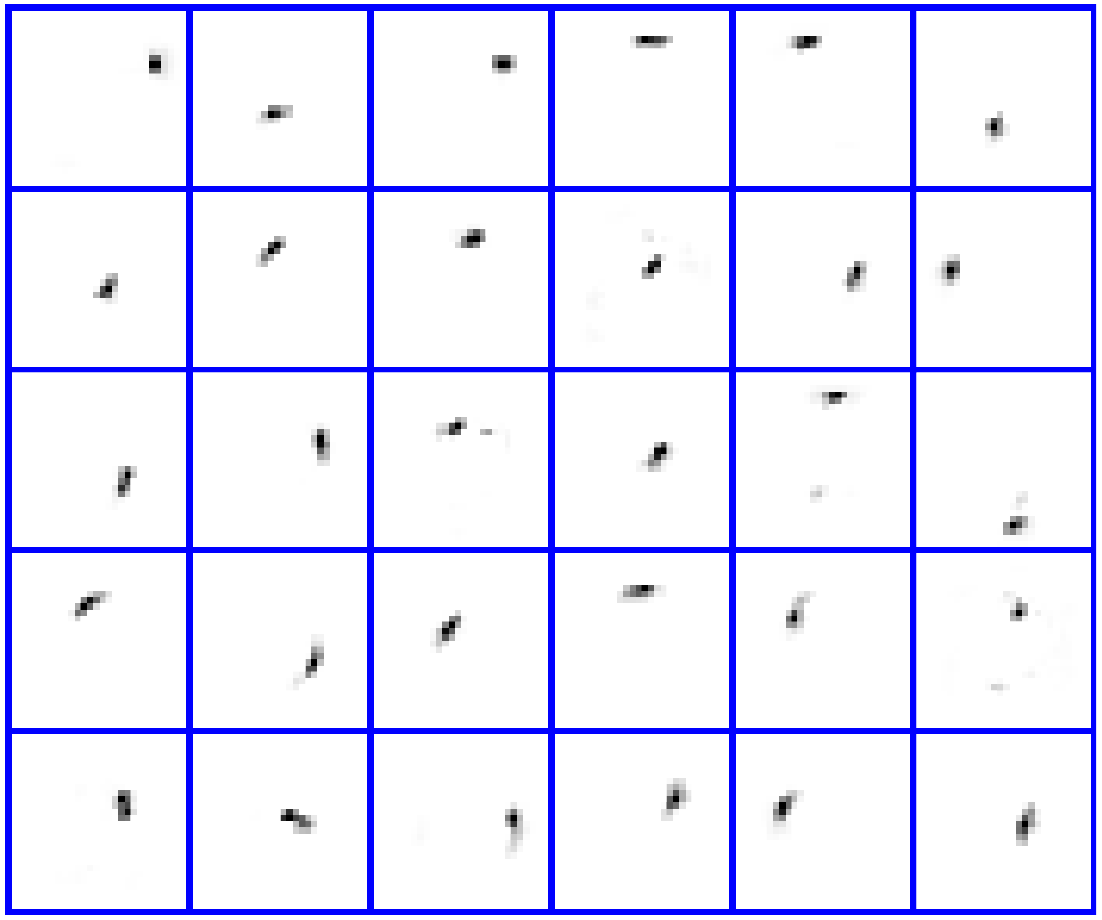}\label{fig:topic_MNISTe}} $\quad$
  \subfloat[]{\includegraphics[scale=0.21]{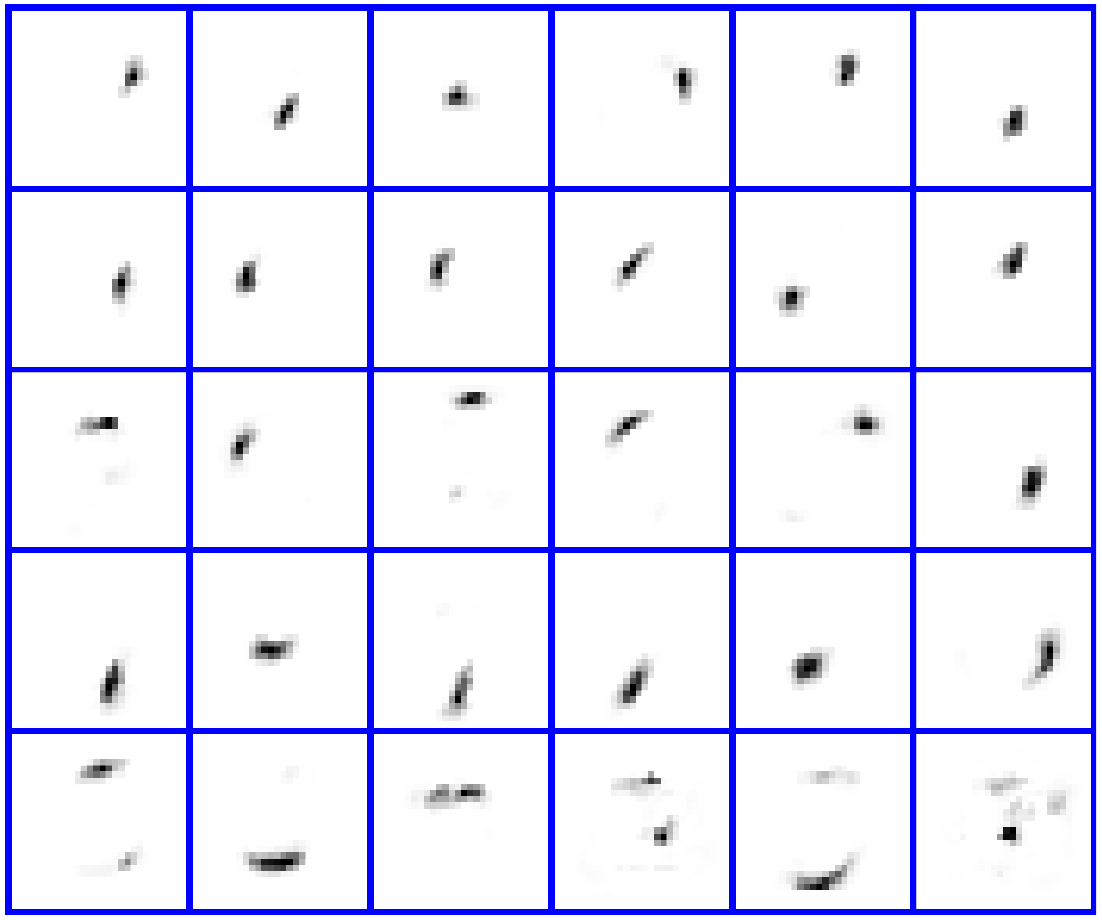}\label{fig:topic_MNISTf}}
  \caption{Learned topics on MNIST digits with a three-hidden-layer DATM of size 128-64-32. Shown in (a)-(c) are example topics for layers 1, 2 and 3, respectively, learned with a deterministic-upward-stochastic-downward encoder (DATM-WHAI), and shown in (d)-(f) are the ones learned with a deterministic-upward encoder (DATM-IWHAI).}
  \label{fig:topic_MNIST}
\end{figure}

\begin{figure*}[ht]
  \centering
  \includegraphics[scale=0.16]{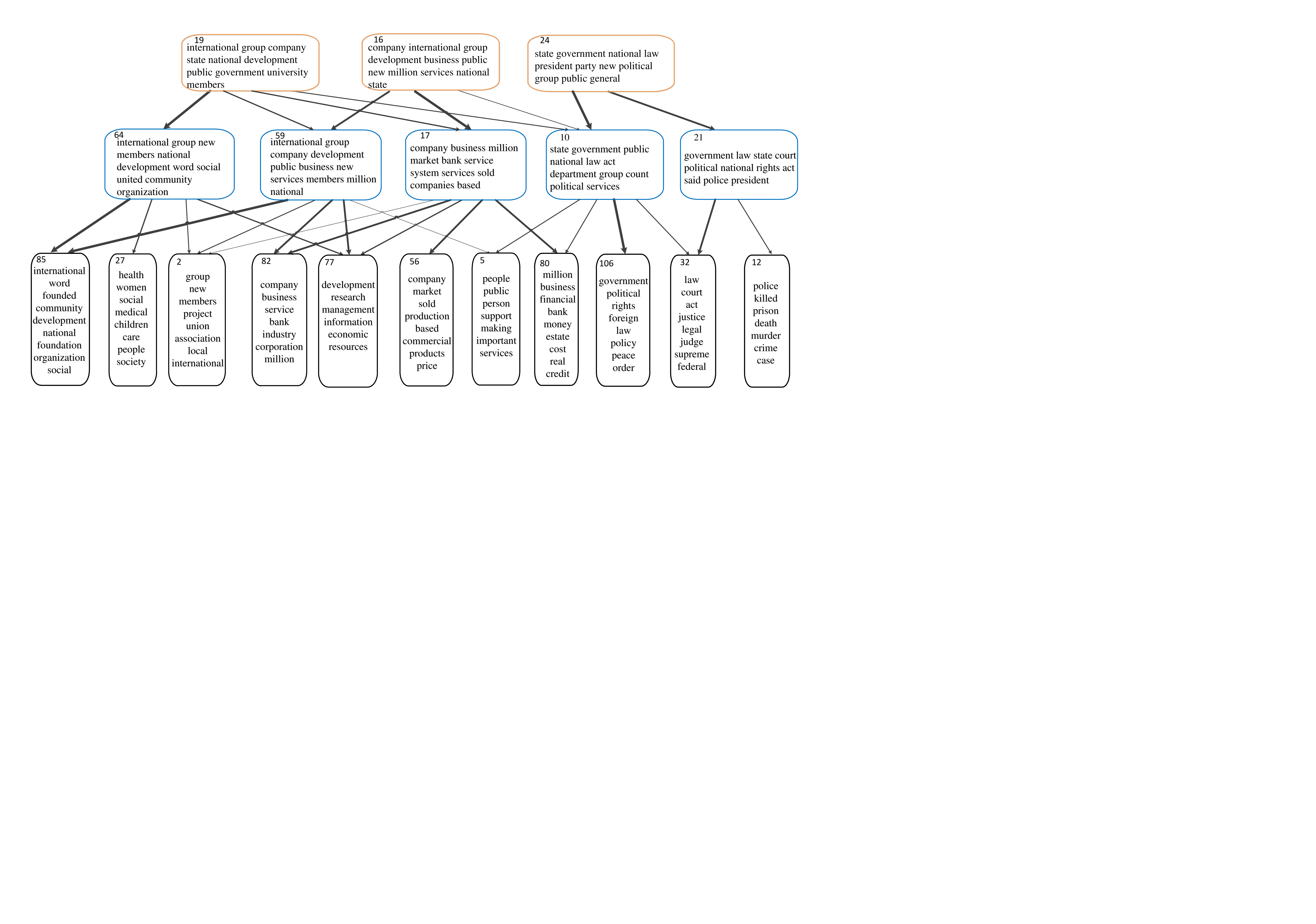}
  \caption{Example of hierarchical topics learned from Wiki by a three-hidden-layer DATM-WHAI of size 128-64-32.}
  \label{fig:topic_wiki}\vspace{-4mm}
\end{figure*}

To further illustrate the effectiveness of the multi-layer representation in our model, we apply a three-layer DATM to MNIST digits and present the learned dictionary atoms.
We use the Poisson likelihood directly to model the MNIST digit pixel values that  are nonnegative integers ranging from 0 to 255. As shown in Figs. \ref{fig:topic_MNISTa}-\ref{fig:topic_MNISTc}, it is clear that the factors at layers one to three represent localized points, strokes, and digit components, respectively, that cover increasingly larger spatial regions.
This type of hierarchical visual representation is difficult to achieve with other types of deep NNs \cite{srivastava2013modeling,kingma2014stochastic,rezende2014stochastic,sonderby2016ladder}.
WUDVE, the inference network of WHAI, has a deterministic-upward--stochastic-downward structure, in contrast to a conventional VAE that often has a pure deterministic bottom-up structure.
Here, we further visualize the importance of the stochastic-downward part of WUDVE through a simple experiment.
We remove the stochastic-downward part of WUDVE in \eqref{posterior-completely} represented as DATM-IWHAI, in other words, we ignore the top-down information.
As shown in Figs. \ref{fig:topic_MNISTd}-\ref{fig:topic_MNISTf}, although some latent structures are learned, the hierarchical relationships between adjacent layers almost all disappear, indicating the importance of having a stochastic-downward structure together with a deterministic-upward
one in the inference network.

\begin{figure}[!t]
	\centering
	\subfloat[]{\includegraphics[scale=0.16]{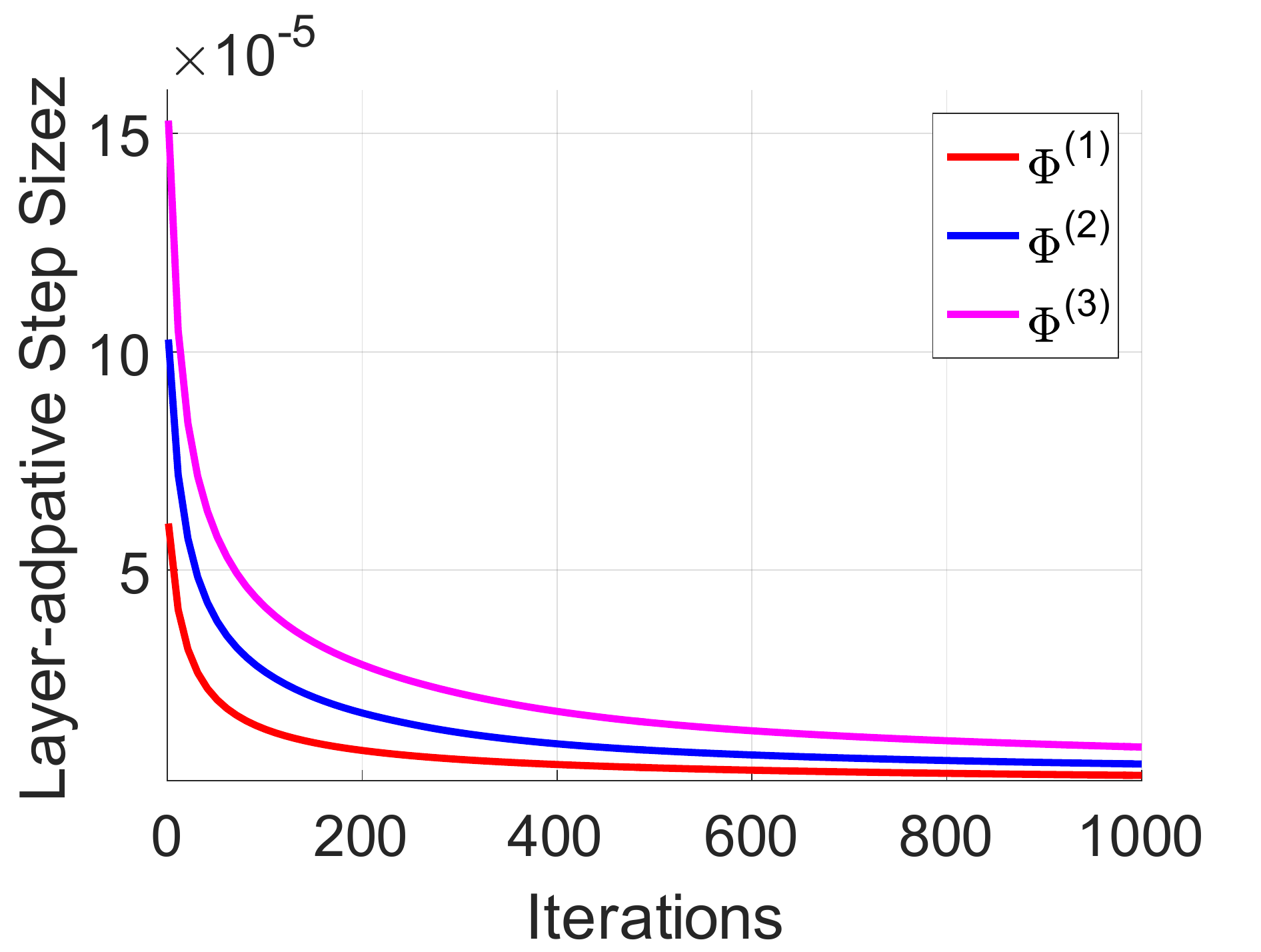}}
	\subfloat[]{\includegraphics[scale=0.16]{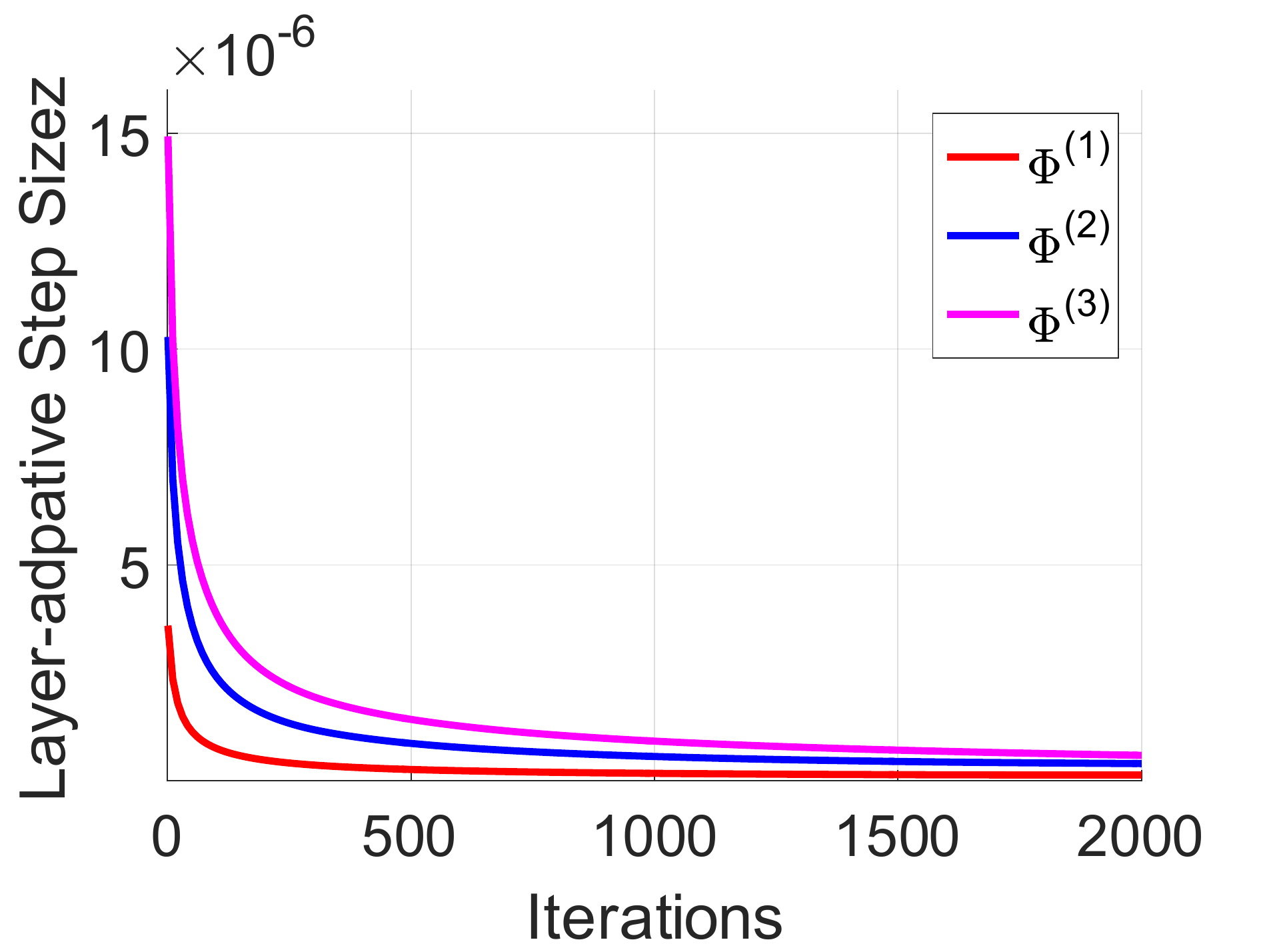}}
	\subfloat[]{\includegraphics[scale=0.16]{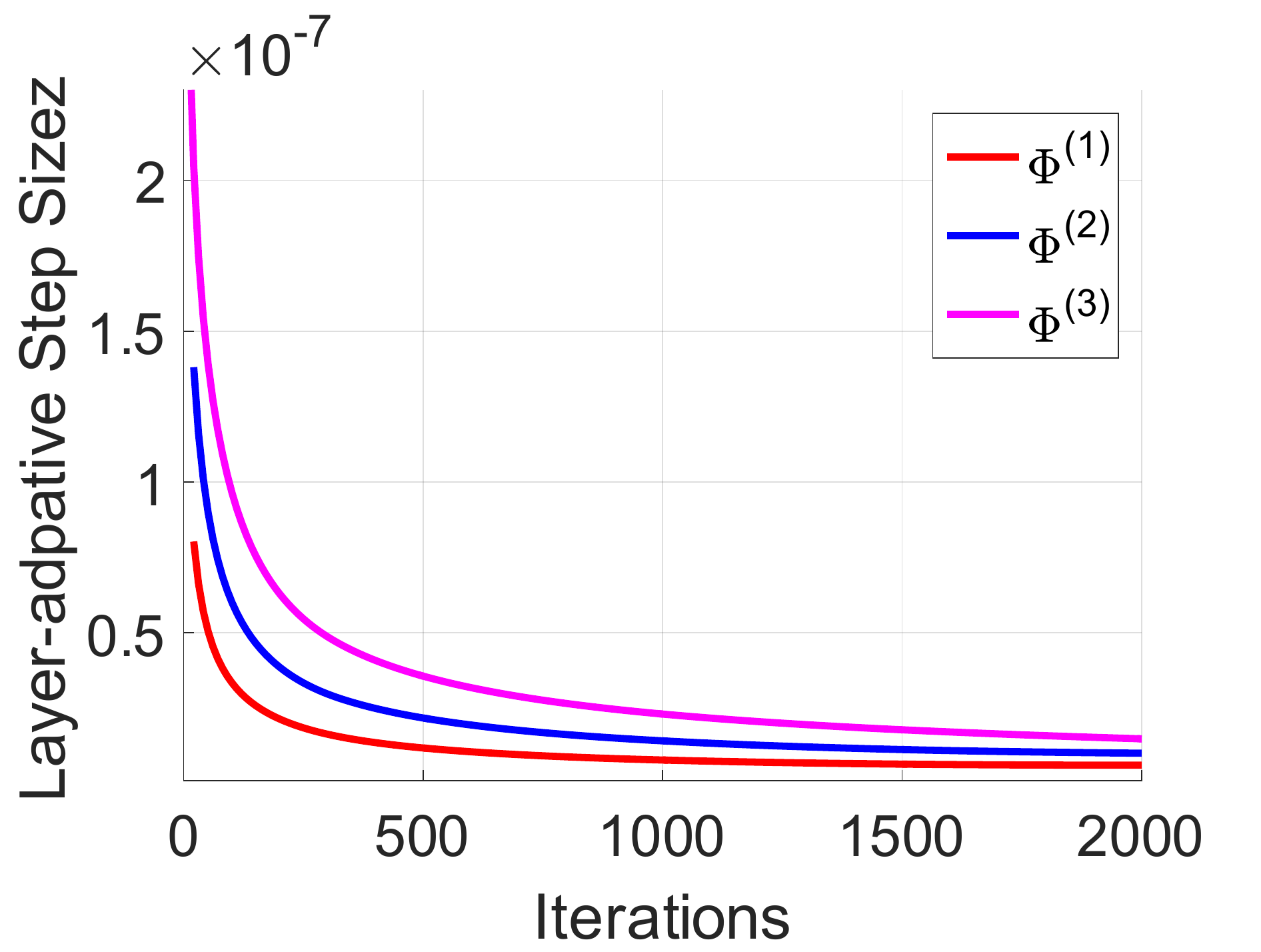}}
	\caption{Topic-layer-adaptive learning rates inferred with a three-layer DATM of size 128-64-32. (a) 20News. (b) RCV1. (c) Wiki}
	\label{fig:adpative stepsize}\vspace{-2mm}
\end{figure}

\subsubsection{Topic-layer-adapative stepsize}
To illustrate the working mechanism of our proposed topic-layer-adapative stepsize, we show how its inferred learning rates are adapted to different layers in Fig. \ref{fig:adpative stepsize}, which is obtained by averaging over the learning rates of all $\phiv_k^{(l)}$ for $k=1,\cdots,K_l$.
For $\Phimat^{(l)}$, higher layers prefer larger step sizes, which may be attributed to the enlarge-partition-augment data generating mechanism of DLDA.
In addition, we find that larger datasets prefer slower learning rates, which demonstrates that since the stochastic noise brought by minibatch learning increases, the model needs a smaller learning rate.

\subsubsection{Topic manifold}
As a sanity check for whether DATM %
overfits the data, we show in Fig. \ref{fig:manifold_MNIST} the latent space interpolations between the test set examples on MNIST dataset, and provide related results  in the Supplement for the 20News corpus.
In Fig. \ref{fig:manifold_MNIST}, the leftmost column is from the same image represented as $\xv_1$ and the rightmost column is random sampled from a class represented as $\xv_2$.
With the 3-layer model learned before, following Dumoulin et al. \cite{dumoulin2017adversarially},  $\xv_1$ and $\xv_2$ are projected into $\zv_1^{(3)}$ and $\zv_2^{(3)}$.
We then linearly interpolate between $\zv_1^{(3)}$ and $\zv_2^{(3)}$, and pass the intermediate points through the generative model to generate the input-space interpolations, shown  in the middle columns.
We observe smooth transitions between the examples in all pairs, and the intermediate images remain interpretable. These observations suggest that 
the inferred latent space of the model resides on a manifold %
and
WHAI has learned a generalizable  latent feature representation rather than concentrating its probability mass  around the training examples.
\begin{figure}[t]
  \centering
  \subfloat[]{\includegraphics[scale=0.9]{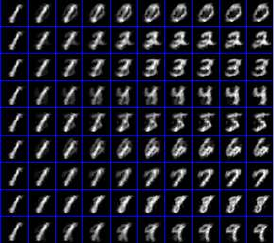}} $\quad$
  \subfloat[]{\includegraphics[scale=0.9]{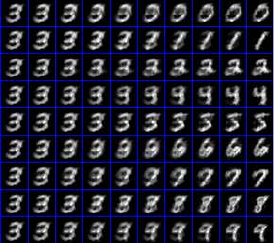}}$\quad$
  \subfloat[]{\includegraphics[scale=0.9]{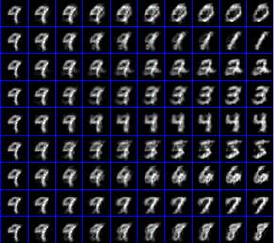}}
  \caption{Latent space interpolations on the MNIST test set. Left and right columns correspond to the images generated from $\zv_1^{(3)}$ and $\zv_2^{(3)}$, and the others are generated from the latent representations interpolated linearly from $\zv_1^{(3)}$ to $\zv_2^{(3)}$.}
  \label{fig:manifold_MNIST}\vspace{-2mm}
\end{figure}

\subsection{Supervised feature learning for classification}
To evaluate how well sDATM
 leverages the label information for feature learning, we compare its classification performance with a variety of algorithms on both MNIST digits and  several benchmark datasets for text classification.
For all experiments, the first 100 epochs are used to train DATM without the label information, %
and then another 300 epochs are used to train sDPATM with the label information.
Note as the ELBO in \eqref{ELBO-of-classification} contains several KL regularization terms, following Sonderby et al.  \cite{sonderby2016ladder}, %
warm-up is used during the first several epochs  to gradually impose the KL regularization terms.

{\bf{Digit classification.}}
We first test sDATM on the MNIST dataset,  which consists of 60,000 training handwritten digits and 10,000 testing ones.
We list the results in Table \ref{Tab:MNIST_class}.
Among them, DLDA inferred by Gibbs sampling and Gaussin VAE inferred by SGD are the unsupervised feature learning models, which are followed by a linear SVM, represented as DLDA+SVM and VAE+SVM, respectively.
Other models, including a supervised VAE model under Gaussian assumption called max-margin variational autoencoder (MMVA) \cite{li2015max-margin}, a supervised generative adversarial network called  Adversarial Autoencoders (AAE) \cite{AAE}, a fully-connected NN (FNN) \cite{li2015preconditioned-pSGLD}, and the deep belief network (DBN) \cite{Hinton06}, are also compared.
In addition, except for sDATM and DLDA that use the original gray-scale pixel values from 0 to 255, all the other algorithms divide them by 255, normalizing them to nonnegative real values from 0 to 1. %

 \begin{table}[t]
\centering
\caption{Error rates ($\%$) and testing time (average seconds per image) on MNIST dataset. %
}
\begin{tabular}{c|c|c}
\hline
{Model} & Error Rate & Test Time  \\ \hline\hline
DLDA+SVM \cite{zhou2015poisson} &  2.82 & 0.523 \\
VAE+SVM \cite{li2015max-margin} &  1.04 & 0.081 \\ \hline
DBN \cite{Hinton06} &  1.20 & 0.021 \\
FNN \cite{li2015preconditioned-pSGLD} &  1.14 & 0.013 \\
MMVA \cite{li2015max-margin} &  0.90 & 0.014 \\
AAE \cite{AAE} &  0.85 & 0.015 \\ \hline
sDATM-L &  1.03 & 0.011 \\
sDATM-N &  0.97 & 0.013 \\ \hline
\end{tabular}\label{Tab:MNIST_class}
\end{table}

Shown in Table \ref{Tab:MNIST_class} are the error rates of various algorithms, which are provided in their corresponding papers, except for that of DLDA+SVM which is obtained by running the author provided  code;
the test times are obtained by running the author provided code on the same  computer with a 3.0 GHz CPU.
Since both sDATM-L and sDATM-N extract features and learn classifier jointly by a principled fully-generative model, it is unsurprising that their test errors are clearly lower than that of DLDA+SVM.
Meanwhile, the nonlinear $\text{sDATM-N}$ only  slightly outperforms  the linear  sDATM-L, demonstrating the effectiveness of sDATM in transforming the data into a discriminative latent space.
In addition, thanks to the encoder network in sDATM, it takes substantial less time than DLDA+SVM does at the testing stage. %
In contrast to both DBN and FNN that use deterministic ``black-box'' deep NNs, sDATM learns an interpretable multi-stochastic-layer  latent space and provides a lower testing error.
MMVA and AAE perform slightly  better but takes longer time at the testing stage than sDATM does,  which may be attributed to more complex networks used in them. %

Shown in Fig. \ref{MNIST_classification} are how  the test errors of sDATM change as the layer width and network depth vary.
There is a clear trend of test error reduction as the network depth of sDATM increases, suggesting  the effectiveness of having a multi-layer representation and feature fusion.
When the network depth is fixed, sDATM with a larger network width performs better, suggesting sDATM is able to use a larger capacity network  to learn a more discriminative latent space.

\begin{figure}[t]
  \centering
  \subfloat[]{\includegraphics[scale=0.15]{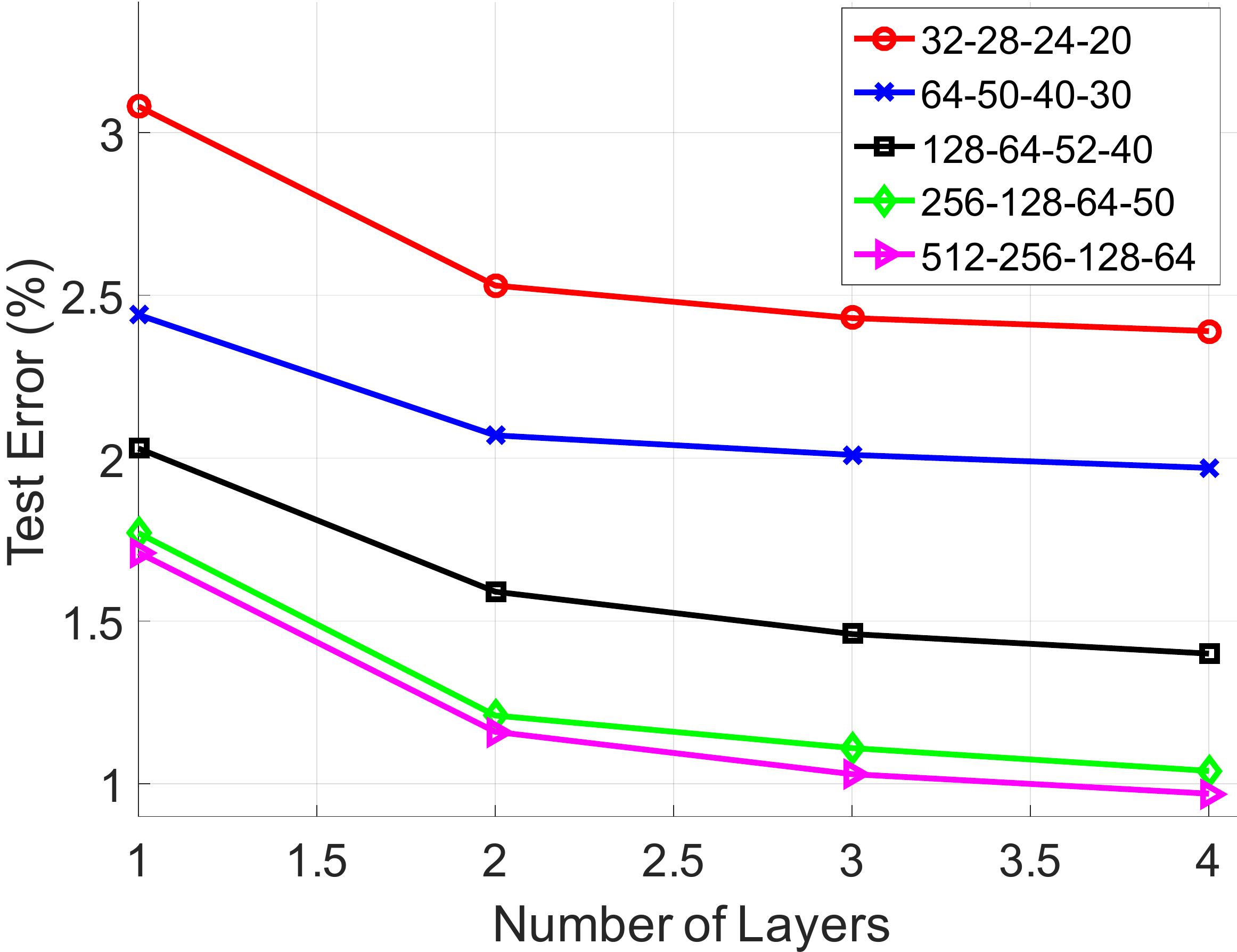}\label{MNIST_classification}}$\quad$
  \quad
  \subfloat[]{\includegraphics[scale=0.15]{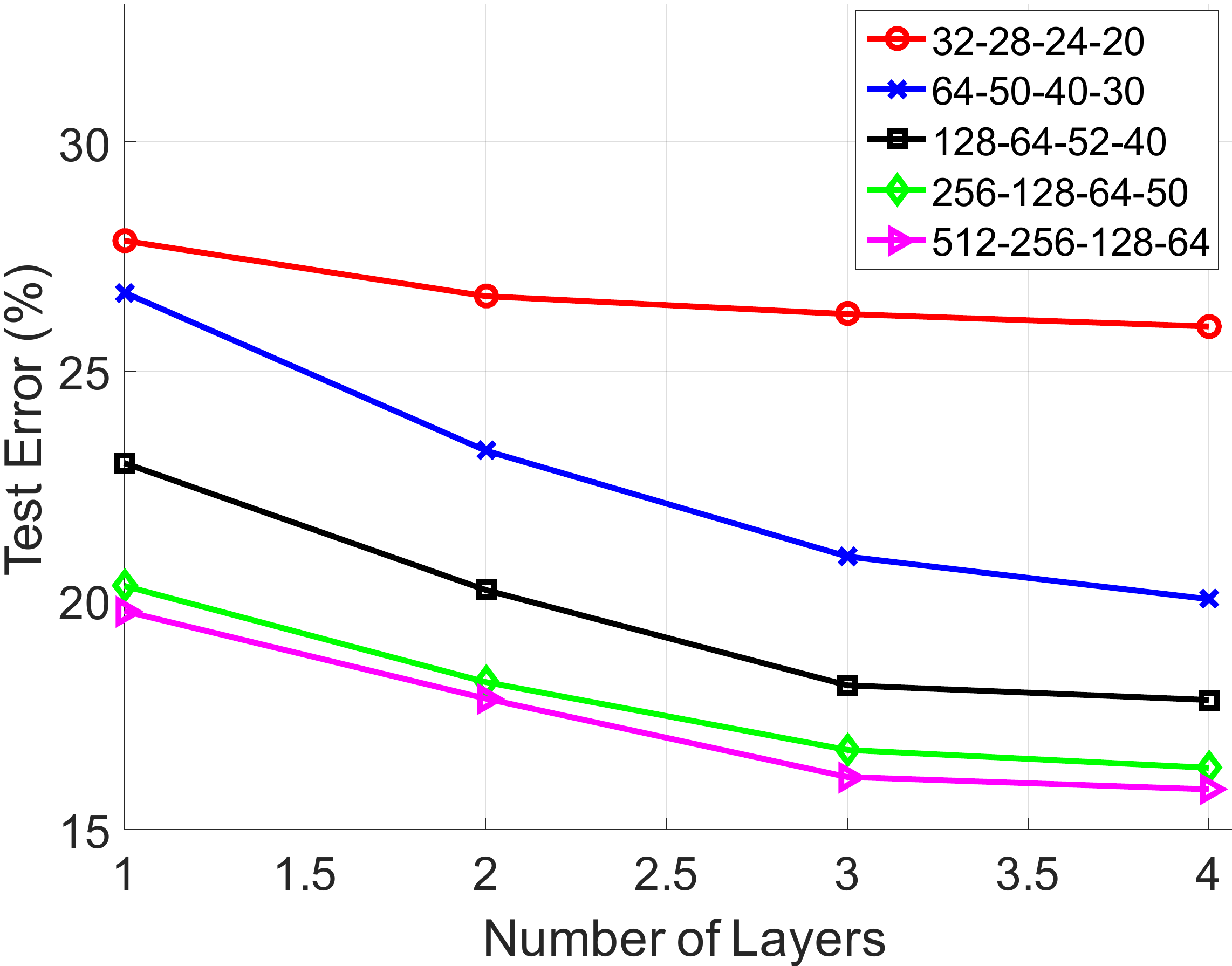}\label{20News_classification}}
  \caption{The test errors change with different layers and width on (a) MNIST and (b) 20News by sDATM-N.}\label{classification}
\end{figure}

{\bf{Document classification.}}
We also test sDATM on the following three document classification tasks: 20News, RCV1, and IMDB \cite{maas2011learning}.
Different from the perplexity experiments,  a lager vocabulary of 33,420 words is used for 20News to achieve better performance \cite{GBN}.
Since RCV1 has 103  topic categories in a hierarchy and one document may be associated with more than one topic, we transform them to a single-label classification with 55 different classes as discussed in Johnson \& Zhang \cite{johnson2015effective,johnson2016supervised}.
The IMDB dataset is used for sentiment classification on movie reviews with a vocabulary size of 30,000 after preprocessing.
The task is to determine whether the movie reviews are positive or negative.
For a fair comparison, the training/testing random splits follow the same settings in previous work \cite{GBN,johnson2015effective,johnson2016supervised}.

Besides a number of  unsupervised models, including LDA, DLDA, DocNADE, OR-softmax, and AVITM, we also make comparison to several representative supervised models:

\begin{itemize}
\item {\bf{FNN-BOW}} and {\bf{FNN-tfidf}}: Four-layer fully-connect feedforward  NNs (FNN) of size $512$-$256$-$128$-$64$ with bag-of-words (BOW) and tfidf document features.
\item {\bf{MedLDA}}: Gibbs max-margin supervised topic models \cite{zhu2014gibbs} based on LDA.
  \item {\bf{sAVITM}}: A supervised AVITM through adding a linear  softmax classifier to AVITM.
  \item {\bf{wv-LSTM}}: A LSTM model based on word vector sequence \cite{dai2015semi-supervised}.
\end{itemize}
For DLDA and sDATM, we choose a four-layer structure with the size of $512$-$256$-$128$-$64$.

\begin{table}[t]
\centering
\caption{Test error rates ($\%$) and testing time (average seconds per document) on 20News, RCV1, and IMDB datasets.}
\begin{tabular}{c|c|c|c|c|c|c}
\hline
\mr{2}*{Model} & \mc{3}{c|}{Error Rate} & \mc{3}{c}{Test Time} \\ \cline{2-7} 
                          & {20News} & {RCV1} & {IMDB} & {20News} & {RCV1} & {IMDB}  \\ \hline\hline
LDA & 25.40 & 24.17 &  21.46 & 0.60 & 0.25 & 0.49 \\
DLDA & 22.01 & 20.18 &  18.13 & 1.22 & 0.92 & 1.06 \\
DocNADE & 23.21 & 21.07 &  18.79 & 0.031 & 0.019 & 0.024 \\
OR-softmax & 22.05 & 20.19 &  18.24 & 0.69 & 0.21 & 0.58 \\
AVITM & 26.31 &  25.16 & 21.75 &  0.017 & 0.014 & 0.015 \\
DATM & 24.28 &  23.10 & 20.42 &  0.018 & 0.015 & 0.016 \\
 \hline
FNN-BOW & 31.29 &  28.16 & 19.25 &  0.014 & 0.010 & 0.012 \\
FNN-tfidf & 24.16 &  19.28 & 17.14 &  0.013 & 0.009 & 0.010 \\
sAVITM & 20.15 &  18.61 & 16.13 & 0.015 & 0.012 & 0.014 \\
MedLDA & 18.76 &  16.38 & 15.28 & 0.240 & 0.098 & 0.202 \\
wv-LSTM & 18.00 & 16.04 & 13.50 & - & - & - \\
sDATM-L & 18.63 & 15.42 & 13.66 & 0.016 & 0.013 & 0.014 \\
sDATM-N & 15.81 & 13.40 & 10.92 & 0.018 & 0.014 & 0.015\\ \hline
\end{tabular}\label{Tab:RCV1+IMDB}
\end{table}

Listed in Table \ref{Tab:RCV1+IMDB} are the results for various algorithms, where these of wv-LSTM are provided in Dai \& Le \cite{dai2015semi-supervised}, these for FNN-bow and FNN-tfidf are obtained by our own carefully optimized code, %
and all the others are obtained by running the author provided code, all on the same data used by sDATM.
Although DLDA  achieves the lowest  testing errors among all unsupervised models, which illustrates the effectiveness of the multi-layer representation of DLDA, it underperforms all supervised topic models.
Having the same document features (BOW) and similar encoder structures, supervised FNN-BOW underperforms DLDA, indicating  that the interpretable latent feature space of DLDA is amenable to classification.
Both sDATM-L and sDATM-N,  which integrate feature extraction and  classification via a fully generative model for both documents and labels, clearly outperform DLDA.
sDATM-L outperforms both sAVITM and MedLDA, which suggests that its multilayer latent representation provides more discriminative power. sDATM-N further improves over $\text{sDATM-L}$ by introducing nonlinearity into the mapping from the latent features to labels.
Note LSTM is a popular method to model the sequence information between words in a document.
The proposed sDATM-L and sDATM-N, only using BOW features, achieve comparable or better performance in comparison to wv-LSTM.
In Fig. \ref{20News_classification}, we also show on the 20News dataset how the test errors of sDATM-N change as the network depth and width  vary.

{\bf{Supervised topic modeling.}}
It can be noted from \eqref{ELBO-of-classification} that the topics and latent representations of sDATM are related to not only   the documents but also their labels, which is why sDATM is able to provide more discriminative power than DATM.
For illustration, we compare the top five first-layer topics  learned by DATM and  that by sDATM-L on the IMDB dataset in Fig.~\ref{sDATM_topic}.  The documents in IMDB %
contain both  movie descriptions and viewer comments.
Clearly, the top topics inferred by DTAM are focused on the content of the movies, while these by   sDATM are focused on the sentiments of the viewers, which helps explain why sDATM performs much better than DATM in document classification for IMDB.

\begin{figure}[t]
  \centering
  \includegraphics[scale=0.15]{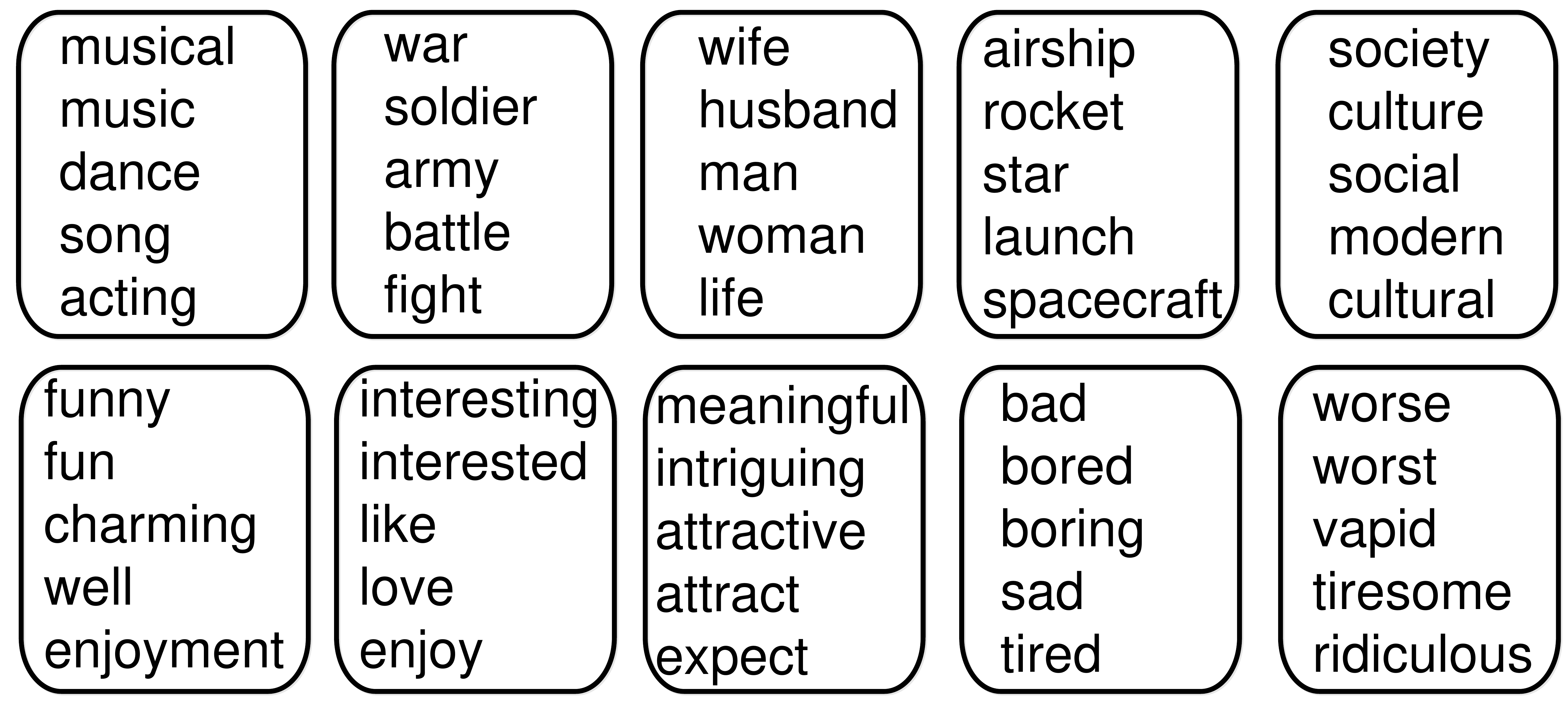}\\
  \caption{The top five first-layer topics learned  by DATM (the first row) and those by sDATM (the second row).}\label{sDATM_topic}
\end{figure}

{%
In order to better understand the changes of topics from DATM to sDATM, we first train DATM with 100 epochs and then add label information to train sDATM  with another 100 epochs.
Figs. \ref{IMDB_dictionary_unsupervised} and \ref{IMDB_dictionary_supervised} show how one topic tree changes by changing the model from DATM to sDATM.
Clearly,  some connection weights between the topics of adjacent layers change and some topics become more focused on viewer sentiments after adding the label likelihood.

}

\begin{figure}[t]
  \centering
  \includegraphics[scale=0.32]{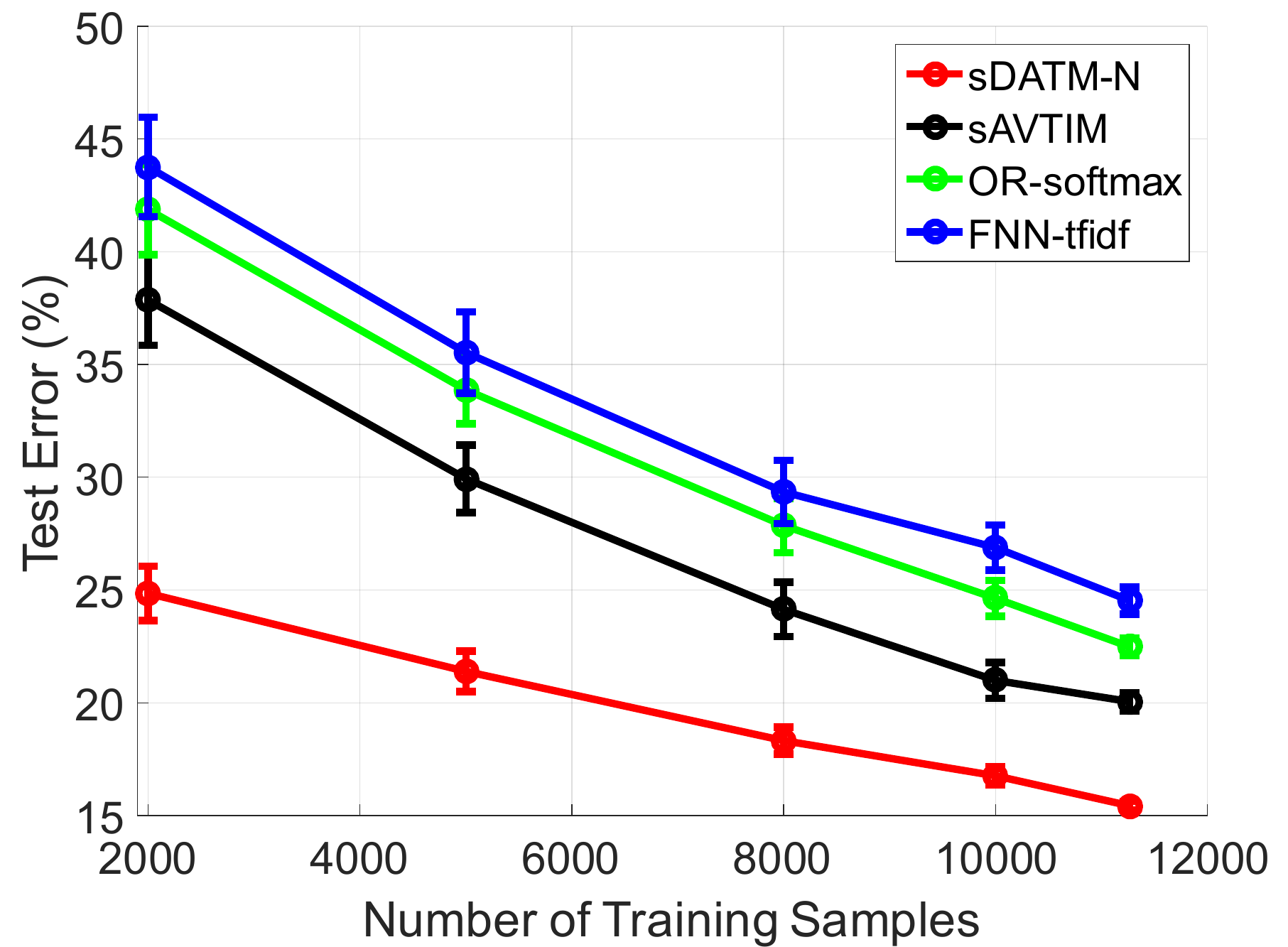}\\
  \caption{The test errors change with different sizes of training dataset on 20News. }\label{20News_reduce_samples}
\end{figure}

\begin{figure*}[th]
  \centering
  \includegraphics[scale=0.2]{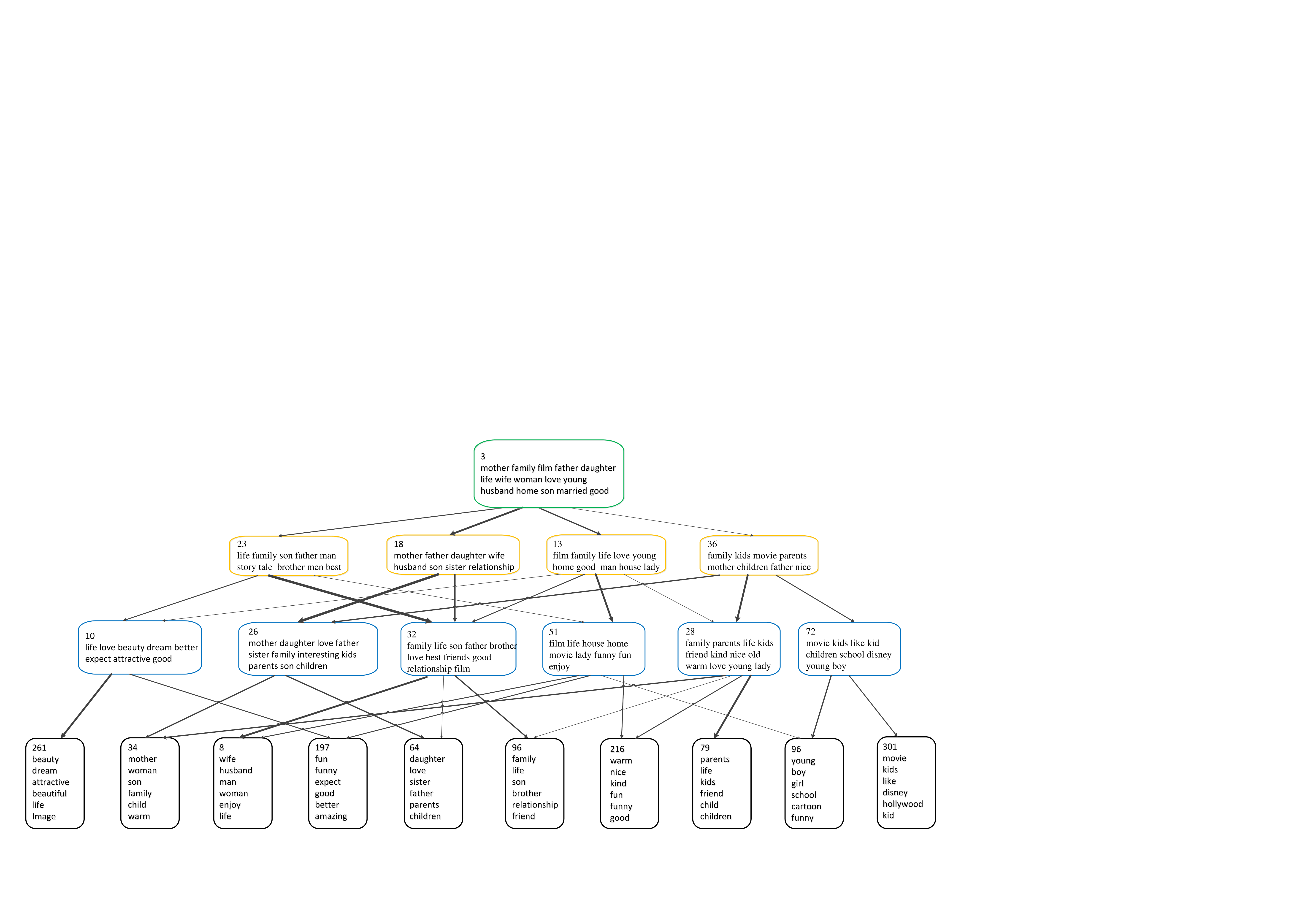}\\
  \caption{Topics by unsupervised learning. }\label{IMDB_dictionary_unsupervised}
\end{figure*}

\begin{figure*}[th]
  \centering
  \includegraphics[scale=0.2]{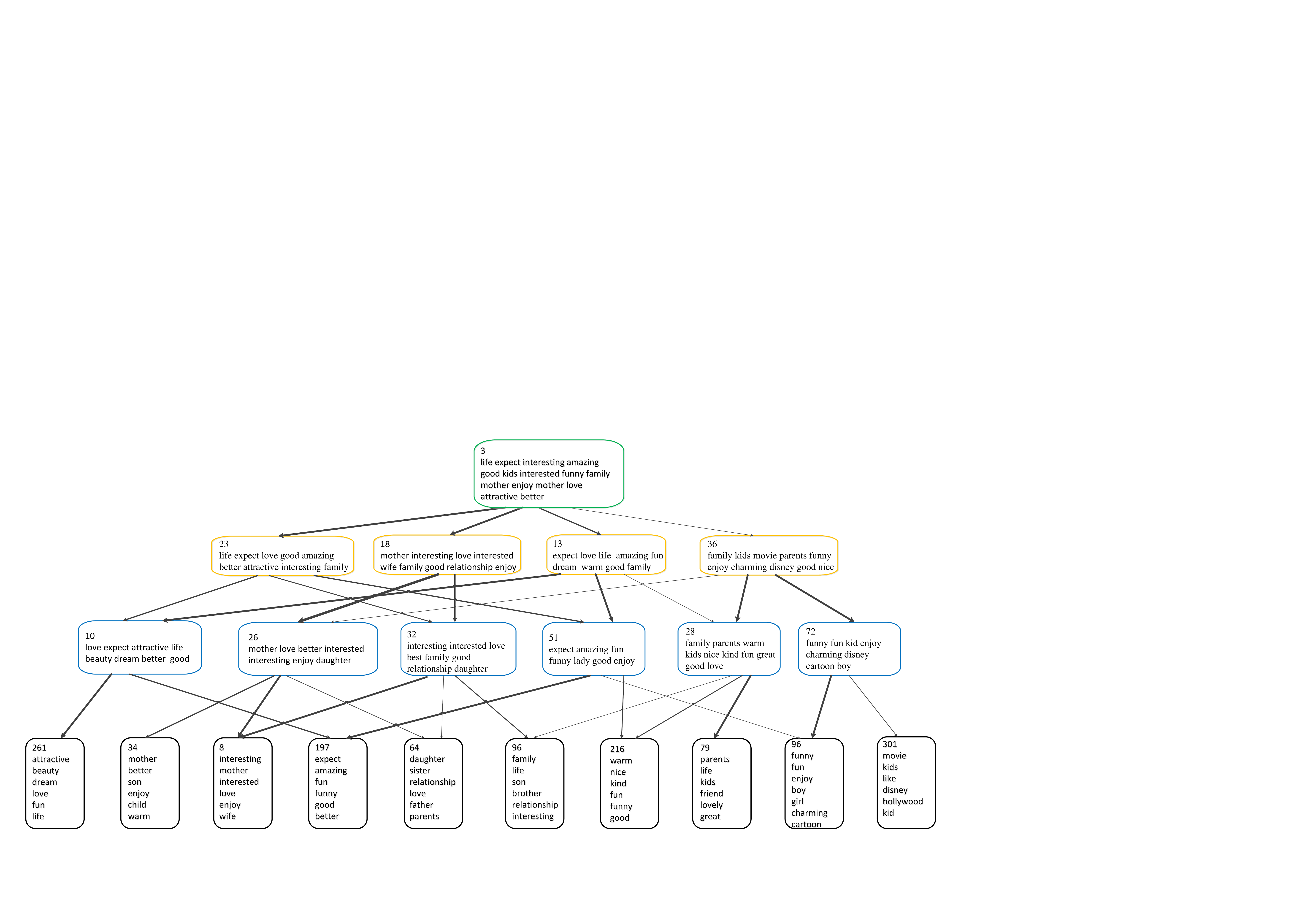}\\
  \caption{Topics by supervised learning. }\label{IMDB_dictionary_supervised}
\end{figure*}

{\bf{Robust to the smaller training set.}}
Compared with the deterministic mapping in DNNs, sDATM constructs a probabilistic model to perform distribution estimation, which may bring more robustness to the smaller training set.
In order to demonstrate this advantage, we train the models on 20News with different training data sizes, as shown in Fig. \ref{20News_reduce_samples} achieved by 50 independent experiments.
Due to the relative small training dataset, deterministic FNN with tfidf document features performs worst among all the models.
The parameters in the generative model of sAVITM (the topics) and OR-softmax (the weights) are updated by SGD with a point estimate, which results in an obvious increase in test error.
In addition, the variance of sDATM is smaller than that of others especially when the training data is small.
We attribute it to the following three reasons: 1) the fully distribution estimate of sDATM, no matter for the networks' parameters or the classifiers; 2) the model average when it perform prediction; and 3) the robustness brought by the multi-layer feature fusion.

\section{Conclusion}
We propose an interpretable deep generative model for document analysis, referred to as deep autoencoding topic model (DATM), where deep latent Dirichlet allocation is used as the generative network, and a Weibull upward-downward variational encoder is used to approximate the posterior distribution of the latent representation. Scalable Bayesian inference for DATM is realized by  a hybrid stochastic gradient MCMC and variational inference algorithm. We further construct supervised DATM that can jointly model the documents and their labels.  The efficacy and scalability of the proposed models are demonstrated on a variety of unsupervised and supervised learning tasks with big corpora.

\bibliography{mybibTEX,References052016}
\bibliographystyle{IEEEtran}

\clearpage
\onecolumn 
\appendices
\section{Naive derivation of the Fisher information matrix of PGBN}
We have discussed that it is difficult to calculate the FIM for PGBN expressed in (1) because of the coupled relationships between layers.
In this section, we provide more detailed discussions.

For simplicity, we take for example a two-layer PGBN, expressed as
 \begin{align}\label{PGBN_generate}
  &\thetav_n^{(2)} \sim \mbox{Gam}\left(\rv,1/c_n^{(3)} \right), \nonumber \\
  &\thetav_n^{(1)} \sim \mbox{Gam}\left(\Phimat^{(2)} \thetav_n^{(2)} ,1/c_n^{(2)} \right), \nonumber\\
  & \xv_n \!\sim \!\mbox{Pois} \left(\Phimat^{(1)} \thetav_n^{(1)} \right),\!
\end{align}
and focus on a specific element $\Phimat_{vk}^{(2)}$ only.
With the FIM definition
\begin{equation}\label{FIM}
  G(\zv) = \mathbb{E}_{{\Pimat}|\zv} \left[ -\frac{\partial^{2}}{\partial \zv^2} \ln p({\Pimat}|\zv) \right],
\end{equation}
where $\Pimat$ denotes the set of all observed and local variables, and $\zv$ denotes the set of all global variables, one may show that the $\Phimat^{(2)}$-relevant part in $\ln p ({\Pimat}|\zv)$ is
\begin{equation}\label{likelihood_in_FIM}
  \sum_{vn} \left[ \Phimat_{v:}^{(2)} \thetav_{:n}^{(2)} \ln \left( c_n^{(2)} \thetav_{vn}^{(1)} \right) -\ln \Gamma \left( \Phimat_{v:}^{(2)} \thetav_{:n}^{(2)} \right) \right].
\end{equation}
Accordingly, with $\psi'(\cdot)$ denoted as the the trigamma function, for $\Phimat_{vk}^{(2)}$ we have
\begin{equation}\label{FIM_for_phi2}
  \mathbb{E} \left[ -\frac{\partial^2}{\partial [\Phimat_{vk}^{(2)}]^2 } \ln p ({\Pimat}|\zv) \right] = \mathbb{E} \left[ \sum_n \psi' \left( \Phimat_{v:}^{(2)} \thetav_{:n}^{(2)} \right) \left[ \thetav_{:n}^{(2)} \right]^2 \right],
\end{equation}
which is an expectation that is difficult to evaluate. 

\section{Proof of DLDA expression}
Note that the counts in $x_{vn}^{(l)} \sim \mbox{Pois}(q_j^{(l)} \sum_{k=1}^{K} \phi_{vk}^{(l)} \theta_{kn}^{(l)})$ can be augmented as
\begin{align}\label{augmented_x1}
  x_{vn}^{(l)} &= \sum_{k=1}^{K} x_{vkn}^{(l)}, \nonumber \\
  x_{vkn}^{(l)} &\sim \mbox{Pois} \left( q_j^{(l)} \phi_{vk}^{(l)} \theta_{kn}^{(l)} \right),
\end{align}
which, according to Lemma 4.1 of \cite{zhou2012beta}, can be equivalently expressed as
\begin{align}\label{augmented_x2}
  \left( x_{vkn}^{(l)} \right)_v &\sim \mbox{Mult} \left( m_{kn}^{(l)(l+1)}, \phiv_k^{(l)} \right), \nonumber \\
  m_{kn}^{(l)(l+1)} &\sim \mbox{Pois} \left( q_j^{(l)} \theta_{kn}^{(l)}  \right),
\end{align}
where $m_{kn}^{(l)(l+1)} := \sum_{v=1}^{K_{l-1}} x_{vkn}^{(l)}$.
Marginalizing out $\theta_{vn}^{(l)} \sim \mbox{Gam} \left( \sum_{k=1}^{K_{l+1}} \phiv_{vk}^{(l+1)} \thetav_{kn}^{(l+1)}, 1/c_n^{(l+1)} \right)$ from \eqref{augmented_x2} leads to
\begin{equation}\label{m}
  m_{vj}^{(l)(l+1)} \sim \mbox{NB} \left( \sum_{k=1}^{K_{l+1}} \phiv_{vk}^{(l+1)} \thetav_{kn}^{(l+1)}, p_j^{(l+1)} \right),
\end{equation}
which can be augmented as
\begin{align}\label{augmented_x3}
   m_{vj}^{(l)(l+1)} &\sim \mbox{SumLog} \left( x_{vn}^{(l+1)}, p_n^{(l+1)} \right), \nonumber \\
   x_{vj}^{(l+1)} &\sim \mbox{Pois} \left( q_n^{(l+1)} \sum_{k=1}^{K_{l+1}} \phiv_{vk}^{(l+1)} \thetav_{kn}^{(l+1)} \right).
\end{align}

\section{Derivation of the $\Gamma (\cdot)$ functions in SG-MCMC}
With $\Dmat(\zv) = \Gmat(\zv)^{-1}$, $\Qmat(\zv) = \zerov$, and the block-diagonal Fisher information matrix (FIM) $\Gmat(\zv)$ in~(6), it is straightforward to show that $\frac{\partial}{\partial \varphiv_k}\left[ \Dmat(\zv) + \Qmat(\zv) \right]$ is non-zero only in the $\varphiv_k$-related block $\Imat(\varphiv_k)$ in~(7).
Therefore, we focus on this block and have
\begin{equation}\label{Gamma}
  \Gamma_v(\varphiv_k) = \sum_u \frac{\partial}{\partial \varphi_{uk}} \left[ \Imat_{vu}^{-1} (\varphiv_k) \right],
\end{equation}
where $\Imat_{vu}^{-1} (\varphiv_k) = M_k^{-1} \left[ \mbox{diag} (\varphiv) - \varphiv \varphiv^T \right]$. Accordingly, we have
\begin{align}\label{Gamma2}
  \Gamma_v(\varphiv_k) = & M_k^{-1} \sum_u \frac{\partial}{\partial \varphi_{uk}} [\delta_{u=v} \varphi_{uk} - \varphi_{vk}\varphi_{uk}]  \nonumber \\
  & = M_k^{-1} (1-V \varphi_{vk}).
\end{align}

Since $\Gmat(\zv)$ is a block-diagonal with its $\rv$-relevant block being $\Imat(\rv) = M^{(L+1)} \mbox{diag}(1 / \rv)$, according to the definition of $\Gamma (\cdot)$, it is straightforward to show that
\begin{align}\label{Gamma_r}
  \Gamma_k(\rv) &= \sum_u \frac{\partial}{\partial r_u} \left[ \Imat_{ku}^{-1} (\rv) \right] \nonumber \\
  & = \sum_u \frac{\partial}{\partial r_u} \left[ \delta_{u=k} \frac{r_u}{M^{(L+1)}} \right] \nonumber \\
  & = 1/M^{(L+1)}.
\end{align}

\section{Hierarchical topics learned from Wiki}
In this section, we present more examples of the hierarchical topics learned from the Wiki dataset in Figs. \ref{fig:topic_wiki1} and \ref{fig:topic_wiki2}, where Fig. 1 shows a subnetwork from the same topic tree as Fig. 5 in the main manuscript does, while Fig. 2 is obtained based on a four-layer DATM-WHAI.

The semantic meaning of each topic and the connections between different topics are highly interpretable.
For example, the subnetwork in Fig. 2 mainly talks about ``war'' and ``government,'' shown as Topic 6, which is further divided into two subtopics focusing on ``public/law'' and ``military/battle,'' respectively.

Moreover, comparing Fig. 1 with Fig. 5 in the main manuscript, although the same or similar words might appear in different topics belonging to the same subnetwork, the differences between the two subnetworks are evident, which demonstrates that different subnetworks could capture distinct semantics existing in the corpus.

		Comparing with Fig. 5 in the main manuscript, the higher-layer topics of these two additional example subnetworks exhibit more distinct meanings.
For example, in Fig. \ref{fig:topic_wiki2}, Topics 5 and 28 at layer 3 talk about "government/public/law'' and "government/war/army.''
Similar results can also be found in Fig. \ref{fig:20News_DATM_dic} and Fig. \ref{fig:nips_DATM_dic}.
Moreover, although some similar or same words are in the different topics belonging to the same subnetwork, examining Fig. 5 in the main manuscript and Figs. \ref{fig:topic_wiki1} and \ref{fig:topic_wiki2}, for the same dataset (WIKI), the different subnetworks are evidently different from each other. %

\begin{figure}[ht]
  \centering
  \includegraphics[scale=0.25]{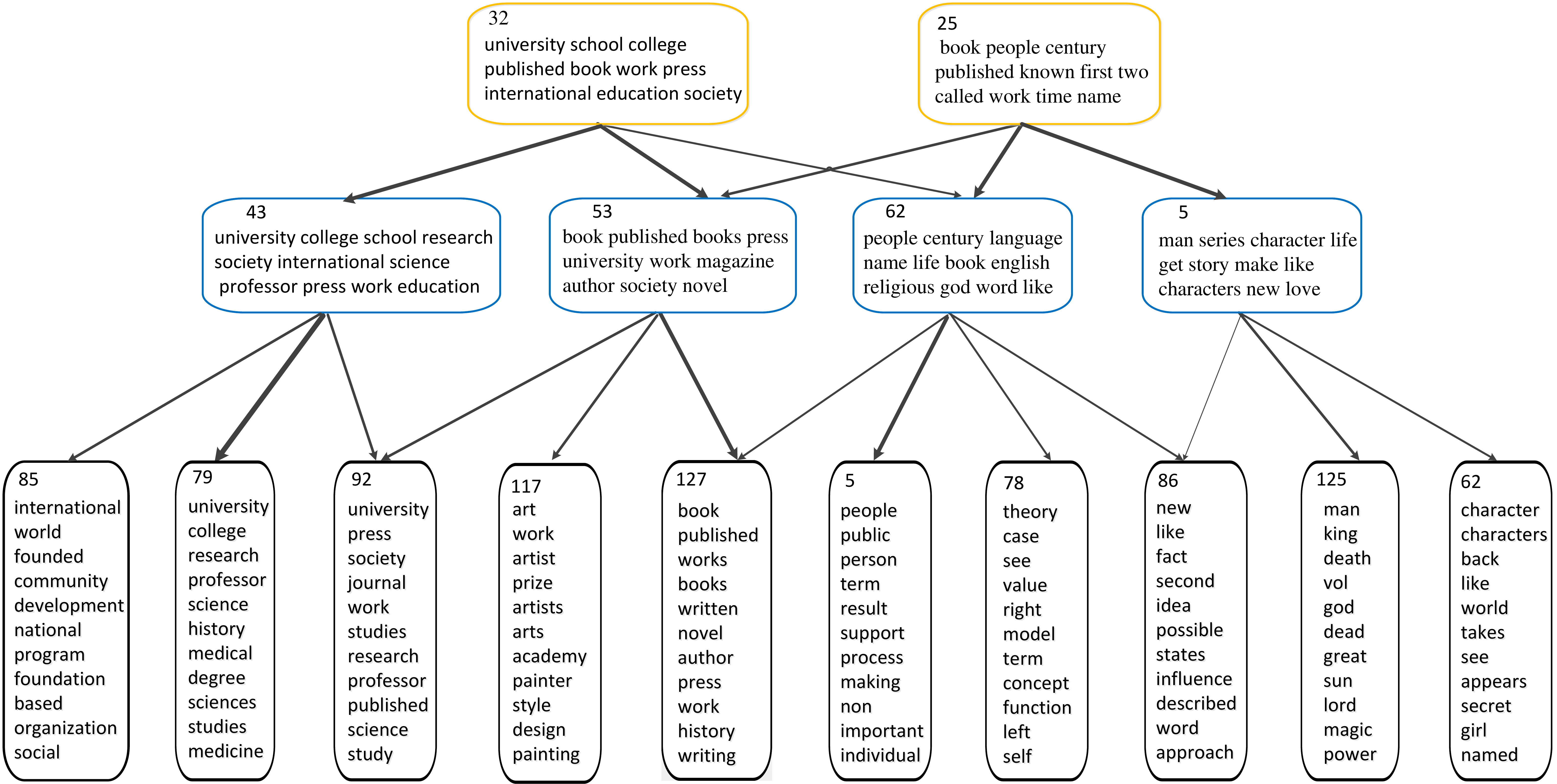}
  \caption{Example of hierarchical topics learned from Wiki by a three-layer DATM-WHAI of size 128-64-32}
  \label{fig:topic_wiki1}
\end{figure}

\begin{figure}[ht]
  \centering
  \includegraphics[scale=0.25]{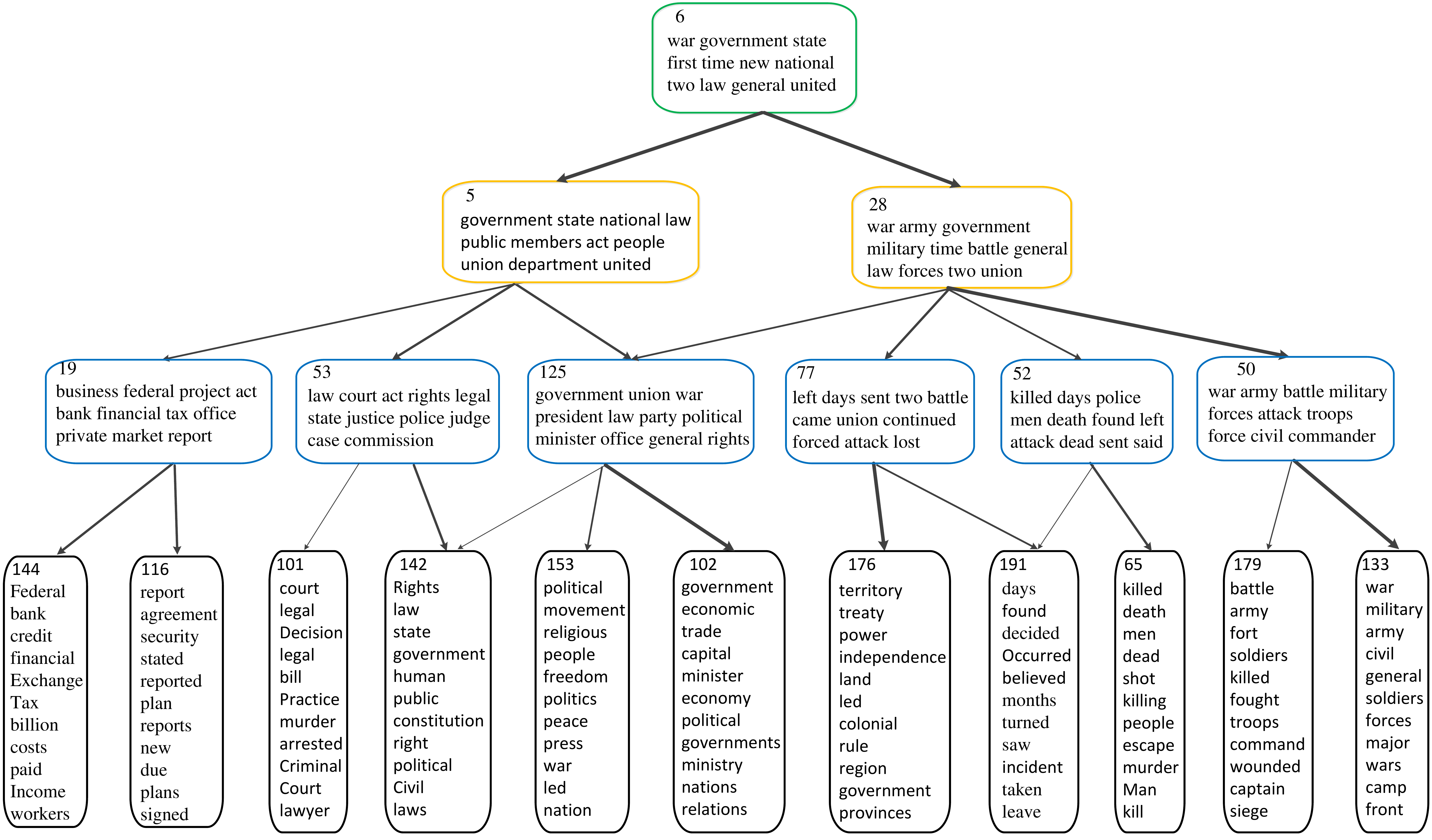}
  \caption{Example of hierarchical topics learned from Wiki by a four-layer DATM-WHAI of size 256-128-64-32}
  \label{fig:topic_wiki2}
\end{figure}

\clearpage
\section{Hierarchical topics learned from 20news and NIPS12 by DATM, hLDA and DEF}
In order to better understand the distinction of DATM in learning hierarchical topics, we compare the results of DATM with hLDA \cite{griffiths2004hierarchical} and DEF \cite{def}; we refer to the original publications on how their topics are visualized. %
Different from DATM and DEF, hLDA arranges its topics into a tree-structured $L$-level hierarchy and a document can only choose a mixture of $L$ topics along a document-specific root-to-leaf path. This construction makes the topics of hLDA closer to the root node to contain more commonly used words, as shown in Fig. 4.

On the contrary, in both DATM and DEF, a document is not restricted to choose the topics in a single path of the inferred topic hierarchy. 
DATM 
relates the topic weights of adjacent layers 
via the gamma shape parameters, %
as discussed in Eq. (2) in the main manuscript, whereas  DEF does so via %
 the gamma scale parameters.
As shown in Figs. \ref{fig:20News_DATM_dic} and \ref{fig:nips_DATM_dic} for DATM and Figs. \ref{fig:20News_DEF_dic} and \ref{fig:nips_DEF_dic} for DEF, the topics at the first layer of both models share similarities and their higher-layer topics are the weighted combinations of the lower-layer ones.
However, it is straightforward for DATM  to visualize %
its topics at higher layers. %
By contrast, %
as  in Ranganath et al. \cite{def}, the higher-layer topics of DEF need to be manually summarized. %

\begin{figure}[ht]
	\centering
	\includegraphics[scale=0.25]{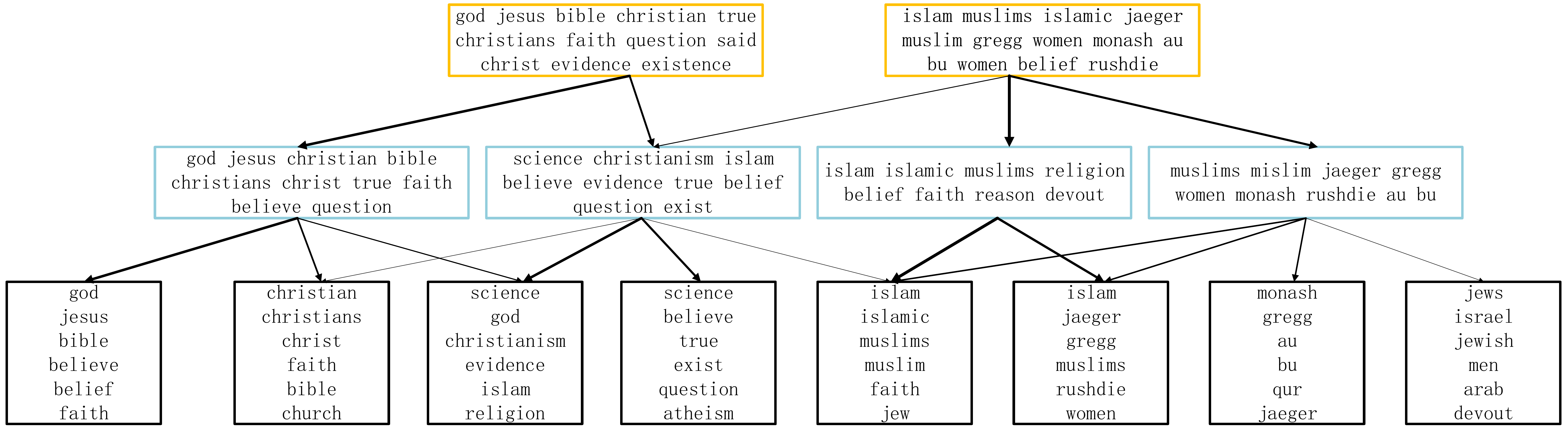}
	\caption{Example of hierarchical topics learned from 20News by a three-hidden-layer DATM-WHAI.}
	\label{fig:20News_DATM_dic}
\end{figure}

\begin{figure}[ht]
	\centering
	\includegraphics[scale=0.25]{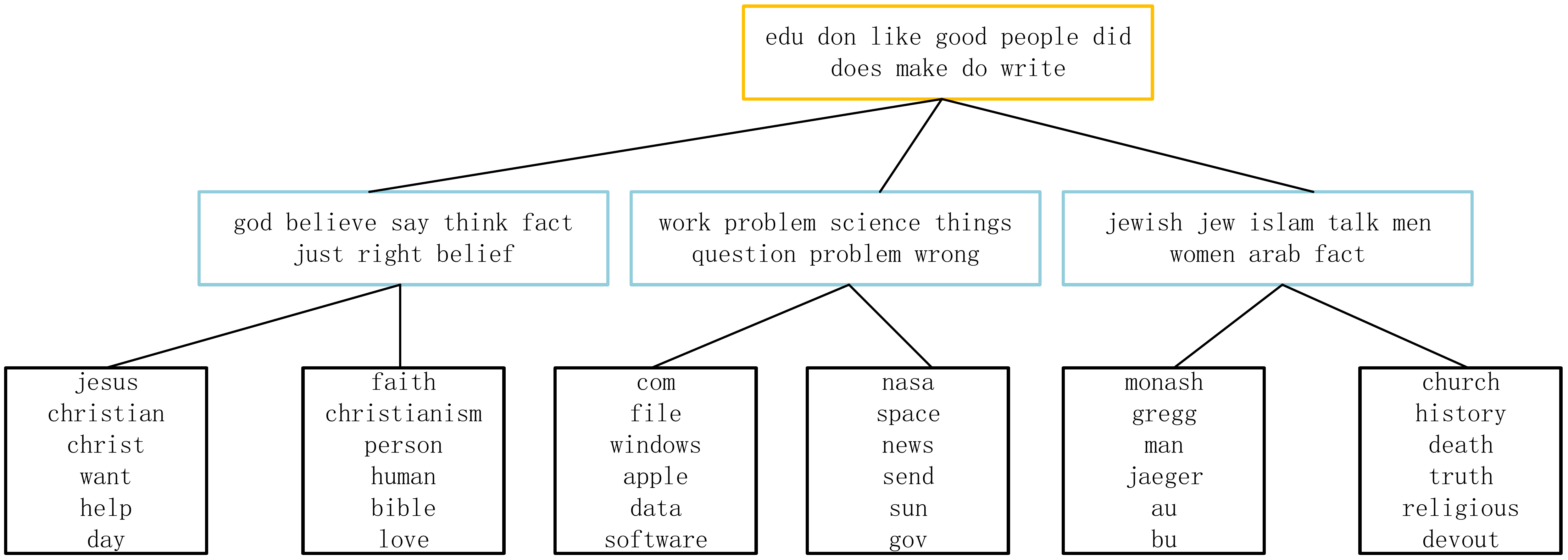}
	\caption{Example of hierarchical topics learned from 20News by a three-hidden-layer hLDA.}
	\label{fig:20News_hLDA_dic}
\end{figure}

\begin{figure}[ht]
	\centering
	\includegraphics[scale=0.25]{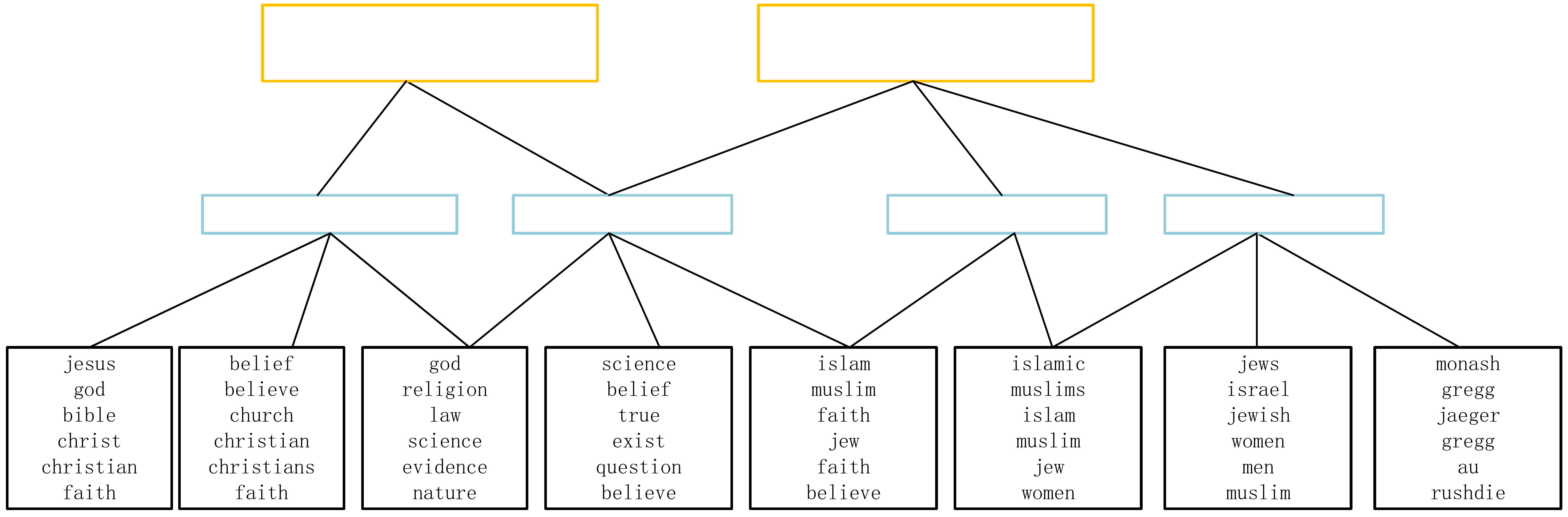}
	\caption{Example of hierarchical topics learned from 20News by a three-hidden-layer DEF. Such visualization follows Ranganath et al.\cite{def}, where the high-level topics are vacant.}
	\label{fig:20News_DEF_dic}
\end{figure}

\begin{figure}[ht]
	\centering
	\includegraphics[scale=0.18]{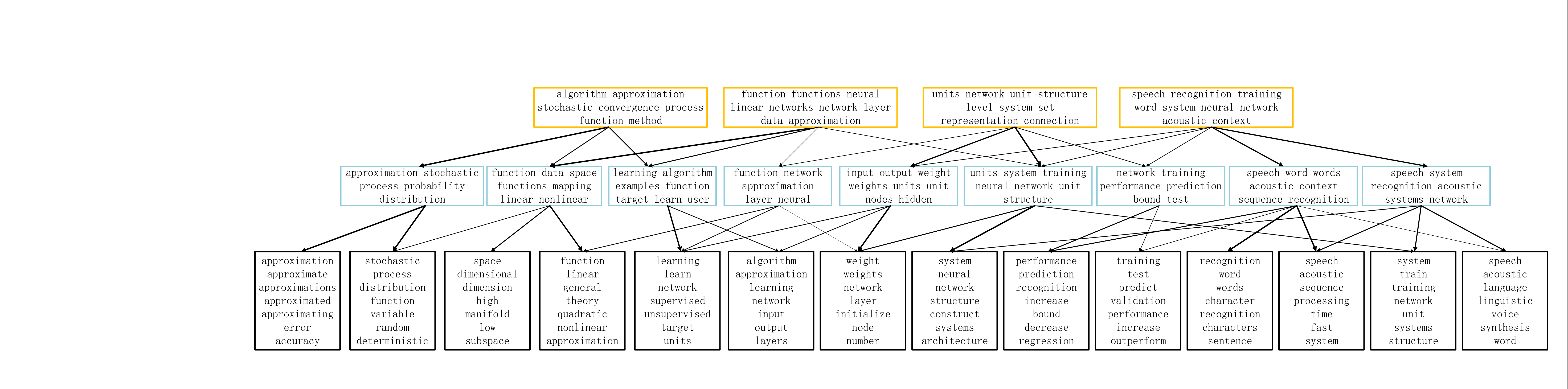}
	\caption{Example of hierarchical topics learned from NIPS12 by a three-hidden-layer DATM-WHAI.}
	\label{fig:nips_DATM_dic}
\end{figure}

\begin{figure}[ht]
	\centering
	\includegraphics[scale=0.25]{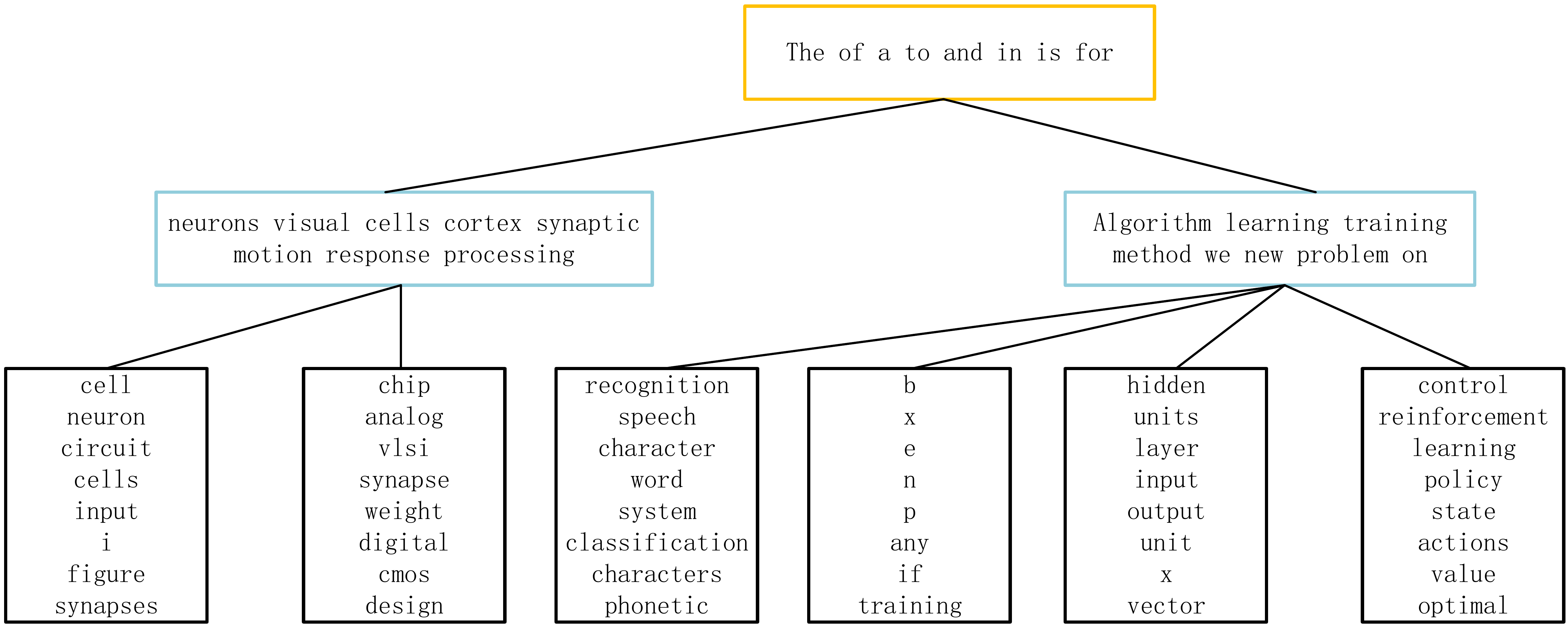}
	\caption{Example of hierarchical topics learned from NIPS12 by a three-hidden-layer hLDA (from the original paper). }
	\label{fig:nips_hLDA_dic}
\end{figure}

\begin{figure}[ht]
	\centering
	\includegraphics[scale=0.2]{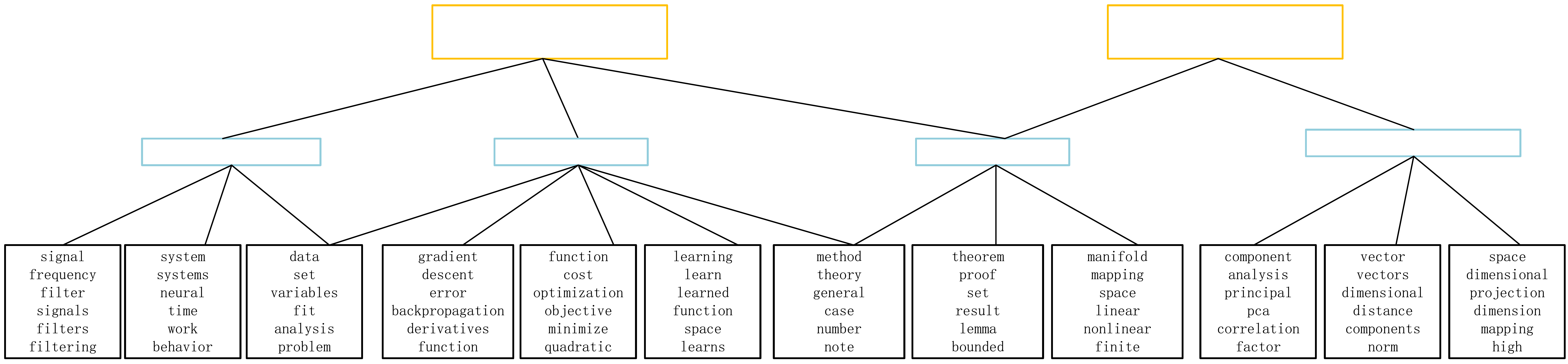}
	\caption{Example of hierarchical topics learned from NIPS12 by a three-hidden-layer DEF. Such visualization follows Ranganath et al. \cite{def}, where the high-level topics are vacant.}
	\label{fig:nips_DEF_dic}
\end{figure}

\clearpage
\section{manifold on documents}
{\bf{From a sci.medicine document to an eci.space one}}

1. com, writes, article, edu, medical, pitt, pain, blood, disease, doctor, medicine, treatment, patients, health, ibm

2. com, writes, article, edu, space, medical, pitt, pain, blood, disease, doctor, data, treatment, patients, health

3. space, com, writes, article, edu, data, medical, launch, earth, states, blood, moon, disease, satellite, medicine,

4. space, data, com, writes, article, edu, launch, earth, states, moon, satellite, shuttle, nasa, price, lunar

5. space, data, launch, earth, states, moon, satellite, case, com, shuttle, price, nasa, price, lunar, writes,

6. space, data, launch, earth, states, moon, orbit, satellite, case, shuttle, price, nasa, system, lunar, spacecraft

{\bf{From a alt.atheism document to a soc.religion.christian one}}

1. god, just, want, moral, believe, religion, atheists, atheism, christian, make, atheist, good, say, bible, faith

2. god, just, want, believe, jesus, christian, atheists, bible, atheism faith, say, make, religious, christians, atheist

3. god, jesus, just, faith, believe, christian, bible, want, church, say, religion, moral, lord, world, writes

4. god, jesus, faith, just, bible, church, christ, believe, say, writes, lord, religion, world, want, sin

5. god, jesus, faith, church, christ, bible, christian, say, write, lord, believe, truth, world, human, holy

6. god, jesus, faith, church, christ, bible, writes, say, christian, lord, sin, human, father, spirit, truth

{\bf{From a com.graphics document to a comp.sys.ibm.pc.hardware one}}

1. image, color, windows, files, image, thanks, jpeg, gif, card, bit, window, win, help, colors, format

2. image, windows, color, files, card, images, jpeg, thanks, gif, bit, window, win, colors, monitor, program

3. windows, image, color, card, files, gov, writes, nasa, article, images, program, jpeg, vidio, display, monitor

4. windows, gov, writes, nasa, article, card, going, program, image, color, memory, files, software, know, screen

5. gov, windows, writes, nasa, article, going, dos, card, memory, know, display, says, screen, work, ram

6. gov, writes, nasa, windows, article, going, dos, program, card, memory, software, says, ram, work, running

\end{document}